\documentclass[letterpaper,10pt,journal,twoside]{IEEEtran}

\IEEEoverridecommandlockouts                              

\usepackage{times}
\usepackage{graphicx}
\usepackage[caption=false]{subfig}
\usepackage{capt-of}
\usepackage{amsmath,amssymb,amsopn,amstext,amsfonts,amsthm}
\usepackage{cancel}
\usepackage[space]{cite}
\usepackage{pdfsync}
\usepackage{balance}
\usepackage{color}
\usepackage{mathtools}
\usepackage{bm}
\usepackage{diagbox}
\usepackage{float}
\usepackage{epstopdf}
\usepackage{pifont}
\usepackage{fixltx2e}
\usepackage{multirow}
\usepackage{url}
\usepackage{verbatim}
\usepackage[linkcolor=black,citecolor=black,urlcolor=black,colorlinks=true]{hyperref}
\usepackage{soul,xcolor}
\usepackage[linesnumbered,ruled,vlined]{algorithm2e}
\usepackage{lipsum}
\usepackage{subfiles}
\usepackage{array,booktabs}
\usepackage{mwe}
\usepackage{tablefootnote}

\newcolumntype{M}[1]{>{\centering\arraybackslash}m{#1}}
\SetKwComment{Comment}{$\triangleright$\ }{}

\graphicspath{{./img/}}
\DeclareGraphicsExtensions{.png,.jpg,.eps,.pdf,.jpeg}

\newcommand{\etal}{\textit{et al.}}
\newcommand{\vect}{\mathbf}
\newcommand{\matr}[1]{\mathbf{#1}}
\newcommand{\tr}{\mathrm{Tr}}
\newcommand{\Dopt}{\mathop{\mbox{$D$-$\mathit{opt}$}}}
\newcommand{\Topt}{\mathop{\mbox{$T$-$\mathit{opt}$}}}
\newcommand{\Aopt}{\mathop{\mbox{$A$-$\mathit{opt}$}}}
\newcommand{\Eopt}{\mathop{\mbox{$E$-$\mathit{opt}$}}}

\DeclareMathOperator*{\argmin}{arg\,min}

\newcommand{\TODO}[1]{\textcolor{red}{\textbf{TODO}}}

\title{RAVEN: A Unified Active SLAM System for Rapid Autonomous Visual Exploration with Aerial Robots under Localization Uncertainty}

\author{Xinyi Chen$^\dagger$, Yichen Zhang, Chenxing Jiang, and Shaojie Shen%
\thanks{$^\dagger$ Corresponding author: Xinyi Chen.

All authors are with the Department of Electronic and Computer Engineering, Hong Kong University of Science and Technology, Hong Kong, China.

\tt\footnotesize $\{$xchencq,yzhangec,cjiangan,eeshaojie$\}$@ust.hk

}%
}

\begin{document}
\IEEEaftertitletext{\vspace{-1\baselineskip}}
\maketitle

\begin{abstract}
Active simultaneous localization and mapping (SLAM) addresses the problem of planning robot motion to build complete and accurate maps of unknown environments through efficient exploration while maintaining high localization accuracy.
This raises a dilemma between exploring unknown spaces to expand coverage and exploiting visited regions to perform active loop closures.
Despite numerous prior works on ground robots, aerial robots with high agility and constrained sensor suites face more significant localization drift and thereby demand dedicated active SLAM planning.
Moreover, a unified strategy that adapts performance based on task-specific priorities remains an open question.
In this article, we introduce RAVEN, a unified active SLAM framework for Rapid Autonomous Visual ExploratioN using quadrotors, which systematically accounts for localization uncertainty.
The framework starts with identifying frontiers and active loop closure candidates, from which global tour candidates are generated.
These candidates are evaluated using a utility vector that separately quantifies exploration efficiency and localization accuracy, and the final tour is selected according to a prioritization parameter ranging from $[0,1]$.
Finally, a perception-aware navigation planner generates smooth execution trajectories that minimize visual state estimation error based on the online Fisher information field that provides visual localization information in real-time.
Benchmark evaluations demonstrate that RAVEN outperforms state-of-the-art active SLAM methods, highlighting its advantages in producing a spectrum of performance across prioritizations.
Fully onboard field experiments with a customized stereo-inertial quadrotor further validate its practical applicability in challenging environments.
The source code will be released to benefit the community.

\end{abstract}

\begin{IEEEkeywords}
Aerial Systems: Perception and Autonomy, Aerial Systems: Applications, Motion and Path Planning, Active SLAM.
\end{IEEEkeywords}

\bstctlcite{bstctl:etal, bstctl:nodash, bstctl:simpurl}

\section{Introduction}
\label{sec:intro}

\IEEEPARstart{A}{erial} robots are increasingly deployed in a wide range of real-world applications, such as search and rescue \cite{erdelj2017help}, structural inspection \cite{bircher2015structural}, and 3D reconstruction \cite{song2021view}, capitalizing on their superior maneuverability and versatility.
To achieve full autonomy in these applications, aerial robots must be capable of performing \textit{localization} of themselves within the environment, \textit{mapping} the traversed space, and \textit{planning} subsequent navigation motions.

\begin{figure}[t]
	\centering
  \includegraphics[width=1.0\columnwidth]{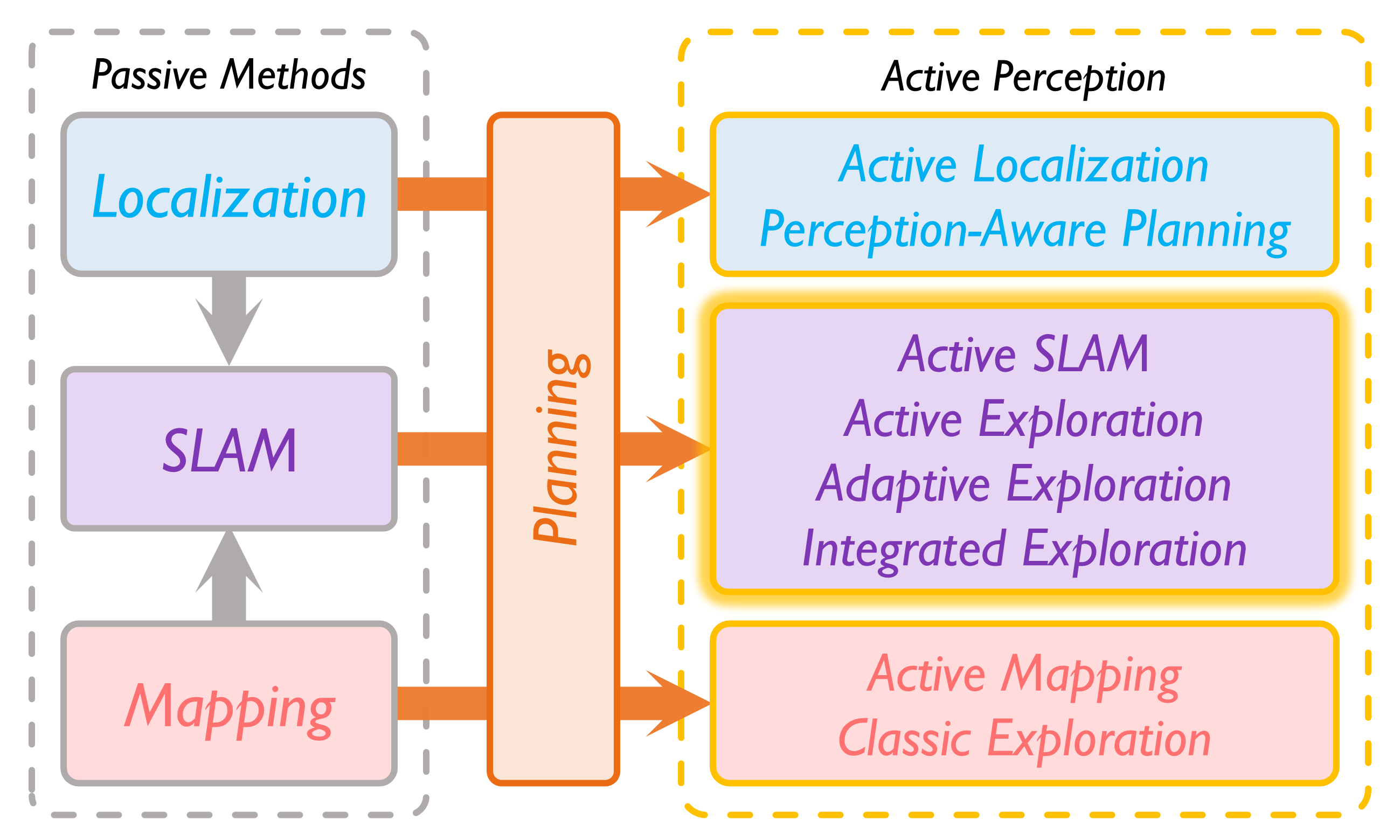}  
  \vspace{-0.8cm}
  \caption{\label{fig:field} The conceptual picture of the mobile robotics field integration and several community terminologies \cite{stachniss2004exploration,makarenko2002experiment}. The highlighted block indicates our focus on active SLAM within this broader context.}
  \vspace{-0.8cm}
\end{figure}

While the fundamental problem of simultaneous localization and mapping (SLAM) \cite{cadena2016past} often relies on passive operations, active perception \cite{bajcsy1988active,bajcsy1985active} introduces the paradigm of incorporating decision-making strategies to optimize sensor data acquisition for specific perceptual objectives.
As conceptually illustrated in Fig. \ref{fig:field}, active localization \cite{fox1998active} focuses on navigation strategies with the perceptual objective of enhancing the robot's state estimation accuracy, typically assuming prior environmental knowledge.
Active mapping \cite{connolly1985determination}, often referred to as classic exploration, addresses the problem of complete environment coverage and map construction for unknown environments under ideal localization assumptions.
Unifying both perspectives, active SLAM \cite{leung2006active} aims for the ultimate goal of constructing the most accurate and complete environmental map, through efficient exploration of the unknown environments while maintaining high localization accuracy.

Since the pioneering studies \cite{bajcsy1988active,bajcsy1985active}, the increasing attention toward active SLAM has been producing research advances from diverse perspectives and methodologies.
One of the core challenges of active SLAM is the \textit{exploration-exploitation dilemma} \cite{thrun1991active, thrun1995exploration}, where the robot must balance two conflicting objectives: exploring unknown spaces to expand mapping coverage and exploiting previously visited regions to reduce localization uncertainty through loop closures.
Various metrics have been proposed to quantify these exploration and exploitation utilities based on information theory \cite{stachniss2005information} or the theory of optimal experiment design \cite{carrillo2012comparison}.
However, the incompatible magnitudes of these fundamentally different quantities require careful manual tuning of weighting parameters for different environments, which is later circumvented by automatic balancing utility functions \cite{carrillo2018autonomous, popovic2020informative}.
Nevertheless, this implicit mechanism disregards user preferences on prioritizing either exploration efficiency or localization accuracy to meet specific task requirements.
Another challenge is that active SLAM frameworks for aerial robots remain relatively rare \cite{placed2023survey}, despite extensive research on ground robots.
The high agility of aerial robots exacerbates localization drift, which can severely degrade map quality and increase the risk of loop closure failures.
Consequently, minimizing state estimation error during navigation execution is also crucial yet often neglected by existing active SLAM methods \cite{lee2021real} that primarily focus on active loop closure.
Furthermore, aerial robots typically operate with constrained sensor suites and limited computational resources, necessitating active SLAM algorithms to balance performance with computational efficiency for real-time operation.
These aspects underscore the critical need for a systematic active SLAM framework for aerial exploration under localization uncertainty.

To bridge these gaps, this work introduces \textbf{RAVEN}, a unified active SLAM framework for \textbf{R}apid \textbf{A}utonomous \textbf{V}isual \textbf{E}xploratio\textbf{N} in unknown environments using a stereo-inertial quadrotor.
This framework is designed to ensure exploration efficiency while accounting for localization uncertainty throughout the standard active SLAM planning workflow\cite{placed2023survey}: goal identification, utility evaluation, and action execution.
In the first phase, vantage points are identified, including frontiers \cite{yamauchi1997frontier} and active loop closure (ALC) candidates.
ALC candidates are carefully selected historical poses with high potential to establish valid loop closures that improve localization accuracy.
To extend the planning horizon beyond the next immediate goal, we generate global tour candidates that cover unexplored regions and integrate active loop closures.
In the second phase, these global tour candidates are evaluated using utility vectors that separately quantify the expected exploration duration and expected pose graph information.
Instead of combining these two objectives into a single utility function, this framework treats them as separate components and allows for user-defined prioritization ranging from $[0,1]$. 
This enables system flexibility in task-specific planning behavior and unifies the classical exploration problem into the active SLAM framework.
Finally, action execution toward the target goal from the selected global tour is performed with a perception-aware planner, which generates trajectories minimizing visual state estimation error.
To support real-time replanning, an online Fisher information field (OFIF) is incrementally constructed, allowing constant-time queries of the amount of information from observations at specific camera poses for visual odometry.
In summary, the proposed framework optimizes exploration motions that balance exploration efficiency and vehicle localization accuracy according to a user-defined prioritization, producing complete volumetric maps with task-specific accuracy in unknown environments.

We conduct extensive benchmark comparisons with state-of-the-art exploration planners \cite{zhang2024falcon,carrillo2018autonomous,lee2021real,placed2022explorb} using a lightweight simulator, demonstrating that RAVEN achieves a more effective balance between exploration efficiency and localization accuracy.
Real-world experiments conducted with a fully autonomous aerial robot further validate RAVEN's practical applicability in challenging environments.
The source code will be released to benefit the community.
The contributions of the present work are as follows:

\begin{enumerate}
  \item A systematic active SLAM framework designed for autonomous aerial exploration that comprehensively accounts for localization uncertainty throughout goal identification, utility evaluation, and action execution;
  \item A trading-off mechanism to the exploration-exploitation dilemma, which eliminates the need for manual parameter tuning while allowing user-defined prioritization between exploration efficiency and localization accuracy;
  \item An incrementally constructed online Fisher information field (OFIF) for constant-time queries of visual localization information at specific poses, which facilitates real-time replanning of the active SLAM pipeline;
  \item Extensive validation through a lightweight simulator and fully onboard real-world experiments in challenging environments. The implementation will be released for reproducible benchmarking and further research.
\end{enumerate}

\bstctlcite{bstctl:etal, bstctl:nodash, bstctl:simpurl}

\section{Related Work}
\label{sec:related}
Active perception has been widely studied since the 1980s \cite{bajcsy1985active, cowan1988automatic, aloimonos1988active}.
This literature review explores studies with perceptual goals of enhancing localization, mapping, or both, under the topics of perception-aware planning (Sec.~\ref{subsec:related_pa_planning}), classical exploration (Sec.~\ref{subsec:related_exploration}), and active SLAM (Sec.~\ref{subsec:related_active_slam}).

\subsection{Perception-Aware Path Planning}
\label{subsec:related_pa_planning}
Perception-aware planning involves determining vehicle motion that improves localization accuracy, which is a topic first introduced as active localization in \cite{fox1998active}.
Various metrics have been developed to evaluate the information gain of potential actions, bridging perception and planning.

Sabdat \etal \cite{sadat2014feature} propose a perception quality metric that combines mesh triangle density as texture quality and surfaces viewing angle as area visibility from a given viewpoint, and integrate it into Rapidly-Exploring Random Trees (RRT) \cite{lavalle98rapidly}.
Falanga \etal \cite{falanga2018pampc} introduce a perception-aware model predictive control (MPC) framework that maximizes the visibility of points of interest while minimizing their velocities in the image frame.
Bartolomei \etal \cite{bartolomei2020perception} utilize semantic segmentation to avoid textureless and problematic areas by penalizing navigation away from perceptually informative regions.
Murali \etal \cite{murali2019perception} propose a yaw trajectory optimization algorithm that ensures both feature covisibility and motion agility with a predetermined position trajectory.
Chen \etal \cite{chen2024apace} generate trajectories for quadrotors that reduce state estimation error by maximizing the number of covisible features while maintaining small parallax angles between consecutive frames.
For perception-aware planning with hard constraints, Spasojevic \etal \cite{spasojevic2020perception} solve for the minimum time trajectory with a novel trajectory parametrization, while maintaining a given set of landmarks within the FoV.
Penin \etal \cite{penin2017vision} address a minimum-time optimization problem with a hard constraint on keeping certain features visible, which is later extended to consider obstacle avoidance and vision occlusion \cite{penin2018vision}.

A more general information gain metric is the Fisher information matrix (FIM), which quantifies how well the vehicle state can be estimated.
Costante \etal \cite{costante2016perception, costante2018exploiting} incorporate photometric information of the environment into localization uncertainty, evaluated using FIM and integrated into the RRT-based planner.
Kim \etal \cite{kim2021topology} leverage global topological information of the environment and select the path that minimizes travel cost and perception cost in terms of FIM among homology classes.
Zhang \etal \cite{zhang2018perception} employ a receding horizon scheme that maximizes the reward function considering obstacle avoidance, goal progress, and perception quality derived from FIM.
To address the computational complexity of FIM that scales linearly with the number of landmarks, Zhang \etal \cite{zhang2019beyond, zhang2020fisher} propose the Fisher information field (FIF), achieving constant time complexity by storing and reusing computed Fisher information in grid cells.
However, most of the perception-aware planning methods concentrate on navigation in known environments with known visual feature points.
The proposed work extends the scope of perception-aware planning to exploration tasks that operate in unknown environments without prior knowledge.

\subsection{Classic Autonomous Exploration}
\label{subsec:related_exploration}
Classic exploration is the task of exploring unknown environments and building a complete volumetric map, which is also known as active mapping \cite{lluvia2021active}.
Although studies under this category may have different focuses, such as exploration efficiency \cite{gao2018improved}, object-centric exploration \cite{papatheodorou2023finding}, and multi-robots collaboration \cite{zhou2023racer}, they have a common assumption of perfect vehicle localization.

\subsubsection{Classical Methods}
\label{subsubsec:related_exploration_classical}
Sampling-based exploration methods randomly generate candidate viewpoints using RRT \cite{lavalle98rapidly} and select the next best view (NBV) \cite{connolly1985determination} based on information gain.
The concept of NBV was first introduced to autonomous exploration by Bircher \etal \cite{bircher2016receding, bircher2018receding} using a receding horizon scheme.
Based on NBV, several subsequent works employ various information gain metrics for different applications.
Witting \etal \cite{witting2018history} incorporate navigation history positions into the sampling scheme to facilitate the quick finding of informative regions, which overcomes the curse of dimensionality as the explored area increases.
Dharmadhikari \etal \cite{dharmadhikari2020motion} sample dynamically feasible motion primitives that exploit the aerial robot dynamics and achieve a higher exploration rate.
To ensure scalability, Duberg \etal \cite{duberg2022ufoexplorer} maintain a graph structure for efficient path planning and drive the robot to the closest unknown space with a simple exploration heuristic.

Frontier-based exploration has gained popularity since Yamauchi \cite{yamauchi1997frontier, yamauchi1998frontier} defined frontiers as the boundary between known and unknown space in the 1990s.
Freda \etal \cite{freda2005frontier} incorporate frontiers in the Sensor-based Random Tree (SRT) method to guide exploration motion towards frontier regions.
Shen \etal \cite{shen2012stochastic, shen2012autonomous} adopts a stochastic differential equation to identify the most significant particle expansion region as the next target frontier in 3D environments.
Dornhege \etal \cite{dornhege2013frontier} adapt 2D frontier-based method to 3D environments by introducing void cells to efficiently determine high-visibility locations for autonomous victims searching.
Zhu \etal \cite{zhu20153d} extend the concept of frontiers to 3D space with the OctoMap representation \cite{wurm2010octomap} and select the closest frontier as the next goal.
Senarathne \etal \cite{senarathne2016towards} exploit the computationally efficient surface frontiers rather than free-space frontiers using OctoMap to actively map the object surfaces.
Cieslewski \etal \cite{cieslewski2017rapid} selects the frontier that minimizes the velocity change to maintain high flight speed for rapid exploration completion.

\subsubsection{Hybrid Methods}
\label{subsubsec:related_exploration_hybrid}
Since sampling-based methods allow for explicit quantification of information gain from different viewpoints, while frontier-based methods offer a more computationally efficient way to identify informative regions, hybrid approaches have been proposed to combine the advantages of both.
Unlike the greedy strategy of most classical methods, which selects the next target based on immediate rewards, hybrid methods often consider global information as a global guidance of the exploration.

Dai \etal \cite{dai2020fast} propose a hybrid strategy that samples candidate viewpoints in frontier regions and evaluates them based on the expected information gain over time.
Selin \etal \cite{selin2019efficient} utilize frontier-based exploration for global planning, while employing a receding horizon next-best-view planning approach for local exploration.
Zhou \etal \cite{zhou2021fuel} first generate efficient global tours for all frontiers and then locally refine the tour with viewpoints sampled around these frontiers.
Tang \etal \cite{tang2023bubble}, Yu \etal \cite{yu2023echo}, and Zhao \etal \cite{zhao2023autonomous} improve viewpoints generation, decision determination, and adaptive yaw planning based on \cite{zhou2021fuel}, demonstrating higher exploration efficiency.
Yang \etal \cite{yang2021graph} adopt a bifurcated planning strategy that explores locally distinctive frontier regions while being guided by a sparse global topological graph.
Cao \etal \cite{cao2021tare, cao2021exploring, cao2023representation} plan a global path through subspaces requiring detailed exploration and then sample viewpoints to locally cover these subspaces.
Recently, coverage path (CP) of the entire unexplored space has been explored as a more effective means of global guidance.
Song \etal \cite{song2018surface, song2020online} propose a global coverage planning algorithm that partitions the entire map into sectors, along with a local inspection planning algorithm that samples informative viewpoints of low-confidence surfaces.
Zhou \etal \cite{zhou2023racer} extends \cite{zhou2021fuel} by integrating a global coverage path of hgrids \cite{ericson2004real}, which guides the swarm agents to explore different unknown regions with balanced workloads.
Zhang \etal \cite{zhang2024falcon} incrementally divide the space with connectivity awareness for efficient coverage path construction and consciously incorporate the intention of global guidance into local planning, demonstrating a significant improvement in exploration efficiency.
Despite developments and advancements in autonomous exploration frameworks, the absence of localization consideration can lead to drifted state estimation and distorted map reconstruction.

\subsection{Active SLAM}
\label{subsec:related_active_slam}
Active SLAM has been referred to by various terminologies in the past \cite{placed2023survey}, including active exploration \cite{thrun1991active}, adaptive exploration \cite{feder1999adaptive}, and integrated exploration \cite{makarenko2002experiment}.
The following reviews literature under these topics, categorized according to the types of utility functions used for action evaluation, which is one of the core modules of active SLAM systems. 

\subsubsection{Heuristic Decision-Making}
\label{subsec:related_active_slam_heuristic}
A straightforward exploration planning strategy can be defined using heuristic thresholds to switch between exploring unknown space and revisiting known regions.
Stachniss \etal \cite{stachniss2004exploration} identify active loop closure candidates that are close on the grid map but distant along the historical trajectory, and activate the revisit policy when the vehicle uncertainty exceeds a certain threshold.
A similar active loop closure strategy has been adopted in a frontier-based exploration framework \cite{lee2021real}. 
Leung \etal \cite{leung2006active, leung2008active} transition the planning state among exploration, map improvement, and localization improvement with increasing predefined thresholds of state estimation uncertainty, which drives the vehicle towards the nearest unknown space, poor features region, or good features region, respectively.
Zhang \etal \cite{zhang2022exploration} propose an active loop closure strategy that allows the vehicle to drop by a historical position if the cost of detouring for active loop closure is only slightly higher than that of the original route.
However, these heuristically defined thresholds require manual tuning, which may not adapt well to different environments with varying scales and complexities.

\subsubsection{Information Theory}
\label{subsec:related_active_slam_info}
Instead of direct quantification of uncertainties, information-theoretic utility functions evaluate information gain based on the uncertainty in the joint belief state of the vehicle pose and the map, building upon the foundation concept of Shannon entropy \cite{shannon1948mathematical}.
Many approaches assess the utility by the expected entropy reduction, known as expected mutual information gain (MI).
The pioneering work by Bourgault \etal \cite{bourgault2002information} computes entropy for occupancy grids \cite{elfes1989using} and proposes a utility function that linearly combines the MI of map entropy with the area summation of features and vehicle uncertainty ellipses.
Stachniss \etal \cite{stachniss2005information} compute the joint entropy of the posterior after taking an action for Rao-Blackwellized particle filters (RBPF).
Since direct computation of the joint entropy is computationally intractable, they in practical use an approximation that linearly combines the separately calculated entropies of the vehicle pose and the map.
This approximation has been widely adopted, as seen in pose graph SLAM, where Valencia \etal \cite{valencia2012active} approximate trajectory entropy by averaging individual pose marginals and add it to the map entropy. 
Vallve \etal \cite{vallve2014dense} further compute the utility of approximated joint entropy in \cite{valencia2012active} densely for the entire configuration space and drive exploration through a gradient descent.
Saravanan \etal \cite{saravanan2024fit} apply this linear utilities combination by evaluating a weighted sum of path length, map entropy gain, and Fisher information for traversability.

However, this approximation faces a scale problem, as the map entropy is often orders of magnitude larger than the vehicle entropy, requiring a manual tuning of weights\cite{cadena2016past}.
To achieve an autonomous trade-off, Carlone \etal \cite{carlone2014active} derive a measure of RBPF uncertainty using the Kullback-Leibler divergence following \cite{fox2003adapting}, which is explicitly incorporated in the expected information gain from a policy over the possible outcomes.
However, this method can only be used for particle-filter-based SLAM methods, limiting its application to other SLAM systems.
Carrillo \etal \cite{carrillo2015autonomous, carrillo2018autonomous} propose a utility function that works for any SLAM algorithms based on Shannon-Rényi entropy, fusing map and robot uncertainties with an automatic weighting parameter.
Nevertheless, the trade-off between exploration and exploitation in these autonomous strategies is implicit and fails to incorporate user preference.

\subsubsection{Theory of Optimal Experiment Design (TOED)}
\label{subsec:related_active_slam_toed}
In contrast to heuristically switching state among exploration modes, utility functions have been proposed to evaluate potential actions, including frontier visitation and active loop closure.
One class of utility functions attempts to quantify uncertainties directly in the task space using the theory of optimal experiment design (TOED) \cite{pazman1986foundations}.
This theory enables comparison of actions based on their associated covariance matrices through various optimality criteria functions, such as T-optimality, A-optimality, E-optimality, and D-optimality.
Formal definitions and comparisons of these optimality criteria can be found in \cite{placed2022general, carrillo2012comparison} and are briefly introduced in Sec.~\ref{subsec:preli_toed}.

Feder \etal \cite{feder1999adaptive} were the first to apply optimality criteria in active SLAM, computing utility as the area of error ellipses, which is essentially D-optimality.
Sim \etal \cite{sim2005global} show the drawbacks of the traditional D-optimality definition and instead use T-optimality in a breadth-first search for action sequences.
Tao \etal \cite{tao20243d} also utilize T-optimality to evaluate the information gain of loop closure candidates in a metric-semantic active SLAM framework.
Carrillo \etal \cite{carrillo2012comparison} analyze and disregard the traditional definition of D-optimality as an informative metric, while showing that Kiefer's formulation of D-optimality \cite{kiefer1974general} is an effective uncertainty measure for active SLAM.
Carrillo \etal \cite{carrillo2015monotonicity} further conclude that only D-optimality guarantees the monotonicity of the uncertainty measure, which is a crucial assumption in active SLAM. This finding established D-optimality as the most appropriate optimality criterion and has been widely used in subsequent research since then.
For instance, Papachristos \etal \cite{papachristos2017uncertainty, papachristos2017autonomous, papachristos2019autonomous} employ a two-step sampling-based receding horizon planning paradigm that first generates exploration viewpoints and then plans local paths optimizing the D-optimality of the vehicle pose and the tracked landmarks covariance.
Suresh \etal \cite{suresh2020active} introduce a saliency-aware underwater exploration framework that directly evaluates the expected uncertainty with D-optimality in the utility function of revisit policy, which is triggered once the vehicle uncertainty exceeds a threshold.
Placed \etal \cite{placed2022explorb} implicitly combines the exploration information gain into the pose graph's Hessian matrix and evaluates graph uncertainty through D-optimality as formulated in \cite{placed2022general}.
Bai \etal \cite{bai2024multi, bai2024graph} employ similar pose graph uncertainty evaluation to plan active loop closures for 2D exploration tasks with prior topo-metric knowledge of the environments.
Placed \etal \cite{placed2020deep} further formulate a deep reinforcement learning paradigm for active SLAM, incorporating traditional utility functions with D-optimality in the rewards.
However, despite careful derivation of the utility function, many of these methods employ a shortsighted planning strategy that selects the next target greedily and neglects global information of the entire exploration space.

\section{Preliminaries}
\label{sec:preliminary}

\subsection{Optimality Criteria Based on TOED}
\label{subsec:preli_toed}
Following the theory of optimal experiment design (TOED), Kiefer \cite{kiefer1974general} introduced a general expression of optimality criteria, which compresses the covariance matrix $\matr{\Sigma}$ using a real-valued mapping with a parameter $p$, defined as
\begin{equation}
    \label{eqn:p-opt}
    ||\cdot||_p: \mathbb{R}^{n\times n} \to \mathbb{R}, \quad ||\matr{\Sigma}||_p \triangleq \Bigl(\frac{1}{n}\tr(\matr{\Sigma}^p)\Bigr)^\frac{1}{p}.
\end{equation}
The general formulation (\ref{eqn:p-opt}) yields a family of optimality criteria, including $\Topt, \Aopt, \Dopt$, and $\Eopt$, corresponding to $p$ values of $1$, $-1$, $0$, and $\pm\infty$, respectively.
Definition (\ref{eqn:p-opt}) can be expressed in terms of the eigenvalues $\{\lambda_1, \dots, \lambda_n\}$ of $\matr{\Sigma}$ depending on $p$ value \cite{pukelsheim2006optimal}.
For brevity, we only outline the eigenvalue expressions for the optimality criteria used in this paper, namely T-optimality ($p=1$) and D-optimality ($p=0$):
\begin{align}
    \Topt(\matr{\Sigma}) & = \frac{1}{n}\tr(\matr{\Sigma}) = \frac{1}{n}\sum_{i=1}^{n}\lambda_i, \label{eqn:topt} \\
    \Dopt(\matr{\Sigma}) & = \exp\Bigl(\frac{1}{n}\sum_{i=1}^{n}\log\lambda_i\Bigr), \label{eqn:dopt}
\end{align}
where (\ref{eqn:topt}) follows directly from the definition and the proof for (\ref{eqn:dopt}) is provided in Appendix~\ref{apx:a}.

\subsection{Graph Theory}
\label{subsec:preli_graph_theory}
Given an undirected graph $\mathcal{G} = (\mathcal{V}, \mathcal{E})$  with $n$ vertices and $m$ edges, the \textit{incidence matrix} $\matr{B} \in \{-1,0,1\}^{n\times m}$ of $\mathcal{G}$ captures the relationship between vertices and edges. For each pair of vertex $\vect{v}_i$ and edge $\vect{e}_j = (\vect{v}_{j_1}, \vect{v}_{j_2})$, the entry $\matr{B}(i,j)$ is valued $1$ if $\vect{v}_i = \vect{v}_{j_1}$, $-1$ if $\vect{v}_i = \vect{v}_{j_2}$, and $0$ otherwise. 
The \textit{Laplacian matrix} $\matr{L} \in \mathbb{R}^{n\times n}$ of an unweighted graph $\mathcal{G}$ and the \textit{weighted Laplacian matrix} $\matr{L}_\gamma \in \mathbb{R}^{n\times n}$ of a weighted graph $\mathcal{G}_\gamma$ are defined as
\begin{equation}
    \label{eqn:laplacian}
    \matr{L} = \matr{B}\matr{B}^T = \sum_{j = 1}^m \vect{b}_j \vect{b}_j^T, \quad \matr{L}_\gamma = \sum_{j = 1}^m \gamma_{j} \vect{b}_j \vect{b}_j^T,
\end{equation}
where $\vect{b}_j$ is the $j$-th column vector of $\matr{B}$, and $\gamma_{j}$ is the weight associated with the edge $\vect{e}_{j}$.

The spectrum of the Laplacian matrix is closely related to the graph's topology.
For a connected graph $\mathcal{G}$, the eigenvalues of $\matr{L}$ are $(\lambda_1=0, \lambda_2,\dots, \lambda_n)$.
By Kirchhoff's matrix tree theorem, the number of spanning trees $t(\mathcal{G})$ in $\mathcal{G}$ is
\begin{equation}
    t(\mathcal{G}) = \frac{1}{n} \prod_{i=2}^n \lambda_i = \det{(\matr{L}')}, \quad t(\mathcal{G}_\gamma) = \det{(\matr{L}_\gamma')}
\end{equation}
where $\matr{L}'$ ($\matr{L}_\gamma'$) is the \textit{reduced Laplacian} obtained by removing an arbitrary row and column from $\matr{L}$ ($\matr{L}_\gamma$).

\subsection{Uncertainty Quantification for a Pose Graph}
\label{subsec:preli_pg_uncertainty}
Pose graph is a widely used graph representation in graph-based SLAM\cite{grisetti2010tutorial}.
A pose graph $\mathcal{G}(\mathcal{V}, \mathcal{E})$ attaches robot poses $\vect{x}_i$ to vertices and each edge between vertices $i,j$ encodes a relative pose measurement $\vect{z}_{ij}$ along with its covariance matrix $\matr{\Sigma}_{ij}$.
The problem of solving SLAM estimation using a pose graph, known as pose graph optimization (PGO), involves computing the maximum likelihood estimation \cite{carlone2016planar} of the robot poses given the relative measurements.
For the robot poses $\vect{x} = (\vect{x}_1^T, \dots, \vect{x}_n^T)^T$, the PGO objective function $\matr{F}(\vect{x})$ can be formulated as follows and solved using nonlinear optimization such as the Gauss-Newton method \cite{kummerle2011g},
\begin{equation}
    \label{eqn:pgo}
    \vect{x}^* = \arg\min_{\vect{x}} \matr{F}(\vect{x}) = \arg\min_{\vect{x}}\sum_\mathcal{E} \vect{e}_{ij}(\vect{x})^T \matr{\Sigma}_{ij} \vect{e}_{ij}(\vect{x}),
\end{equation}
where $\vect{e}_{ij}(\vect{x})$ is the residual error of the relative pose between pose $\vect{x}_i$ and $\vect{x}_j$.

For active SLAM problems, uncertainty quantification for the entire pose graph is crucial to assess localization inaccuracies during the whole navigation process. 
However, computation of the full covariance or Fisher information matrix (FIM) of the entire pose graph becomes intractable as the size of the pose graph increases, making downstream online algorithms impractical.
To address this issue, Placed \etal \cite{placed2022general} derived an approximation for the FIM optimality criteria of the entire pose graph by analyzing the pose graph topology.
Specifically, for a general pose graph $\mathcal{G}$, an undirected weighted graph $\mathcal{G}_\gamma$ can be constructed correspondingly with edges weighted by $\gamma_{ij} = \Dopt(\mathcal{I}_{ij}) = \Dopt(\matr{\Sigma}_{ij}^{-1})$, where $\mathcal{I}_{ij}$ and $\matr{\Sigma}_{ij}$ are the FIM and covariance matrix of the measurement $\vect{z}_{ij}$.
The D-optimality of the entire graph FIM can be approximated as
\begin{equation}
    \label{eqn:pgo_fim}
    \Dopt(\mathcal{I}) \approx \Dopt(\matr{L}_\gamma) = (nt(\mathcal{G}_\gamma))^\frac{1}{n} = (n \det(\matr{L}_\gamma'))^\frac{1}{n},
\end{equation}
where $\matr{L}_\gamma'$ is the reduced Laplacian matrix of the graph $\mathcal{G}_\gamma$, and the last two equations are proved in Appendix~\ref{apx:a}.
In this work, we focus on the D-optimality criterion due to its monotonicity property \cite{carrillo2015monotonicity}, which makes it particularly suitable for active SLAM problems.
For detailed derivations and results on other optimality criteria, readers are referred to \cite{placed2022general}.

\begin{figure*}[t]
	\centering
  \includegraphics[width=2.0\columnwidth]{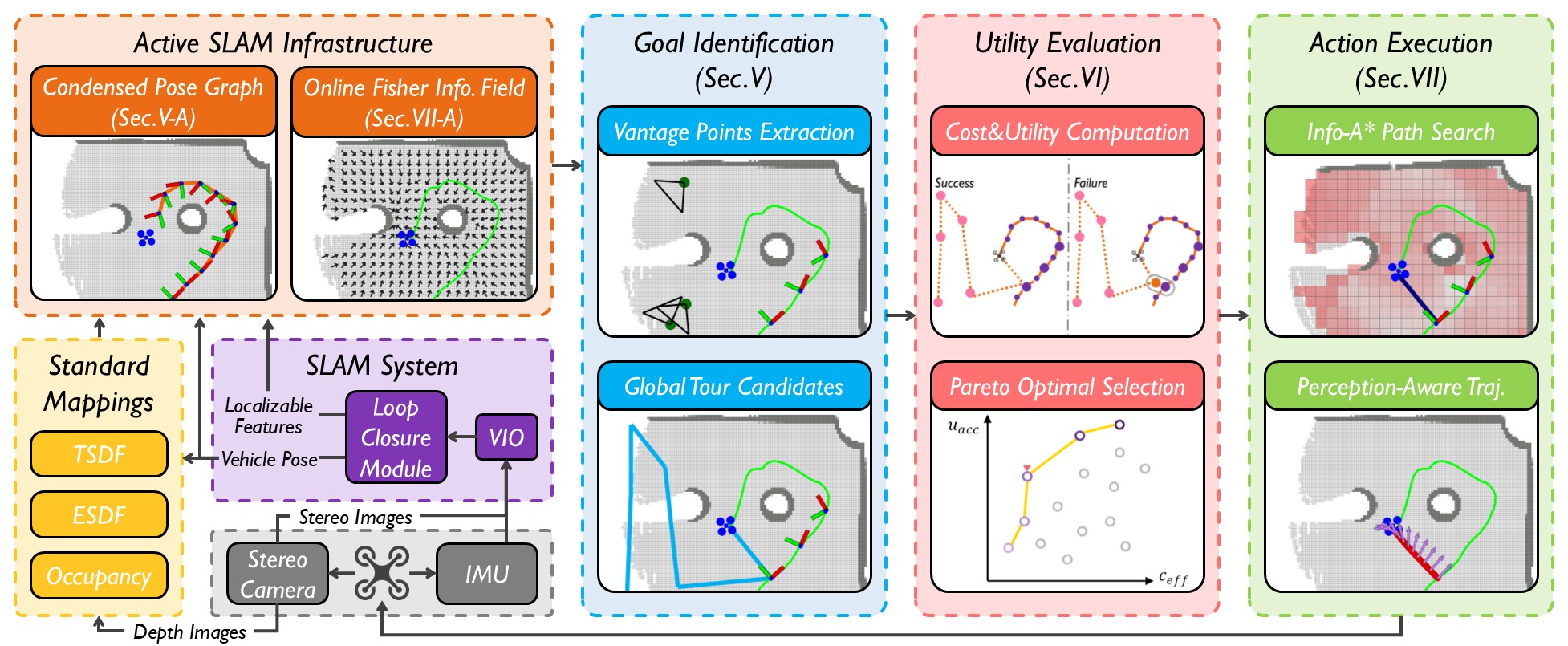}  
  \vspace{-0.2cm}
  \caption{\label{fig:system_overview} System overview of the \textbf{RAVEN} active SLAM framework for autonomous exploration.
  After updating vehicle pose estimation and standard volumetric maps with onboard sensors data, the online Fisher information field (OFIF) and condensed pose graph (CPG) are incrementally constructed to facilitate responsive planning. 
  The active SLAM planning proceeds through three phases: goal identification, utility evaluation, and action execution. 
  The framework consistently updates and replans based on the latest sensor measurements until exploration concludes.}
  \vspace{-0.2cm}
\end{figure*}

\section{System Overview}
\label{sec:system_overview}

\subsection{Problem Definition}
\label{subsec:problem_def}
Active SLAM is a decision-making problem in which a robot must determine a sequence of actions that allows it to collect sensor data for constructing the most accurate and complete map representation of an unknown environment.
Consider a robot performing SLAM that uses its sensor data to incrementally build a map representation $\mathbf{m}_t\in\mathcal{M}$ and estimate its pose $\mathbf{x}_t\in\mathcal{X}$, which jointly define the state space $\mathcal{S} \triangleq \mathcal{M} \times \mathcal{X}$.
Formally, the active SLAM problem can be formulated as a particularization of the partially observable Markov decision process (POMDP) as follows \cite{placed2023survey}
\begin{equation}
  \label{eqn:prob_def}
  \boldsymbol{a}_{t: t+k}^{\star}=\underset{\boldsymbol{a}_{t: t+k} \in \mathcal{A}^k}{\arg \max } \sum_{\tau=t}^{t+k} u(b(\boldsymbol{s}_\tau), \boldsymbol{a}_\tau),
\end{equation}
where $\boldsymbol{a}_{t: t+k}^{\star}$ is the optimal sequence of actions over a $k$-step planning horizon, $\mathcal{A}^k = \mathcal{A} \times \cdots \times \mathcal{A}$ denotes the action space, $b(\boldsymbol{s}_\tau) \triangleq p(\boldsymbol{s}_\tau \mid \boldsymbol{z}_{1: \tau}, \boldsymbol{a}_{1: \tau-1})$ is the belief over the state posterior at time $\tau$, and $u$ is a scalar utility function that evaluates the quality of a belief-action pair.
Active SLAM for autonomous exploration addresses the problem of efficiently covering an unknown and bounded 3D task space $V \subset \mathbb{R}^3$ while maintaining high localization accuracy in GPS-denied environments, jointly contributing to building a complete and accurate volumetric map of the accessible space $V_\text{acc} \subset V$.

\subsection{Proposed Framework Overview}
\label{subsec:overall_framework}

An overview of the proposed active SLAM framework is depicted in Fig. \ref{fig:system_overview}. 
The modular SLAM architecture comprises a visual-inertial odometry (VIO) module that fuses images and inertial measurement unit (IMU) measurements to provide 6-DOF state estimates, which are rectified by the loop closure module upon detecting revisited locations.
The rectified state estimates and depth measurements are used to update the standard mappings.
The exploration planning follows the widely recognized workflow for active SLAM algorithms \cite{placed2023survey, cadena2016past, lluvia2021active} consisting of three main components:
\textit{(1) Goal Identification (Sec.\ref{sec:goal_iden}):} Extract vantage points for both exploration and exploitation, then generate global tours as evaluation candidates that cover unexplored space and integrate active loop closures.
\textit{(2) Utility Evaluation (Sec.\ref{sec:util_eval}):} Compute the expected utility vectors for each global tour candidate to quantify exploration efficiency and localization accuracy, then select the final global tour balancing exploration and exploitation based on a user-defined prioritization parameter.
\textit{(3) Action Execution (Sec.\ref{sec:local_planning}):} Generate smooth and safe perception-aware trajectories towards the target viewpoint using the Online Fisher Information Field, which minimize vehicle visual odometry drift during navigation.
The entire framework including the above three phases is consistently replanned, triggered by frontier observation progress, trajectory execution completion, or a periodic timer.
The exploration task is terminated when no frontier persists in the volumetric map, which means full coverage of accessible space is achieved.

\section{Global Tour Candidates Identification}
\label{sec:goal_iden}

This section identifies vantage points for both exploration and exploitation, then generates a finite subset of high-quality global tours for evaluation, reducing computational overhead for real-time planning.

\subsection{Condensed Pose Graph for Visual SLAM}
\label{subsec:con_pose_graph}

We incrementally construct a \textit{Condensed Pose Graph} (CPG) as a compact high-level representation of the vehicle's visual odometry trajectory estimation. 
The CPG compresses high-frequency visual state estimates into a sparse graph structure, reducing computational complexity while preserving essential information for active loop closure candidates selection (Sec.~\ref{subsec:vantage_point}) and localization accuracy utility evaluation (Sec.~\ref{subsec:loc_util}).
The CPG is a pose graph with weighted edges, where pose nodes correspond to 6-DoF camera poses uniformly selected at fixed geodesic intervals $d_\text{cpg}$ along the odometry trajectory. 
Edges between consecutive nodes encode relative transformation with covariance defined as the summation of covariance matrices related to the feature reprojection error of all intermediate SLAM keyframes, which is detailed in Appendix \ref{apx:cpg_cov}.
Whenever a loop closure is detected, an additional edge is added between the nodes closest to the loop closure frames.
The CPG operates based on the fundamental module of feature matching, thereby removing requirements on specific SLAM algorithms such as graph-based \cite{carrillo2018autonomous} or particle-filter-based \cite{carlone2014active} approaches.
This representation efficiently encapsulates visual odometry and loop closure information, ensuring adaptability of the proposed active SLAM framework across diverse SLAM algorithms.


\subsection{Vantage Points Selection}
\label{subsec:vantage_point}
\begin{figure}[t]
	\centering
  \includegraphics[width=\columnwidth]{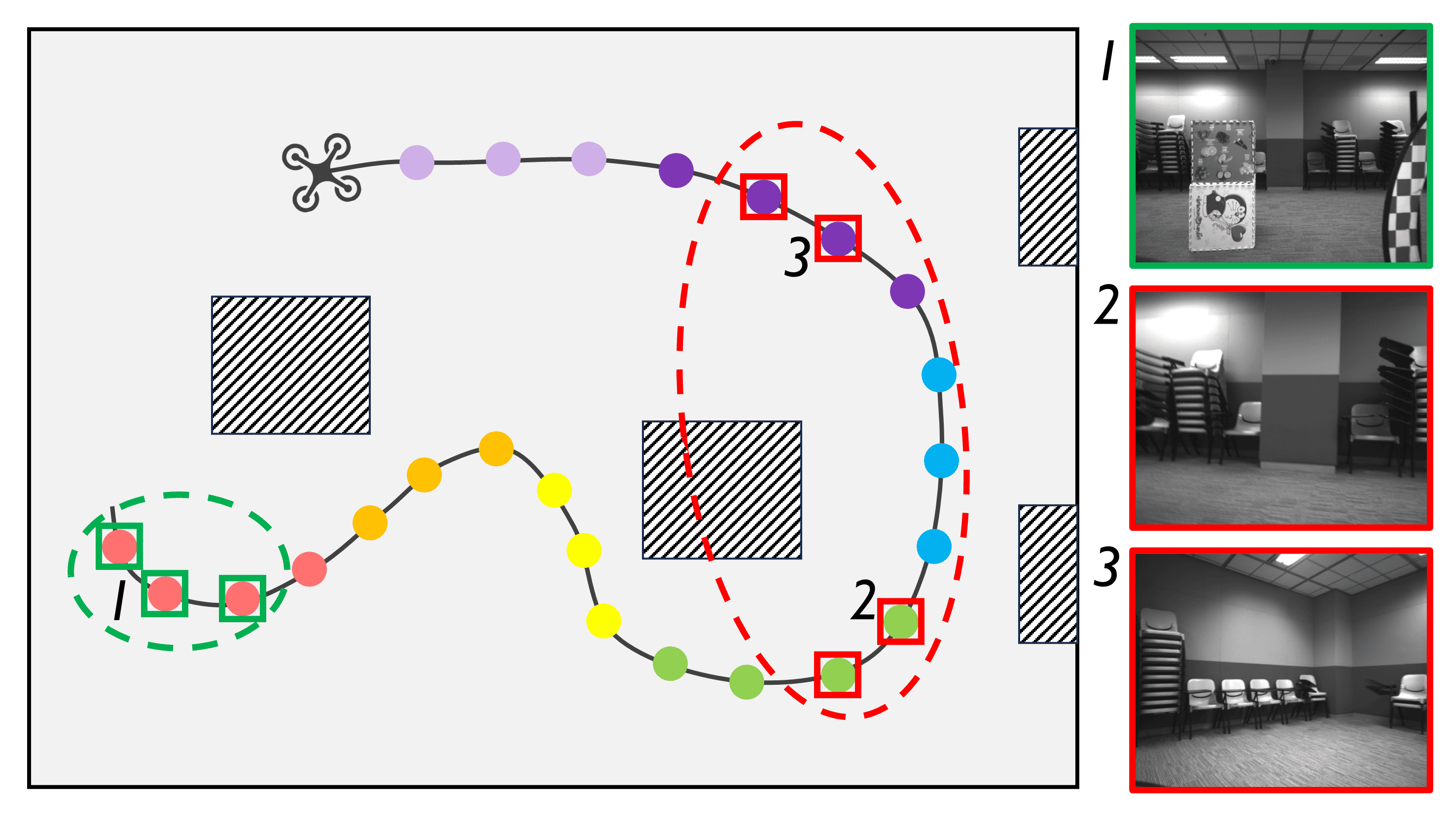}
  \vspace{-0.6cm}
  \caption{\label{fig:alc_candidate} Illustration of ALC candidate selection.
  CPG node clusters are shown in different colors. 
  The green squares and dashed ellipse highlight the similar frames queried by the pink cluster representative and their spatial distribution, while the red ones indicate those from the purple cluster representative.
  The pink representative, with concentrated frame recalls, is selected as an ALC candidate, whereas the purple one is excluded due to dispersed frame recalls.
  The right panel displays the first-person views of the numbered frames, where image 1 is visually distinctive, while image 2 and 3 are similar.}
  \vspace{-1.2cm}
\end{figure}

An active SLAM algorithm must address two objectives simultaneously: exploring unknown regions to expand map coverage and exploiting previously explored regions to improve localization accuracy through active loop closures.
For exploration vantage points, frontiers are extracted using \cite{zhou2021fuel}, which are boundaries between explored and unexplored regions \cite{yamauchi1997frontier}.
For exploitation vantage points, active loop closure (ALC) candidates are carefully chosen from the historical trajectory based on their potential to produce valid loop closure.
As shown in Fig.~\ref{fig:alc_candidate}, this process begins by clustering CPG nodes that observe similar regions based on feature co-visibility, whose representative is designated as the node that observes the most co-visible features.
For each cluster representative, we query similar nodes among all CPG nodes, whose Faiss \cite{douze2024faiss} similarity scores on the NetVLAD descriptor \cite{arandjelovic2016netvlad} exceed a threshold.
Cluster representatives with concentrated retrieved nodes are selected as ALC candidates, as they correspond to visually distinctive frames that are likely to yield valid loop closures.

\subsection{Global Tour Candidates Generation}
\label{subsec:global_tour_cand}

While many existing active SLAM approaches greedily select the next goal by maximizing immediate reward among identified frontiers and ALC candidates, such a shortsighted strategy can result in suboptimal performance due to inadequate long-term and global consideration.
To extend the planning horizon, we generate global tours that encompass the entire remaining exploration process until completion.
This is inspired by the classic exploration framework \cite{zhang2024falcon}, which optimizes exploration duration under a perfect localization assumption.
Specifically, it decomposes the voxel map of the task space $V$ into disjoint \textit{zones} based on voxel connectivity and occupancy, where zones containing frontiers or unexplored regions are designated as \textit{zones of interest}.
The \textit{coverage path} is optimized as an ordered tour through the centers of all zones of interest, minimizing the expected total duration formulated as an Asymmetric Traveling Salesman Problem.
As exploration progresses, both the zones and the coverage path are updated and replanned using the latest voxel map.

\begin{figure*}[t]
	\centering
  \includegraphics[width=2.0\columnwidth]{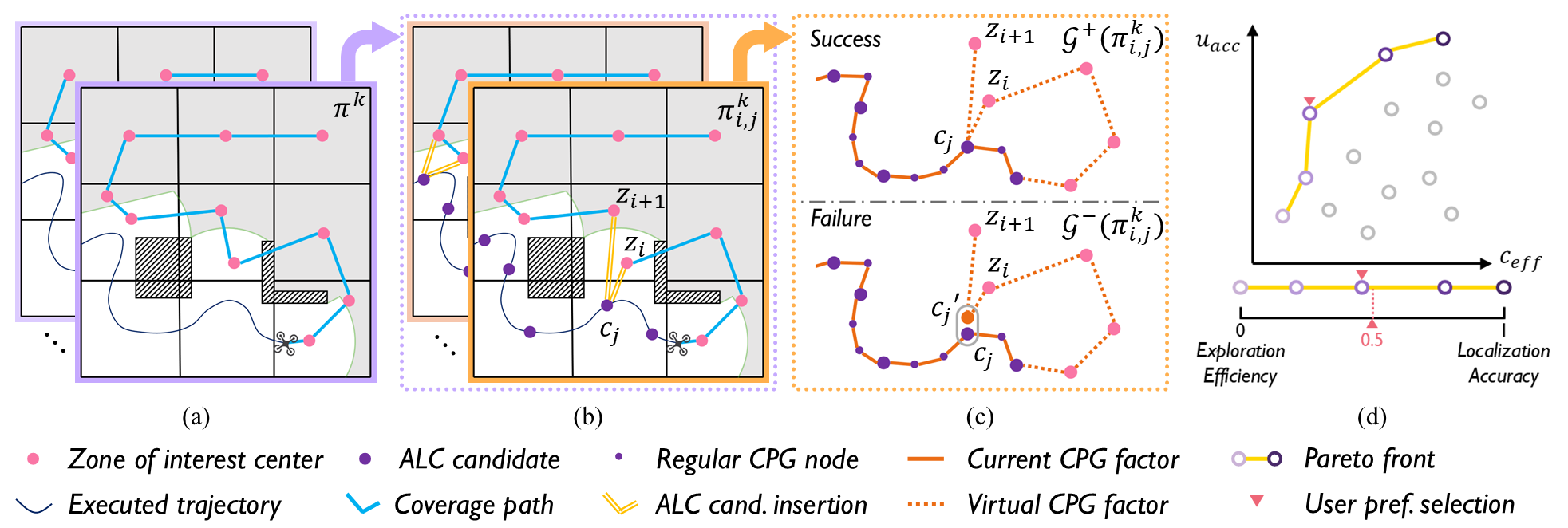}  
  \vspace{-0.2cm}
  \caption{\label{fig:selection} 
  This figure depicts the global tour candidate selection process.
  (a) The top $N_\text{cp}$ coverage paths connecting all zone of interest centers.
  (b) Global tour candidates generated by inserting an ALC candidate into the coverage paths.
  (c) The two possible configurations of virtual CPG, where the upper graph $\mathcal{G}^+$ and lower graph $\mathcal{G}^-$ represent a loop closure success and failure, respectively.
  (d) The utility space of the global tour candidates, where the yellow polyline represents the Pareto front with Pareto-optimal solutions in purple circles.
  The pink inverted triangle marks the selected global tour that is nearest to the interpolated point on the Pareto front according to prioritization parameter $\lambda=0.5$.}
  \vspace{-0.2cm}
\end{figure*}

In this active SLAM work, we consider not only exploration efficiency but also localization accuracy.
Our key idea is that \textit{'a high-quality global tour for the active SLAM problem should also serve as an efficient solution to the classic exploration problem without the ALC candidate'}.
This principle motivates a two-stage approach for generating \textit{global tour candidates}.
For exploration objective, we first solve for the top $N_\text{cp}$ coverage path base tours as a classic exploration problem following \cite{zhang2024falcon} with distance costs derived from the info-A* search discussed in Sec.\ref{sec:local_planning}, as shown in Fig.~\ref{fig:selection}(a).
For exploitation objective, we integrate only the next upcoming ALC candidate into these base tours, resulting in global tours comprising a visitation order for all zones of interest and at most one ALC candidate.
We selectively insert each ALC candidate into the base tours at edges that fall within a distance threshold, as shown in Fig.~\ref{fig:selection}(b).
This strategy efficiently generates a manageable set of high-quality global tour candidates by building upon proven efficient base tours rather than exploring the entire combinatorial space.


\section{Action Evaluation with Pareto Optimality}
\label{sec:util_eval}

Instead of combining the fundamentally different quantities of exploration and exploitation objectives into a single utility function \cite{carrillo2018autonomous}, we evaluate each global tour separately by its expected exploration efficiency cost (Sec.~\ref{subsec:eff_cost}) and localization accuracy utility (Sec.~\ref{subsec:loc_util}), targeting for a set of solutions that effectively balance both objectives (Sec.~\ref{subsec:pareto_selection}).

\subsection{Exploration Efficiency Cost}
\label{subsec:eff_cost}
The exploration efficiency cost $c_\text{eff}$ is determined by the expected task completion duration, calculated as the cumulative traversal time between consecutive elements in the tour.
For a global tour $\pi^k$ without an ALC candidate, $c_\text{eff}(\pi^k)$ can be directly obtained from the coverage path planning in Sec.~\ref{subsec:global_tour_cand}.
For a global tour $\pi^k_{i,j}$ with an ALC candidate $c_j$ inserted between zones $z_i$ and $z_{i+1}$ in $\pi^k$, the expected task completion duration is computed as
\begin{equation}
  c_\text{eff}(\pi^k_{i,j}) = c_\text{eff}(\pi^k) + t(z_i, c_j) + t(c_j, z_{i+1}) - t(z_i, z_{i+1}),
\end{equation}
where $t(\cdot,\cdot) \triangleq \text{dist}(\cdot,\cdot) / v_\text{max}$ denotes the expected traversal time between the two elements.

\subsection{Localization Accuracy Utility}
\label{subsec:loc_util}
The localization accuracy utility $u_\text{acc}$ is computed by evaluating the expected graph information of the CPG corresponding to a global tour candidate.
For a global tour $\pi^k_{i,j}$, we construct a virtual CPG by sequentially appending virtual nodes to the current CPG $\mathcal{G}_0$.
This virtual CPG has two possible configurations depending on whether the active loop closure at $c_j$ succeeds, as shown in Fig.~\ref{fig:selection}(c).
In the successful case $\mathcal{G}^+(\pi^k_{i,j})$, virtual nodes are added for each zone element up to $z_{i+1}$ that follows the ALC candidate $c_j$.
The valid loop closure produces two new edges connecting the existing node $c_j$ from $\mathcal{G}_0$ to both virtual nodes for zones $z_i$ and $z_{i+1}$.
For a failed loop closure, the virtual CPG $\mathcal{G}^-(\pi^k_{i,j})$ extends $\mathcal{G}_0$ linearly through tour elements up to zone $z_{i+1}$, where the ALC candidate is established as a new virtual node $c_j^\prime$ that shares the same pose as $c_j$.
In both configurations, the covariances for the virtual edges are set proportional to the Euclidean distances between the connected nodes. 
The virtual CPG is truncated at the node following the ALC candidate because the global tour incorporates only the upcoming ALC candidate, making the graph beyond this point temporary.

Subsequently, we estimate the success probability of an active loop closure $p_\text{succ}$ by considering whether the vehicle can reliably return to the historical pose at $c_j$ to perform loop closure.
Specifically, we compute the covariance of state estimate difference from the two visitations as $\tilde{\Sigma} \triangleq \Sigma_j - \Sigma_j^\prime$, where $\Sigma_j$ and $\Sigma_j^\prime$ are the marginal covariances of node $c_j$ in $\mathcal{G}_0$ and node $c_j^\prime$ in $\mathcal{G}^-(\pi^k_{i,j})$ respectively.
Assuming the estimation difference $\tilde{w}$ of each differential-flat state variable $w\in\{x,y,z,\psi\}$ follows a Gaussian distribution, we have
\begin{equation}
  \label{eqn:succ_prob}
  p_\text{succ} = \prod_{w\in\{x,y,z,\psi\}} P(|\tilde{w}| < \delta_w | \tilde{w} \sim \mathcal{N}(0, \tilde{\sigma}_{w})),
\end{equation}
where $\delta_w$ represents the empirical threshold of acceptable difference in variable $w$ for a successful loop closure, and $\tilde{\sigma}_{w}$ denotes the diagonal element of $\tilde{\Sigma}$ corresponding to variable $w$. 
This formulation computes the probability that the state estimate deviation remains within an acceptable threshold, indicating that the localization drift is sufficiently small for the vehicle to revisit the historical pose and achieve loop closure.
Finally, the localization accuracy utility is computed as the expected graph information of the virtual CPG,
\begin{equation}
  \label{eqn:loc_util}
  u_\text{acc}(\pi^k_{i,j}) = p_\text{succ}\cdot h(\mathcal{G}^+(\pi^k_{i,j})) + (1-p_\text{succ})\cdot h(\mathcal{G}^-(\pi^k_{i,j})),
\end{equation}
where $h(\cdot) \triangleq \Dopt(\matr{L}_\gamma(\cdot))$ is the approximated Fisher information of the entire graph according to Sec.~\ref{subsec:preli_pg_uncertainty}.

\subsection{Global Tour Selection with Pareto Optimality}
\label{subsec:pareto_selection}
The comparison of evaluated utility vectors $(c_\text{eff}, u_\text{acc})$ is performed based on Pareto optimality. 
Mathematically, \textit{Pareto order} provides a partial order for solution comparison in a multi-objective optimization problem, which maximizes a reward objective function $\vect{f} = (f_1, \dots, f_m) \in \mathbb{R}^m$.
One solution $\vect{x}$ is said \textit{dominates} another $\vect{y}$ if and only if all elements in $\vect{f}(\vect{x})$ are no less than the corresponding ones in $\vect{f}(\vect{y})$ and at least one element of $\vect{f}(\vect{x})$ is strictly larger than the corresponding element of $\vect{f}(\vect{y})$. Formally, $\vect{x} \succ_\vect{f} \vect{y}$ or $\vect{f}(\vect{x}) \succ \vect{f}(\vect{y}) \iff$ 
\begin{equation}
    f_i(\vect{x}) \geq f_i(\vect{y}), \forall i \in I \text{ and } f_j(\vect{x}) > f_j(\vect{y}), \exists j \in I,
\end{equation}
where $I = \{1,\dots,m\}$ is the index set. 
A solution $\vect{x}^*$ is called \textit{Pareto optimal} if and only if there is no other solution $\vect{x}$ in the feasible design space $X$ that dominates $\vect{x}^*$, i.e. $\nexists \vect{x} \in X$ s.t. $\vect{f}(\vect{x}) \succ \vect{f}(\vect{x}^*)$.
The \textit{Pareto front} refers to the set of all Pareto optimal solutions, which are the best possible trade-offs between objectives.
Applying Pareto optimality to our problem, the formulation (\ref{eqn:prob_def}) can be extended to
\begin{align}
  \mathcal{P}_{t: t+k} &= \underset{\boldsymbol{a}_{t: t+k} \in \mathcal{A}^k}{\operatorname{Argmax}}^\succ \sum_{\tau=t}^{t+k} \boldsymbol{u}(b(\boldsymbol{s}_\tau), \boldsymbol{a}_\tau), \\
  \boldsymbol{a}_{t: t+k}^{\star} &= S(\mathcal{P}_{t: t+k}, \lambda),
\end{align}
where $\mathcal{P}_{t: t+k}$ is the set of Pareto-optimal solutions under Pareto order $\succ$, the utility vector $\boldsymbol{u}$ quantifies exploration efficiency and localization accuracy separately, and $S(\cdot, \lambda)$ is a selection function that chooses the final action sequence based on a user-defined prioritization parameter $\lambda \in [0, 1]$.

Specifically, we treat exploration efficiency cost $c_\text{eff}$ and localization accuracy utility $u_\text{acc}$ as two separate objectives to be minimized and maximized, respectively.
After evaluating utility vectors $(c_\text{eff}, u_\text{acc})$ for all global tour candidates, the set of Pareto optimal global tours $\boldsymbol{\Pi}^*$ is extracted from the Pareto front in the utility space, as illustrated in Fig.~\ref{fig:selection}(d).
The final global tour $\pi^*(\lambda)\in\boldsymbol{\Pi}^*$ is selected as the solution nearest to the interpolated point along the Pareto front polyline according to the given prioritization parameter $\lambda \in [0,1]$ which explicitly trades off exploration efficiency and localization accuracy.
If the selected tour $\pi^*(\lambda)$ designates an active loop closure as the next action, the next viewpoint is directly set to the associated historical pose.
Otherwise, for exploration actions, the next frontier is retrieved from the target zone of interest, with viewpoints sampled around it following \cite{zhang2024falcon}.
Among them, the next viewpoint is selected as the one offering the highest frontier voxels visibility and holding sufficient Fisher information queried from the OFIF (Sec.~\ref{subsec:ofif}).
This selection process ensures the chosen viewpoint contributes to exploration coverage, while avoiding texture-less regions, thereby maintaining reliable vehicle localization.


\section{Perception-Aware Action Execution}
\label{sec:local_planning}
Once the next viewpoint is selected, a perception-aware local planner should generate a smooth trajectory to reach the target viewpoint. 
Beyond fundamental requirements such as collision avoidance and smoothness, the planner aims to minimize drift in visual state estimation during navigation.
To achieve this, we introduce an Online Fisher Information Field (Sec.~\ref{subsec:ofif}) that quantifies localization information, on which perception-aware trajectories are optimized for execution (Sec.~\ref{subsec:traj_planning}).

\subsection{The Online Fisher Information Field}
\label{subsec:ofif}

An Online Fisher Information Field (OFIF) is an online updated and differentiable volumetric map representation that provides constant-time queries on the Fisher information of visual localization for real-time exploration planning.
Specifically, the field evaluates how well the 6-DoF camera pose can be estimated from visual observations of nearby features that are already present in the map.
The proposed OFIF is inspired by the previous work \cite{zhang2020fisher}, which focused on offline map construction for known environments and is not suitable for exploration tasks in unknown environments. 
The core idea of OFIF is straightforward: it stores a rotation-independent component of the Fisher information in each voxel, which can later be reused to recover the Fisher information when queried by a camera pose located in the voxel.
Formally speaking, the total Fisher information matrix (FIM) at a camera pose $\matr{T}_{\text{wc}} = (\matr{R}_{\text{wc}}, \vect{t}_{\text{wc}})$ is aimed to be decomposed as
\begin{equation}
    \label{eqn:fim_separation}
    \mathcal{I}_{\matr{T}_{\text{wc}}} \approx g(\sum_{i\in F}h(\vect{t}_{\text{wc}}, \vect{p}_{i}^\text{w}), \matr{R}_{\text{wc}}),
\end{equation}
where $\sum_{i\in F}h(\vect{t}_{\text{wc}}, \vect{p}_{i}^\text{w})$ is the \textit{positional factor} stored in the voxel, $g(\cdot, \cdot)$ is the \textit{query function} that recovers the FIM from camera rotation $\matr{R}_{\text{wc}}$, and $\{\vect{p}_i\}_{i\in F}$ denotes the set of observable and matchable feature points at $\vect{t}_{\text{wc}}$ that are non-occluded and within sensing range. In fact, the total FIM is computed as 
\begin{equation}
    \label{eqn:sum_fim}
    \mathcal{I}_{\matr{T}_{\text{wc}}} = \sum_{i\in F} v(\matr{T}_{\text{wc}}, \vect{p}_{i}^\text{w}) \mathcal{I}(\vect{p}_i),
\end{equation}
where $\mathcal{I}(\vect{p}_i)$ denotes the FIM for the observation of the $i$-th feature at the camera pose $\matr{T}_{\text{wc}}$, and $v(\cdot, \cdot)$ represents the visibility function indicating whether a feature lies within the camera's FoV.
The following subsections analyze these two terms and detail how (\ref{eqn:fim_separation}) is achieved.

\subsubsection{Fisher Information Matrix for a Single Feature}
\label{subsubsec:fim}
For a general parameter estimation problem, the Fisher Information Matrix (FIM) encapsulates the amount of information that an observable random variable $\vect{x}$ carries about a set of unknown parameters $\boldsymbol{\theta}$.
The observed Fisher Information Matrix \cite{efron1978assessing} can be expressed as \cite{zhang2020fisher}
\begin{equation}
    \label{eqn:fisher_def2}
    \mathcal{I}_{\boldsymbol{\theta}}(\vect{x})=\frac{1}{\sigma^{2}}(\matr{J}_{\boldsymbol{\theta}})^T \matr{J}_{\boldsymbol{\theta}}, \quad \text { where } \matr{J}_{\boldsymbol{\theta}}=\frac{\partial \vect{x}}{\partial \boldsymbol{\theta}},
\end{equation}
where $\sigma^2$ is the constant noise variance and can be neglected in subsequent analysis without loss of generality.
The inverse of the FIM defines the Cramér-Rao lower bound \cite{radhakrishna1945information}, which sets a lower limit on the covariance of any unbiased estimator.
Consequently, the FIM reflects the certainty of parameter estimation given the measurements \cite{barfoot2024state}.
Particularly, the equation of the Cramér-Rao lower bound can be achieved in maximum likelihood estimators \cite{daniels1961asymptotic}.

In our context, we consider the unknown parameters as the camera pose $\matr{T}_{\text{wc}}$ and the observation of feature point $\vect{p}_i$ as a bearing vector $\vect{f}_{i}$.
The measurement model, the Jacobian, and the FIM for a single feature observation are given by
\begin{align}
    \vect{f}_{i}=\frac{\vect{p}_{i}^\text{c}}{\left\|\vect{p}_{i}^\text{c}\right\|_{2}}&\triangleq\frac{1}{n_{i}} \vect{p}_{i}^\text{c}, \text{ where }\vect{p}_{i}^\text{c}=\matr{T}_{\text{cw}} \vect{p}_{i}^\text{w}, \ n_i = \left\|\vect{p}_{i}^\text{c}\right\|_{2} \label{eqn:bearing}\\
    \matr{J}_{i}&=\frac{\partial \vect{f}_{i}}{\partial \matr{T}_{\text{wc}}}=\frac{\partial \vect{f}_{i}}{\partial \vect{p}_{i}^\text{c}} \frac{\partial \vect{p}_{i}^\text{c}}{\partial \matr{T}_{\text{wc}}}, \ \mathcal{I}(\vect{p}_i) = \matr{J}_{i}^T \matr{J}_{i}. \label{eqn:jacobian_and_fim}
\end{align}
The second derivative ${\partial \vect{p}_{i}^\text{c}}/{\partial \matr{T}_{\text{wc}}}$ in $\matr{J}_{i}$ related to the rigid body transformation $\matr{T}_{\text{wc}}$ involves Lie algebra and a perturbation $\boldsymbol{\xi}\in\mathfrak{se}(3)$. Depending on the perturbation parameterization employed, the Jacobian $\matr{J}_{i}$ can have different expressions:

\textit{(i) Left perturbation:}
\begin{align}
    \frac{\partial \vect{p}_{i}^\text{c}}{\partial \matr{T}_{\text{wc}}} &= \frac{\partial(\exp (\boldsymbol{\xi}^{\wedge}) \matr{T}_{\text{wc}})^{-1} \vect{p}_{i}^\text{w}}{\partial \boldsymbol{\xi}} =\matr{R}_{\text{cw}}\left[-\matr{I}_{3}, (\vect{p}_{i}^\text{w})^{\wedge}\right], \\
    \matr{J}_{i}^L&=(\frac{1}{n_{i}} \matr{I}_{3}-\frac{1}{n_{i}^{3}} \vect{p}_{i}^\text{c}(\vect{p}_{i}^\text{c})^T) \matr{R}_{\text{cw}}\left[-\matr{I}_{3}, (\vect{p}_{i}^\text{w})^{\wedge}\right].
\end{align}

\textit{(ii) Right perturbation:}
\begin{align}
    \frac{\partial \vect{p}_{i}^\text{c}}{\partial \matr{T}_{\text{wc}}} &= \frac{\partial(\matr{T}_{\text{wc}} \exp (\boldsymbol{\xi}^{\wedge}) )^{-1} \vect{p}_{i}^\text{w}}{\partial \boldsymbol{\xi}} =\left[-\matr{I}_{3}, (\vect{p}_{i}^\text{c})^{\wedge}\right], \\
    \matr{J}_{i}^R&=(\frac{1}{n_{i}} \matr{I}_{3}-\frac{1}{n_{i}^{3}} \vect{p}_{i}^\text{c}(\vect{p}_{i}^\text{c})^T) \left[-\matr{I}_{3}, (\vect{p}_{i}^\text{c})^{\wedge}\right].  \label{eqn:right_perturb}
\end{align}

\begin{table}[t]
  \begin{center}
    \caption{T-optimality of FIM from Different Perturbations}
    \vspace{-0.1cm}
    \label{tab:fim_perturb}
    \begin{tabular}{ccccc}
      \hline\hline
      \multicolumn{2}{c}{\textbf{Feature Position}}           & \textbf{Camera Pos.}  & \textbf{Left Pert.}           & \textbf{Right Pert.} \\
      $\vect{p}^\text{c}$ & $\vect{p}^\text{w}$  & $\vect{t}_\text{wc}$     & $\tr(\mathcal{I}^L(\vect{p}))$ & $\tr(\mathcal{I}^R(\vect{p}))$\\
      \hline
      \multirow{2}{*}{(1, 0, 0)}  & (1, 0, 0) & (0, 0, 0) & 4.0 & 4.0 \\
                                           & (1, 0, 10) & (0, 0, 10) & 104.0 & 4.0 \\
      \hline
      \multirow{2}{*}{(2, 0, 0)}  & (2, 0, 0) & (0, 0, 0) & 2.5 & 2.5 \\
                                           & (2, 0, 10) & (0, 0, 10) & 27.5 & 2.5 \\
      \hline\hline
    \end{tabular}
  \end{center}
  \vspace{-0.1cm}
  \footnotesize{$^*$ The camera rotation is set to identity due to the rotation-invariant T-optimality of FIM from both left and right perturbations.}
  \vspace{-0.8cm}
\end{table}

The previous work \cite{zhang2020fisher} adopted the left perturbation Jacobian to compute the FIM due to its appealing property that the resulting FIM $\mathcal{I}^L(\vect{p}_i)$ is invariant to camera rotation $\matr{R}_{\text{wc}}$.
However, this invariance comes at a cost that $\mathcal{I}^L(\vect{p}_i)$ scales with the feature's world-frame coordinate $\vect{p}_i^\text{w}$ and the camera position $\vect{t}_{\text{wc}}$.
As shown in Table~\ref{tab:fim_perturb}, feature points with identical camera-frame coordinates yield significantly higher information values when farther from the world origin, making comparisons across positions inconsistent.
Intuitively, this occurs because a small world-frame rotation in left perturbation on the camera pose can have a large effect on the observation of a feature point distant from the world origin.
In contrast, the right perturbation Jacobian yields an FIM $\mathcal{I}^R(\vect{p}_i)$ that depends only on the camera-frame coordinate of the features $\vect{p}_i^{c}$, hence giving comparable information values at different world-frame positions.
Although the full FIM $\mathcal{I}^R(\vect{p}_i)$ loses the property of rotation invariance, its T-optimality fortunately remains invariant to the camera rotation, resulting in an elegant closed-form expression,
\begin{equation}
  \label{eqn:fim_closed_form}
  \tr(\mathcal{I}^R(\vect{p}_i)) = \frac{2}{n_i^2} + 2, \quad n_i = \left\|\vect{p}_{i}^\text{c}\right\|_{2}.
\end{equation}
The derivation of the closed-form expression is provided in Appendix \ref{apx:fim_closed_form}. 
By leveraging T-optimality with right perturbation FIMs, the resulting information remains rotation invariant while eliminating dependence on world-frame coordinates.


\subsubsection{Visibility Model Separation}
\label{subsec:visib_model}
\begin{figure}[t]
	\centering
  \includegraphics[width=\columnwidth]{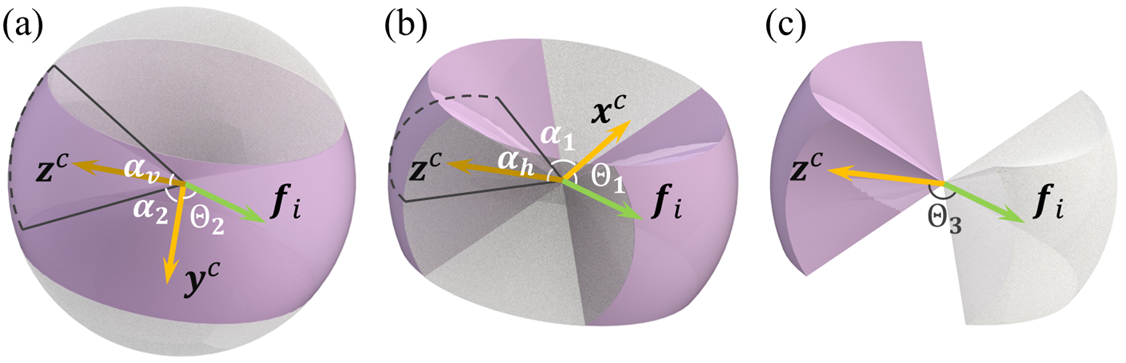}  
  \vspace{-0.6cm}
  \caption{\label{fig:visib_model} The figure shows a visualization of the frustum visibility model, where the grey volumes indicate the deducted components.}
  \vspace{-0.8cm}
\end{figure}

The exact visibility function $v(\matr{T}_{\text{wc}}, \vect{p}_{i}^\text{w})$ in (\ref{eqn:sum_fim}) is a binary indicator that equals $1$ if feature point $\vect{p}_{i}^\text{w}$ lies within the camera's FoV at pose $\matr{T}_{\text{wc}}$, and $0$ otherwise.
In practice, this discontinuous function is approximated by a smooth visibility model.
As the simple cone model from the prior work \cite{zhang2020fisher} lacks expressiveness for asymmetric FoVs, we instead adopt a more general deductive frustum visibility model \cite{chen2024apace} as shown in Fig.~\ref{fig:visib_model}, defined as
\begin{align}
  v'&(\matr{T}_{\text{wc}}, \vect{p}_{i}^\text{w}) = v_2(\theta_2) \cdot v_1(\theta_1) \cdot v_3(\theta_3), \label{eqn:visibility}\\
  v_1(\theta_1) = &(1 + e^{-k(\sin\theta_1 - \sin\alpha_1)})^{-1},\;\sin\theta_1 = |\vect{x}^\text{c} \times \vect{f}_i|,\\
  v_2(\theta_2) = &(1 + e^{-k(\sin\theta_2 - \sin\alpha_2)})^{-1},\;\sin\theta_2 = |\vect{y}^\text{c} \times \vect{f}_i|,\\
  v_3(\theta_3) = &(1 + e^{-k(\cos\theta_3 - \cos\alpha_3)})^{-1},\;\cos\theta_3 = \vect{z}^\text{c} \cdot \vect{f}_i,
\end{align}
where $\alpha_1 = (\pi - \alpha_h) / 2, \alpha_2 = (\pi - \alpha_v) / 2, \alpha_3 = \pi / 2$,  $(\alpha_h, \alpha_v)$ are the horizontal and vertical FoVs. $\theta_1, \theta_2, \theta_3$ are the angles between $\vect{f}_i$ in (\ref{eqn:bearing}) and the camera axes $\vect{x}^\text{c}, \vect{y}^\text{c}, \vect{z}^\text{c}$.

To decompose the frustum visibility model $v'(\matr{T}_{\text{wc}}, \vect{p}_{i}^\text{w})$ into \textit{positional visibility} $\vect{v}^p(\vect{t}_{\text{wc}}, \vect{p}_{i}^\text{w})$ and \textit{rotational visibility} $\vect{v}^r(\matr{R}_{\text{wc}})$ as required by (\ref{eqn:fim_separation}), we employ Gaussian process regression (GPR) with quaternion inputs.
Given a feature point $\vect{p}^\text{w}_i$ and camera position $\vect{t}_\text{wc}$, we sample $N_s$ rotations $\{\vect{q}_i\}_{i=1}^{N_s}$ and compute the corresponding visibilities $\matr{v}^s = [v_1, v_2, \dots, v_{N_s}]^T$ using (\ref{eqn:visibility}).
Using these samples, the GPR model allows us to approximate the visibility for any camera rotation quaternion $\vect{q}_\text{wc}$ associated with $\matr{R}_{\text{wc}}$ as:
\begin{align}
    v(\vect{q}_\text{wc}) = \matr{K}_\vect{q} &\matr{K}^{-1} \matr{v}^s, \; \matr{K}_\vect{q} \in \mathbb{R}^{1\times N_s}, \; \matr{K} \in \mathbb{R}^{N_s \times N_s}, \\
    \matr{K}_\vect{q}(1,i) &= k(\vect{q}_\text{wc}, \vect{q}_i), \; \matr{K}(i,j) = k(\vect{q}_i, \vect{q}_j),
\end{align}
where the kernel function $k(\vect{q}_1, \vect{q}_2) = \sigma_\text{gpr}^2\exp (-||\vect{q}_1 - \vect{q}_2||^2 / (2l_\text{gpr}^2))$ is a standard squared exponential kernel adapted for quaternions \cite{lang2014gaussian}, whose hyperparameters $\sigma_\text{gpr}$ and $l_\text{gpr}$ are learned during training.
By using the same set of rotation samples $\{\vect{q}_i\}_{i=1}^{N_s}$ for all feature points, this formulation naturally achieves the desired decomposition, where the rotational visibility $\vect{v}^r(\matr{R}_{\text{wc}})^T = \matr{K}_\vect{q}$ depends only on the camera rotation, and the positional visibility $\vect{v}^p(\vect{t}_{\text{wc}}, \vect{p}_{i}^\text{w}) = \matr{K}^{-1} \matr{v}^s$ depends only on the camera and feature point positions.


\subsubsection{OFIF Incremental Construction}
\label{subsec:ofif_update}
From Sec.~\ref{subsubsec:fim} and \ref{subsec:visib_model}, we reformulate (\ref{eqn:sum_fim}) as
\begin{align}
  \tr(\mathcal{I}_{\matr{T}_{\text{wc}}}) &= \sum_{i\in F} v(\matr{T}_{\text{wc}}, \vect{p}_{i}^\text{w}) \tr(\mathcal{I}^R(\vect{p}_i)) \\
  &\approx \sum_{i\in F} \vect{v}^r(\matr{R}_{\text{wc}})^T \cdot \vect{v}^p(\vect{t}_{\text{wc}}, \vect{p}_{i}^\text{w}) \tr(\mathcal{I}^R(\vect{p}_i)) \\
  &= \vect{v}^r(\matr{R}_{\text{wc}})^T \sum_{i\in F} \vect{v}^p(\vect{t}_{\text{wc}}, \vect{p}_{i}^\text{w}) \tr(\mathcal{I}^R(\vect{p}_i)) 
\end{align}
This allows us to achieve our goal (\ref{eqn:fim_separation}) through appropriate definitions of the positional factor and query function as
\begin{align}
  \sum_{i\in F} h(\vect{t}_{\text{wc}}, \vect{p}_{i}^\text{w}) &\triangleq \sum_{i\in F} \vect{v}^p(\vect{t}_{\text{wc}}, \vect{p}_{i}^\text{w}) \tr(\mathcal{I}^R(\vect{p}_i)) \label{eqn:pos_factor} \\
  g(\sum_{i\in F} h(\vect{t}_{\text{wc}}, \vect{p}_{i}^\text{w}), \matr{R}_{\text{wc}}) &\triangleq \vect{v}^r(\matr{R}_{\text{wc}})^T \sum_{i\in F} h(\vect{t}_{\text{wc}}, \vect{p}_{i}^\text{w}) \label{eqn:query_func}
\end{align}
The positional factor $\sum_{i\in F} h(\vect{t}_{\text{wc}}, \vect{p}_{i}^\text{w})$ is stored in the voxel corresponding to $\vect{t}_{\text{wc}}$ and supports incremental updates as new feature points are observed. 
This approach allows constant-time queries of the Fisher information at any camera pose $\matr{T}' = (\matr{R}', \vect{t}')$ which multiplies the rotational visibility of $\matr{R}'$ by the precomputed positional factor retrieved from the voxel containing $\vect{t}'$.
The differentiable formulation (\ref{eqn:query_func}) is also suitable for gradient-based optimization in motion planning.

\begin{algorithm}[t]
  \caption{OFIF Incremental Construction}
  \label{alg:ofif_update}
  \DontPrintSemicolon
  \KwData{OFIF map $\mathcal{F}$, feature point cloud $\mathcal{P}$, occupancy map $\mathcal{O}$}
  \KwResult{updated OFIF map $\mathcal{F}$}
  \SetKwFunction{FMain}{updatePosFactor}
  \BlankLine

  \While{\textnormal{$P_\text{new} \leftarrow$ featureObservation()}}{

    \textit{// Update new free voxels with previous features} \label{algln:step1_1}

    $V_\text{new}$ $\leftarrow$ $\mathcal{F}.$getNewFreeVoxels($\mathcal{O}$)

    \ForEach{\textnormal{$\vect{v}_i \in V_\text{new}$}}{
      $P_i \leftarrow$ $\mathcal{P}$.getNearbyFeatures($\vect{v}_i, r_\text{s}$)

      \FMain($\{\vect{v}_i\}$, $P_i$)
    } \label{algln:step1_2}
    
    \textit{// Update new features into all free voxels} \label{algln:step2_1}

    \ForEach{\textnormal{$\vect{p}_i \in P_\text{new}$}}{
      $V_i$ $\leftarrow$ $\mathcal{F}$.getNearbyFreeVoxels($\vect{p}_i$, $r_\text{s}$)

      \FMain($V_i$, $\{\vect{p}_i\}$)
    }

    $\mathcal{P}$.add($P_\text{new}$) \label{algln:step2_2}
  }

  \SetKwProg{Fn}{Function}{:}{}
  \Fn{\FMain{$V$, $P$}}{
    \ForEach{$\vect{v}_i \in V$ and $\vect{p}_j \in P$}{
      \If{\textnormal{$\mathcal{O}$.checkNotOccluded($\vect{v}_i$, $\vect{p}_j$)}}{
        $\vect{v}_i$.info += calcPosFactor($\vect{v}_i$, $\vect{p}_j$) \Comment*[r]{\textnormal{(\ref{eqn:pos_factor})}}
      }
    }
  }
\end{algorithm}
\setlength{\textfloatsep}{0pt}

\begin{figure}[t]
	\centering
  \includegraphics[width=\columnwidth]{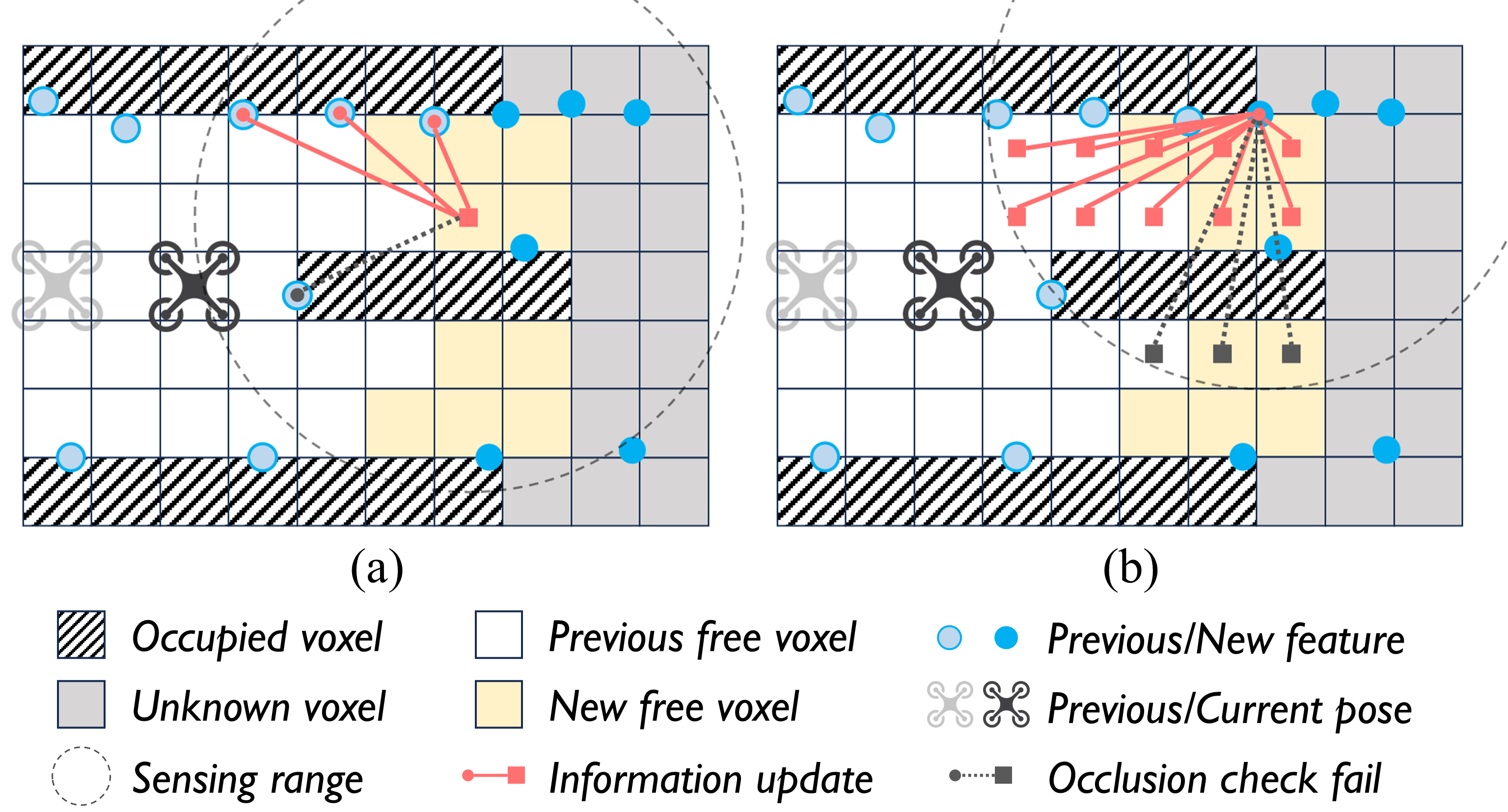}  
  \vspace{-0.6cm}
  \caption{\label{fig:ofif_update} The two-step update of the OFIF incremental construction. (a) In step 1, each new free OFIF voxel is updated with previous localizable features within sensing range. (b) In step 2, each new localizable feature is updated into all nearby free OFIF voxels within sensing range.}
\end{figure}

The OFIF is designed to store the positional factors within its voxels for efficient retrieval during information queries.
The challenges of online OFIF construction are twofold: 1) identifying observable and matchable feature points that contribute to localization during navigation, which are named \textit{localizable} features, and 2) efficiently updating the positional factors in the related voxels when new localizable features are detected during exploration. 
For the first challenge, we track feature point observations in the visual odometry.
A feature point is considered localizable only if it is observed and matched consistently across several consecutive frames.
These localizable features are stably tracked with low position uncertainty, which are considered as deterministic points for OFIF updates.
We also exclude features outside the exploration bounding box or beyond the feature observation range $r_\text{o}$, which is set larger than the sensing range $r_\text{s}$ for the standard mappings.
For the second challenge, whenever new localizable features $P_\text{new}$ are received, an incremental update of the related voxels is performed, as summarized in Alg.~\ref{alg:ofif_update} and illustrated in Fig.~\ref{fig:ofif_update}.
This process first updates the new free OFIF voxels with the previous localizable features within the sensing range $r_\text{s}$.
Then it iterates through the new localizable features $P_\text{new}$ and updates the surrounding free OFIF voxels, including both previous and new ones.
When updating a voxel $\vect{v}_i$ with a feature point $\vect{p}_j$, an occlusion check is performed to confirm that $\vect{p}_j$ is observable from $\vect{v}_i$.
If the feature is not occluded, the positional factor stored in voxel $\vect{v}_i$ is additively updated according to (\ref{eqn:pos_factor}).
This two-step update process ensures that the OFIF voxels in free regions are properly maintained with the latest localizable feature observations, avoiding duplicated or missing information updates during exploration.

The OFIF serves as a computationally efficient localizability measure for exploration planning.
Its construction adopts several approximations, including the trace-based scalarization of the FIM, the separable visibility model, and the bounded local incremental updates, trading off FIM accuracy for memory efficiency and real-time computation.
Despite this approximation, the field preserves sufficient visual localization information required by the planner in the following Sec.~\ref{subsec:traj_planning} to generate visual-information-rich trajectories.


\subsection{Perception-aware Trajectory Planning}
\label{subsec:traj_planning}
To incorporate perception information from the OFIF into trajectory planning, we first define an information potential cost function that penalizes a camera pose $\matr{T}$ with low Fisher information $\tr(\mathcal{I}_\matr{T}) \triangleq x$,
\begin{equation}
    \label{eqn:info_potential}
    f_\text{info}(\matr{T}) = 
    \begin{cases}
        0,  & \epsilon_\text{info} < x\\
        k_1(x - \epsilon_\text{info})^2, & 0 \leq x \leq \epsilon_\text{info},\\
        k_2 x + b_\text{pot}, & x < 0\\
    \end{cases}
\end{equation}
where the sufficient information threshold $\epsilon_\text{info}$ is determined as the cumulative information from $N_\text{pot}$ randomly sampled feature points within the sensing range $r_\text{s}$.
After specifying the parameter $b_\text{pot}$, the potential cost at $0$ information, $k_1$ and $k_2$ are computed to ensure potential function continuity.

\begin{figure}[t]
	\centering
  \includegraphics[width=0.9\columnwidth]{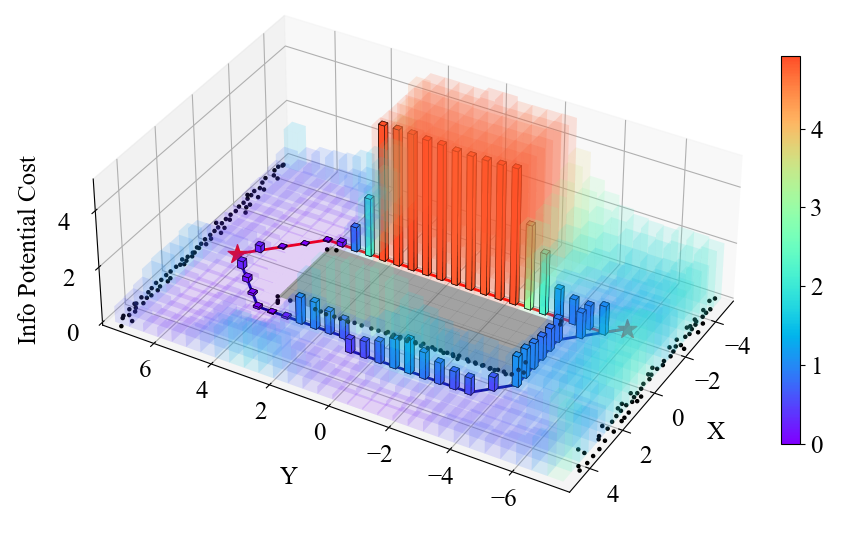}  
  \vspace{-0.6cm}
  \caption{\label{fig:infoA*} A 2D comparison between info-A* search and standard A* search algorithms for trajectory planning.
  The central rectangle represents an obstacle and the black dots indicate localizable features. 
  The transparent rainbow-colored histogram depicts the information potential cost at each voxel.
  The info-A* path (dark blue) favors texture-rich regions with low information panelties, whereas the standard A* search path (red) follows the shortest route, traversing regions with few localizable features.}
\end{figure}

We decompose the trajectory planning problem into independent position and yaw optimization, leveraging the differential flatness property of quadrotor dynamics \cite{mellinger2011minimum}.
For the position trajectory, we strategically incorporates perception information in the initial path search since the trajectory optimization can only perform local refinements to an initial path, as illustrated in Fig.~\ref{fig:infoA*}.
The search provides an informative initial path, deprioritizing texture-less regions where later optimization would have a limited effect.
Such an informative initial path is searched within the free space of the occupancy grid map using an info-A* search, which is a variation of the classical A* algorithm.
The info-A* evaluates each node $\vect{n}_i$ using both a geometric cost and an information potential cost that is calculated at pose $(\vect{n}_i, \psi_i^*)$ ,where $\psi_i^*$ is the most informative yaw direction obtained by sampling.
For a path $\pi = \{\vect{n}_i\}_i$, the path cost $g(\pi)$ incorporates a weighted cumulative information penalties, i.e., $g(\pi) = \text{dist}(\pi) + w_\text{info} \sum_{i} f_\text{info}(\vect{n}_i, \psi_i^*)$.
The heuristic function remains the Euclidean distance to the goal to ensure admissibility.
By using info-A* search, the planner can prioritize paths that traverse more informative regions while maintaining a comparable travel distance.
Based on the obtained initial path, we optimize a minimum-time B-spline trajectory subject to standard constraints, including smoothness, collision avoidance, and dynamical feasibility, to produce the final executable position trajectory.

For the yaw trajectory, we optimize the yaw trajectory for informative viewing directions to enhance localization stability during navigation.
To accelerate optimization convergence and avoid local minima, we first generate a yaw primitives path that serves as the initial input for subsequent yaw trajectory optimization.
This path generation is modeled as a graph search problem that minimizes the accumulated information potential cost at knot points along the position trajectory.
For each knot point $\{\vect{P}_i\}_{i=0}^m$, we uniformly sample a set of yaw candidates $L_i = \{\psi_{i,0}, \psi_{i,1}, \cdots, \psi_{i,l}\}$, each of which serves as a graph node.
To enforce trajectory continuity, the first and last sets $L_0=\{\psi_0\}$ and $L_m=\{\psi_m\}$ are fixed to the start and goal yaw angles, respectively.
Edges are established between nodes from consecutive knot points $(\psi_{i,j_1}, \psi_{i+1,j_2})$, with the associated cost defined as
\begin{equation}
f_\text{info}(\vect{P}_{i+1}, \psi_{i+1,j_2}) + w_\text{ys}\cdot\text{diff}(\psi_{i,j_1}, \psi_{i+1,j_2}),
\end{equation}
where $w_\text{ys}$ is the yaw smoothness weight, and $\text{diff}(\cdot, \cdot)$ denotes the smallest rotation angle between two yaws.
The yaw primitives path is then searched as the minimum-cost path from node $\psi_0$ to $\psi_m$, which can be efficiently solved using Dijkstra's algorithm.

The yaw primitives path serves as the initial solution for subsequent B-spline trajectory optimization. Thanks to our differentiable formulation of perception information, the perception constraint can be seamlessly integrated into the optimization process.
For a B-spline yaw trajectory $\psi(t)$ with control points $\mathrm{\Xi}$, the yaw optimization is formulated as
\begin{equation}
  \argmin_{\mathrm{\Xi}} \int_{t_0}^{t_{m}} \dddot{\psi}(t)^2 dt + \lambda_p \sum_{i=1}^{m-1} f_\text{info}(\vect{P}_i, \psi(t_i)),
\end{equation}
where the first term evaluates the trajectory jerk as smoothness cost, and the second term for perception is the accumulated information potential costs at the knot points, which are computed using (\ref{eqn:query_func}) and (\ref{eqn:info_potential}). 
The weighting parameter $\lambda_p$ balances the trajectory smoothness and perception informativeness.
The optimized yaw trajectory results in smooth motion and informative perception throughout the navigation.

\section{Benchmark Experiments}
\label{sec:benchmark}
This section presents a series of benchmark experiments to comprehensively evaluate the proposed active SLAM framework RAVEN in comparison with state-of-the-art methods \cite{zhang2024falcon,carrillo2018autonomous,lee2021real,placed2022explorb}.
All simulation experiments in Sec.~\ref{sec:benchmark}~-~\ref{sec:ablation} are conducted on a computer with an Intel Ultra 9 275HX CPU, 96GB RAM, and an NVIDIA RTX 5080 Laptop GPU.

\subsection{Implementation Details}
\label{subsec:implement_detail}

\subsubsection{Simulator Setup}
Building upon the open-source exploration planner evaluation environment (EPEE) \cite{zhang2024falcon}, we upgrade this software-in-the-loop simulator into the EPEE+ to facilitate active SLAM algorithms. There are three critical enhancements: (1) multi-camera color images real-time rendering, (2) simulated IMU measurements with configurable noise models, and (3) visual SLAM integration.
These additions enable a realistic simulation of visual state estimation with drift arising from both visual tracking inaccuracies and IMU noises.
As a result, the lightweight simulator EPEE+ offers a photorealistic simulation environment tailored for active SLAM research, promoting fair and comprehensive benchmark and reproducible research.

\subsubsection{Benchmark Candidates and Settings}
The benchmark comparison evaluates the proposed framework RAVEN against a series of state-of-the-art exploration planners: SRE\cite{carrillo2018autonomous}, REAL\cite{lee2021real}, ExplORB\cite{placed2022explorb}, and FALCON\cite{zhang2024falcon}, listed in chronological order of publication.
SRE employs Shannon-Rényi Entropy (SRE) to automatically balance exploration and exploitation trade-offs.
REAL performs active loop closures when the current and historical poses hold far enough geodesic distance along the trajectory and close enough Euclidean distance.
ExplORB adopts a greedy frontier selection strategy based on the D-optimality of the hallucinate pose graph with potential loop closures are established as edges.
FALCON is a recent classic exploration planner that solely optimizes exploration efficiency under a perfect localization assumption.
The proposed method RAVEN is evaluated under three $\lambda$ values of $0.0, 0.5, 1.0$, which represent prioritization between exploration efficiency ($\lambda \to 0$) and localization accuracy ($\lambda \to 1$).
As no official implementation is available, SRE and REAL were reimplemented according to their manuscripts, while ExplORB and FALCON were adapted to the EPEE+ simulator using their open-source code. 
The parameter settings for RAVEN and those for all methods are summarized in Table~\ref{tab:params}.
For fair comparison, all methods utilize VINS\cite{qin2018vins} as SLAM module with loop closure detection\cite{xu2024d, xu2022omni}.

\begin{table}[t]
  \centering
	\caption{Parameter Settings}
  \vspace{-0.2cm}
	\begin{tabular}{p{3.4cm}M{0.6cm}M{1.4cm}M{1.8cm}}
		\toprule\toprule
		\textbf{Parameter} & \textbf{Sec.} & \textbf{Notation} & \textbf{Value} \\
		\midrule
		CPG node interval & \ref{subsec:con_pose_graph} & $d_\text{cpg}$ & $1.0$~m\\
		Top base tours & \ref{subsec:global_tour_cand} & $N_\text{cp}$ & $5$ \\
		Success loop closure thresh. & \ref{subsec:loc_util} & $\delta_x, \delta_y, \delta_z, \delta_\psi$ & $3$~m, $0.5$~rad\\
    GPR dimension & \ref{subsec:visib_model} & $N_s$ & $360$ \\
		Feature observation range & \ref{subsec:ofif_update} & $r_\text{o}$ & $8.0$~m\\
		Sufficient features & \ref{subsec:traj_planning} & $N_\text{pot}$ & $50$ \\
		Potential cost at zero info. & \ref{subsec:traj_planning} & $b_\text{pot}$ & $5.0$ \\
    Occupancy grid resolution & - & - & $0.1$~m \\
    OFIF resolution & - & - & $0.5$~m \\
    \cmidrule{1-4}
		\multicolumn{2}{l}{Max linear velocity} & $v_m$ & $1.0$~m$\slash$s\\
		\multicolumn{2}{l}{Max angular velocity} & $\dot{\psi}_m$ & $0.8$~rad$\slash$s\\
		\multicolumn{2}{l}{Sensing range} & $r_\text{s}$ & $5.0$~m \\
		\multicolumn{2}{l}{Simulated sensor FoV} & $(\alpha_h, \alpha_v)$ & $(90, 73)$~deg \\
		\multicolumn{3}{l}{Simulated IMU accelerometer noise magnitude} & $0.5$~m$\slash$s$^2$\\
		\multicolumn{3}{l}{Simulated IMU gyroscope noise magnitude} & $0.05$~rad$\slash$s \\
		\toprule\toprule
	\end{tabular}
	\label{tab:params}
	\vspace{-0.6cm}
\end{table}

\begin{figure}[t]
  \centering
  \captionsetup[subfigure]{justification=centering}
  \subfloat[Corridor\\$17\times17\times2$ m$^3$]{\includegraphics[width=0.25\columnwidth]{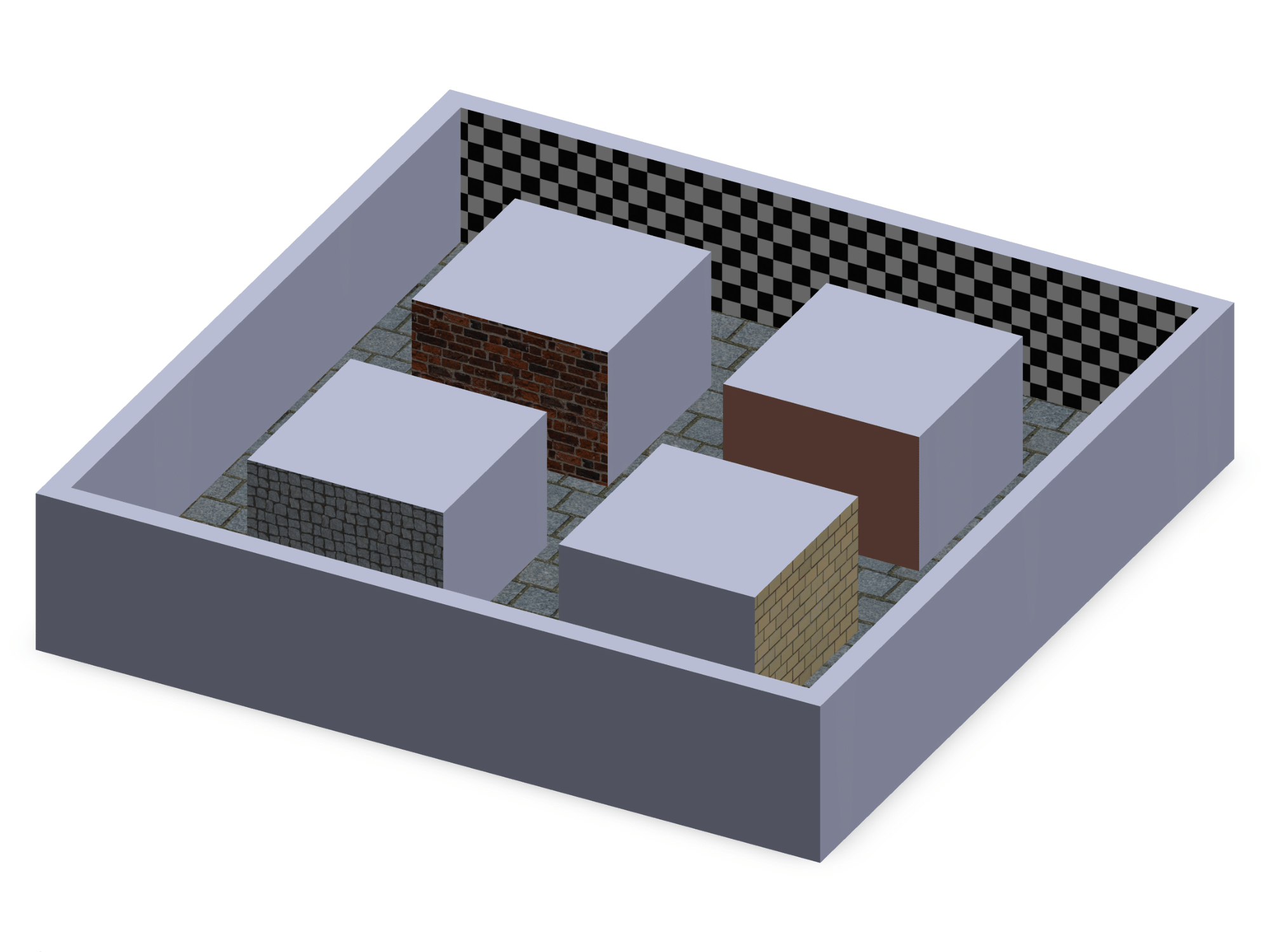}}
  \subfloat[Warehouse\\$21\times15\times2$ m$^3$]{\includegraphics[width=0.25\columnwidth]{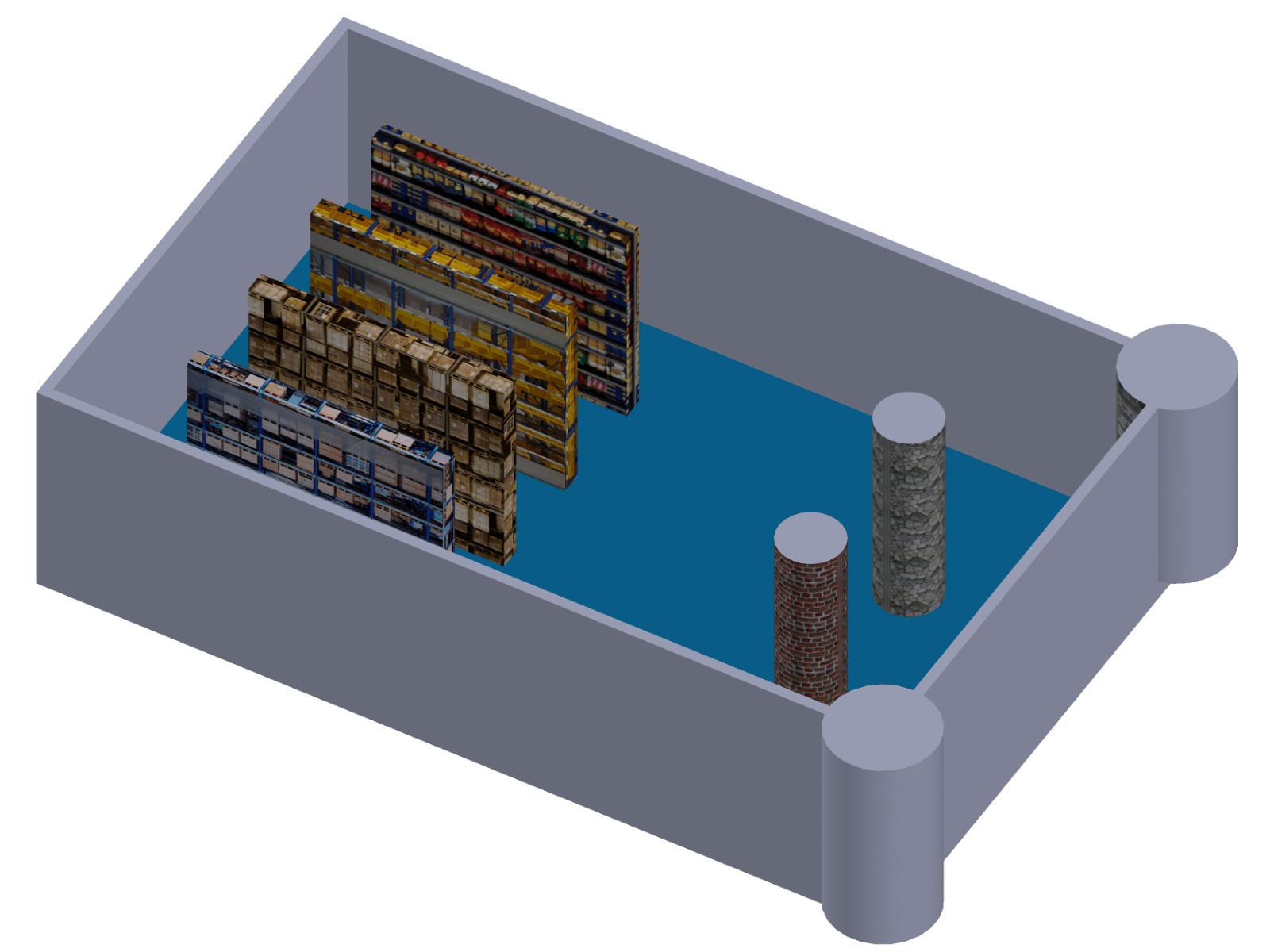}}
  \subfloat[Office\\$15\times10\times2$ m$^3$]{\includegraphics[width=0.25\columnwidth]{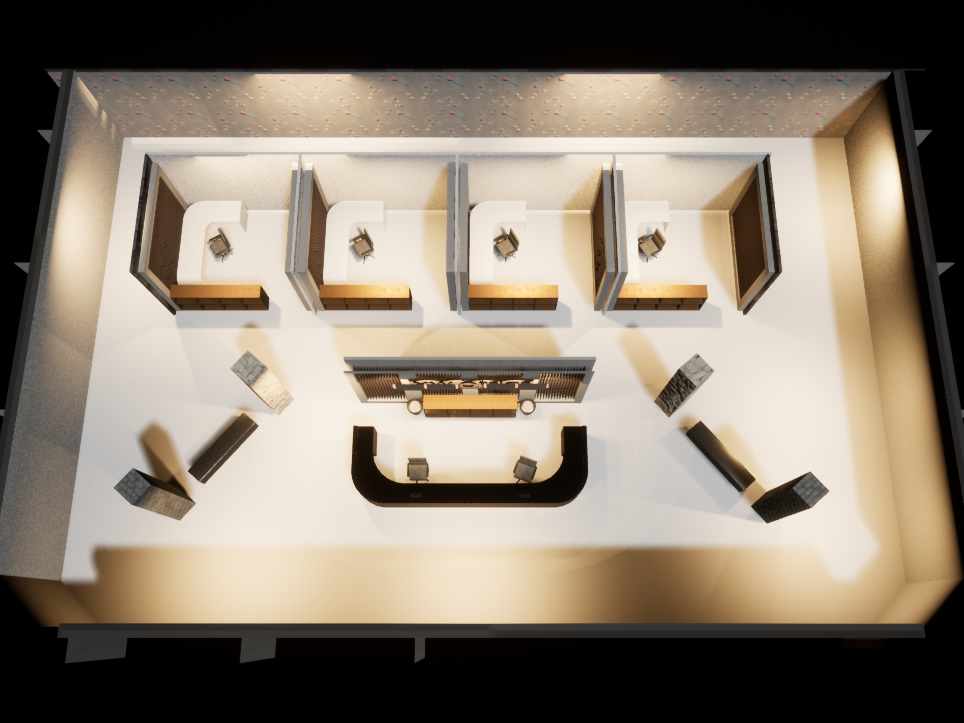}}
  \subfloat[Apartment\\$20\times 10\times 6$ m$^3$]{\includegraphics[width=0.25\columnwidth]{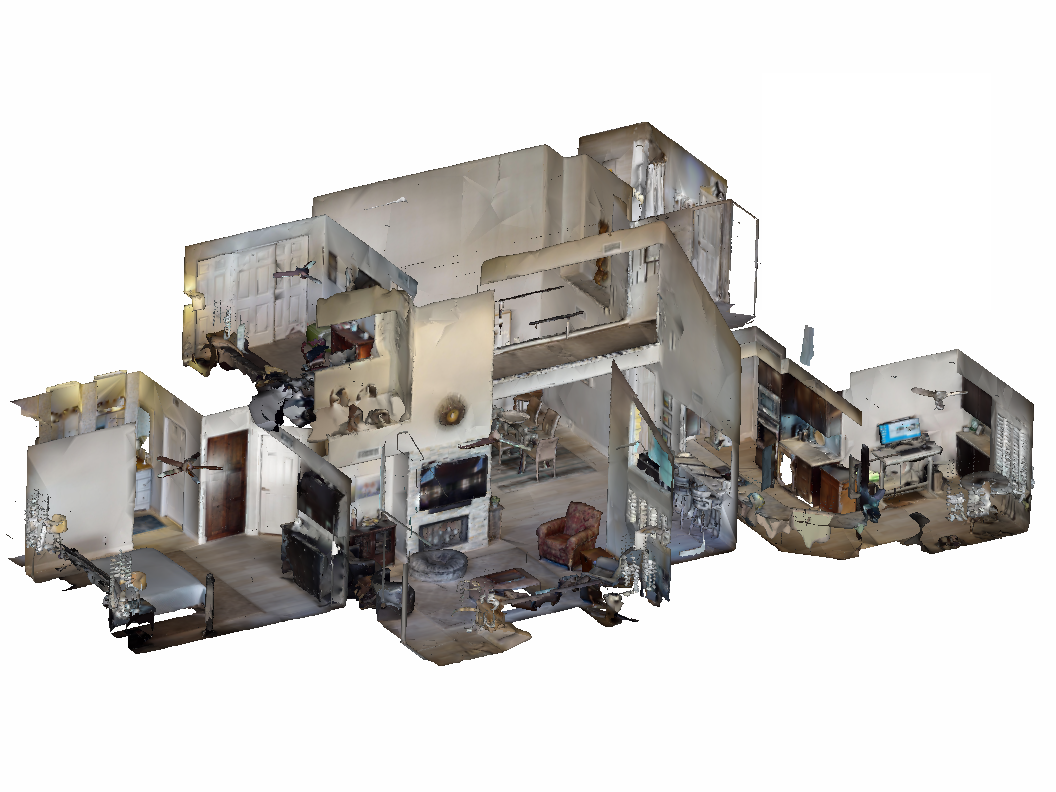}}
  \caption{\label{fig:exp_scenarios} An overview of the four testing scenarios for benchmark experiments.}
  \vspace{-2.0cm}
\end{figure}

\subsection{Simulation Benchmark Analysis}
\label{subsec:sim_benchmark}
\subsubsection{Evaluation Metrics}
\label{subsubsec:metrics}
The evaluation metrics are designed to assess each method's performance with respect to exploration efficiency and localization accuracy. 
Specifically, the metrics include exploration time, absolute trajectory error (ATE), flight distance, and looped images count.
Exploration time and flight distance are computed as the total duration and distance traveled by the vehicle until exploration completion. 
The ATE is calculated as the root mean square error (RMSE) of the Euclidean distance between the estimated trajectory from the SLAM algorithm and the ground truth trajectory provided by the simulator.
The looped images count refers to the number of image pairs successfully matched by the SLAM module, serving as an indicator of loop closure frequency.

\begin{table*}[t]
  \centering
  \caption{Benchmark Result Statistics}
  \vspace{-0.2cm}
  \begin{tabular}{M{1.0cm}M{1.7cm}M{0.55cm}M{0.4cm}M{0.55cm}M{0.55cm}M{0.4cm}M{0.4cm}M{0.4cm}M{0.4cm}M{0.55cm}M{0.4cm}M{0.55cm}M{0.55cm}M{0.55cm}M{0.4cm}M{0.55cm}M{0.55cm}}
    \toprule\toprule
    \multirow{2}{*}{\textbf{Scene}} & \multirow{2}{*}{\textbf{Method}} & \multicolumn{4}{c}{\textbf{Exploration Time (s)}}    
                                                                        & \multicolumn{4}{c}{\textbf{ATE (m)}}          
                                                                        & \multicolumn{4}{c}{\textbf{Flight Distance (m)}}          
                                                                        & \multicolumn{4}{c}{\textbf{Looped Images Count}} \\
                                    &                                  & \textbf{Avg} & \textbf{Std} & \textbf{Max} & \textbf{Min}
                                                                        & \textbf{Avg} & \textbf{Std} & \textbf{Max} & \textbf{Min}
                                                                        & \textbf{Avg} & \textbf{Std} & \textbf{Max} & \textbf{Min}
                                                                        & \textbf{Avg} & \textbf{Std} & \textbf{Max} & \textbf{Min} \\
    \midrule
    \multirow{7}{*}{\begin{tabular}[c]{@{}c@{}}Corridor\end{tabular}} 
      & SRE\cite{carrillo2018autonomous} & 177.3 & 18.3 & 202.0 & 142.4 & 0.24 & 0.15 & 0.54 & 0.07 & 175.2 & 17.0 & 194.1 & 141.7 & 103.7 & 66.9 & 254.0 & 15.0 \\
      & REAL\cite{lee2021real}           & 153.6 & 23.0 & 202.3 & 126.7 & 0.20 & 0.07 & 0.35 & 0.10 & 156.9 & 20.6 & 192.4 & 130.2 & 118.3 & 84.2 & 343.0 & 52.0 \\
      & ExplORB\cite{placed2022explorb}  & 159.9 & 17.9 & 187.8 & 126.7 & 0.17 & 0.06 & 0.28 & 0.10 & 151.7 & 24.0 & 178.0 & 91.6 & 79.4 & 31.3 & 148.0 & 28.0 \\
      & FALCON\cite{zhang2024falcon}     & \textbf{74.8} & 6.8 & 80.9 & 62.0 & 0.43 & 0.09 & 0.56 & 0.26 & \textbf{78.1} & 5.8 & 89.0 & 67.6 & 16.8 & 22.5 & 68.0 & 0.0 \\
      & RAVEN - 0.0                      & 88.8 & 8.0 & 100.6 & 76.7 & 0.29 & 0.09 & 0.39 & 0.12 & 86.4 & 7.3 & 97.3 & 72.0 & 15.5 & 17.3 & 60.0 & 0.0 \\
      & RAVEN - 0.5                      & 109.8 & 7.6 & 123.0 & 100.0 & 0.13 & 0.03 & 0.19 & 0.08 & 99.4 & 8.2 & 115.1 & 86.4 & 87.2 & 34.6 & 166.0 & 39.0 \\
      & RAVEN - 1.0                      & 160.6 & 15.5 & 189.5 & 141.8 & \textbf{0.08} & 0.02 & 0.13 & 0.05 & 142.3 & 12.6 & 165.6 & 126.0 & \textbf{285.1} & 135.3 & 644.0 & 132.0 \\
      \midrule
    \multirow{7}{*}{\begin{tabular}[c]{@{}c@{}}Warehouse\end{tabular}} 
      & SRE\cite{carrillo2018autonomous} & 179.7 & 9.2 & 192.5 & 161.8 & 0.35 & 0.10 & 0.51 & 0.17 & 181.7 & 11.5 & 192.4 & 153.9 & 101.9 & 66.1 & 233.0 & 25.0 \\ 
      & Real\cite{lee2021real} & 138.8 & 13.4 & 158.4 & 118.5 & 0.38 & 0.14 & 0.69 & 0.21 & 139.5 & 12.8 & 159.0 & 119.5 & 74.1 & 30.8 & 122.0 & 17.0 \\ 
      & ExplORB\cite{placed2022explorb} & 158.0 & 10.8 & 175.1 & 141.2 & 0.36 & 0.14 & 0.56 & 0.16 & 147.6 & 5.6 & 161.6 & 140.4 & 57.5 & 40.6 & 151.0 & 14.0 \\ 
      & FALCON\cite{zhang2024falcon} & \textbf{87.7} & 8.6 & 98.2 & 75.2 & 0.55 & 0.12 & 0.74 & 0.32 & 90.7 & 12.3 & 109.5 & 71.8 & 11.3 & 8.6 & 31.0 & 0.0 \\ 
      & RAVEN - 0.0 & 89.2 & 8.2 & 98.9 & 77.1 & 0.36 & 0.13 & 0.54 & 0.17 & \textbf{82.8} & 7.2 & 92.2 & 70.1 & 17.4 & 14.6 & 42.0 & 0.0 \\ 
      & RAVEN - 0.5 & 107.4 & 5.8 & 116.0 & 95.3 & 0.17 & 0.06 & 0.25 & 0.07 & 96.1 & 7.0 & 105.6 & 86.4 & 65.2 & 26.0 & 110.0 & 18.0 \\ 
      & RAVEN - 1.0 & 136.2 & 13.6 & 161.7 & 119.1 & \textbf{0.09} & 0.03 & 0.17 & 0.04 & 123.5 & 12.6 & 146.4 & 110.8 & \textbf{150.6} & 44.8 & 227.0 & 66.0 \\
    \midrule
    \multirow{7}{*}{\begin{tabular}[c]{@{}c@{}}Office\end{tabular}} 
      & SRE\cite{carrillo2018autonomous} & 96.8 & 7.3 & 113.1 & 88.0 & 0.35 & 0.09 & 0.47 & 0.19 & 89.0 & 6.9 & 97.5 & 74.4 & 31.0 & 20.2 & 58.0 & 0.0 \\ 
      & Real\cite{lee2021real} & 125.8 & 10.7 & 137.9 & 105.0 & 0.20 & 0.04 & 0.30 & 0.15 & 113.7 & 9.8 & 127.1 & 93.0 & 88.0 & 30.7 & 138.0 & 30.0 \\ 
      & ExplORB\cite{placed2022explorb} & 92.0 & 12.2 & 116.6 & 78.8 & 0.28 & 0.07 & 0.36 & 0.14 & 78.6 & 10.6 & 101.7 & 65.5 & 18.1 & 8.8 & 34.0 & 5.0 \\ 
      & FALCON\cite{zhang2024falcon} & \textbf{64.6} & 7.0 & 76.8 & 54.8 & 0.49 & 0.09 & 0.69 & 0.37 & \textbf{61.0} & 6.0 & 71.3 & 54.2 & 12.6 & 12.7 & 36.0 & 0.0 \\ 
      & RAVEN - 0.0 & 76.3 & 9.0 & 96.4 & 63.0 & 0.26 & 0.06 & 0.38 & 0.18 & 61.3 & 7.2 & 78.5 & 52.2 & 11.7 & 16.7 & 53.0 & 0.0 \\ 
      & RAVEN - 0.5 & 95.2 & 7.0 & 105.9 & 84.9 & 0.14 & 0.05 & 0.20 & 0.06 & 75.1 & 6.8 & 90.9 & 67.6 & 105.3 & 29.2 & 149.0 & 64.0 \\ 
      & RAVEN - 1.0 & 120.4 & 11.1 & 142.4 & 105.4 & \textbf{0.08} & 0.04 & 0.16 & 0.03 & 93.7 & 10.1 & 115.9 & 80.4 & \textbf{204.3} & 61.3 & 300.0 & 91.0 \\
    \midrule
    \multirow{7}{*}{\begin{tabular}[c]{@{}c@{}}Apartment\end{tabular}} 
      & SRE\cite{carrillo2018autonomous} & 171.9 & 10.6 & 189.8 & 157.4 & 0.17 & 0.04 & 0.24 & 0.13 & 148.9 & 4.8 & 155.5 & 138.5 & 694.5 & 76.6 & 849.0 & 588.0 \\
      & Real\cite{lee2021real} & 178.1 & 12.2 & 202.2 & 163.2 & 0.23 & 0.08 & 0.35 & 0.11 & 163.0 & 10.3 & 175.2 & 143.8 & 778.7 & 109.9 & 961.0 & 621.0 \\
      & ExplORB\cite{placed2022explorb} & 148.8 & 6.9 & 161.5 & 139.2 & 0.22 & 0.07 & 0.30 & 0.10 & 147.9 & 4.9 & 156.2 & 138.5 & 451.6 & 172.7 & 795.0 & 206.0 \\
      & FALCON\cite{zhang2024falcon} & \textbf{106.5} & 15.6 & 128.5 & 84.5 & 0.21 & 0.06 & 0.31 & 0.13 & 82.3 & 5.7 & 89.7 & 71.2 & 328.7 & 89.8 & 488.0 & 130.0 \\
      & RAVEN - 0.0 & 115.7 & 10.5 & 139.0 & 105.1 & 0.20 & 0.05 & 0.26 & 0.11 & \textbf{80.4} & 5.6 & 90.3 & 70.7 & 271.1 & 122.0 & 461.0 & 121.0 \\
      & RAVEN - 0.5 & 136.9 & 7.3 & 151.0 & 129.2 & 0.13 & 0.02 & 0.15 & 0.10 & 104.7 & 6.9 & 112.8 & 89.6 & 675.8 & 57.9 & 758.0 & 588.0 \\
      & RAVEN - 1.0 & 211.9 & 21.7 & 251.8 & 172.9 & \textbf{0.08} & 0.02 & 0.10 & 0.05 & 131.8 & 11.0 & 157.1 & 118.6 & \textbf{904.0} & 419.0 & 1380.0 & 186.0 \\
    \toprule\toprule
  \end{tabular}
  \label{tab:result}
\vspace{-0.4cm}
\end{table*}

\begin{figure*}[t!]
  \centering
  \captionsetup[subfloat]{captionskip=2pt, farskip=0pt}
  \subfloat{\includegraphics[width=2.0\columnwidth]{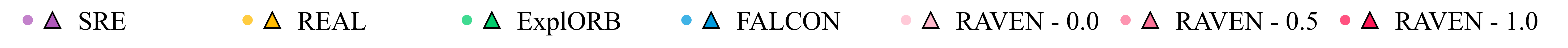}}\\
  \subfloat[Corridor]{\includegraphics[width=0.5\columnwidth]{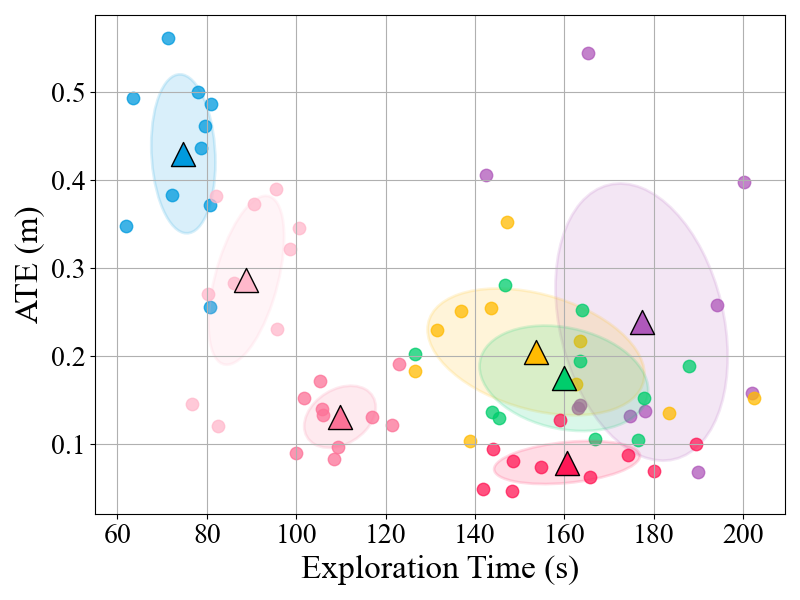}} \hfill
  \subfloat[Warehouse]{\includegraphics[width=0.5\columnwidth]{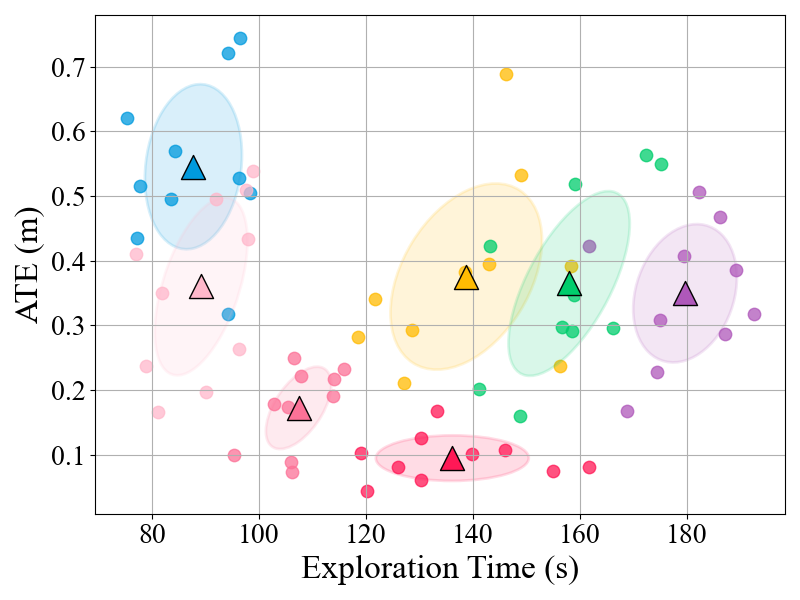}} \hfill
  \subfloat[Office]{\includegraphics[width=0.5\columnwidth]{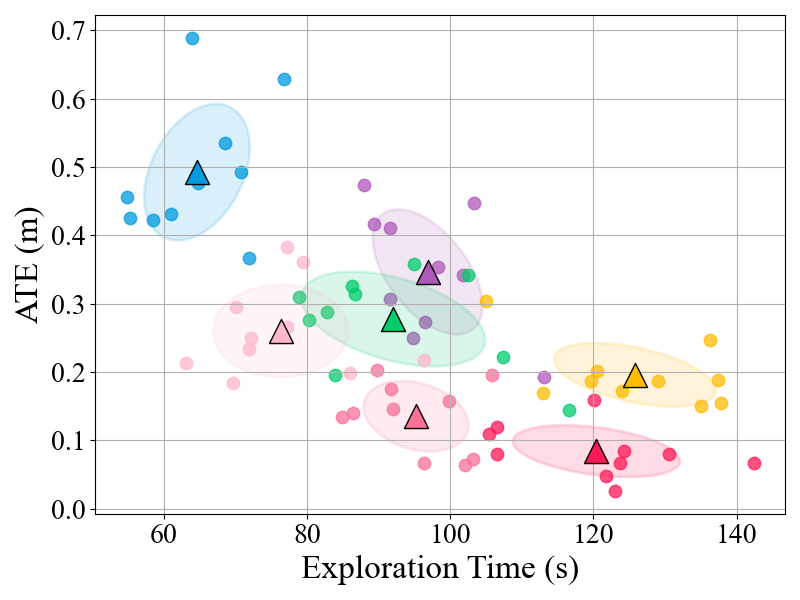}} \hfill
  \subfloat[Apartment]{\includegraphics[width=0.5\columnwidth]{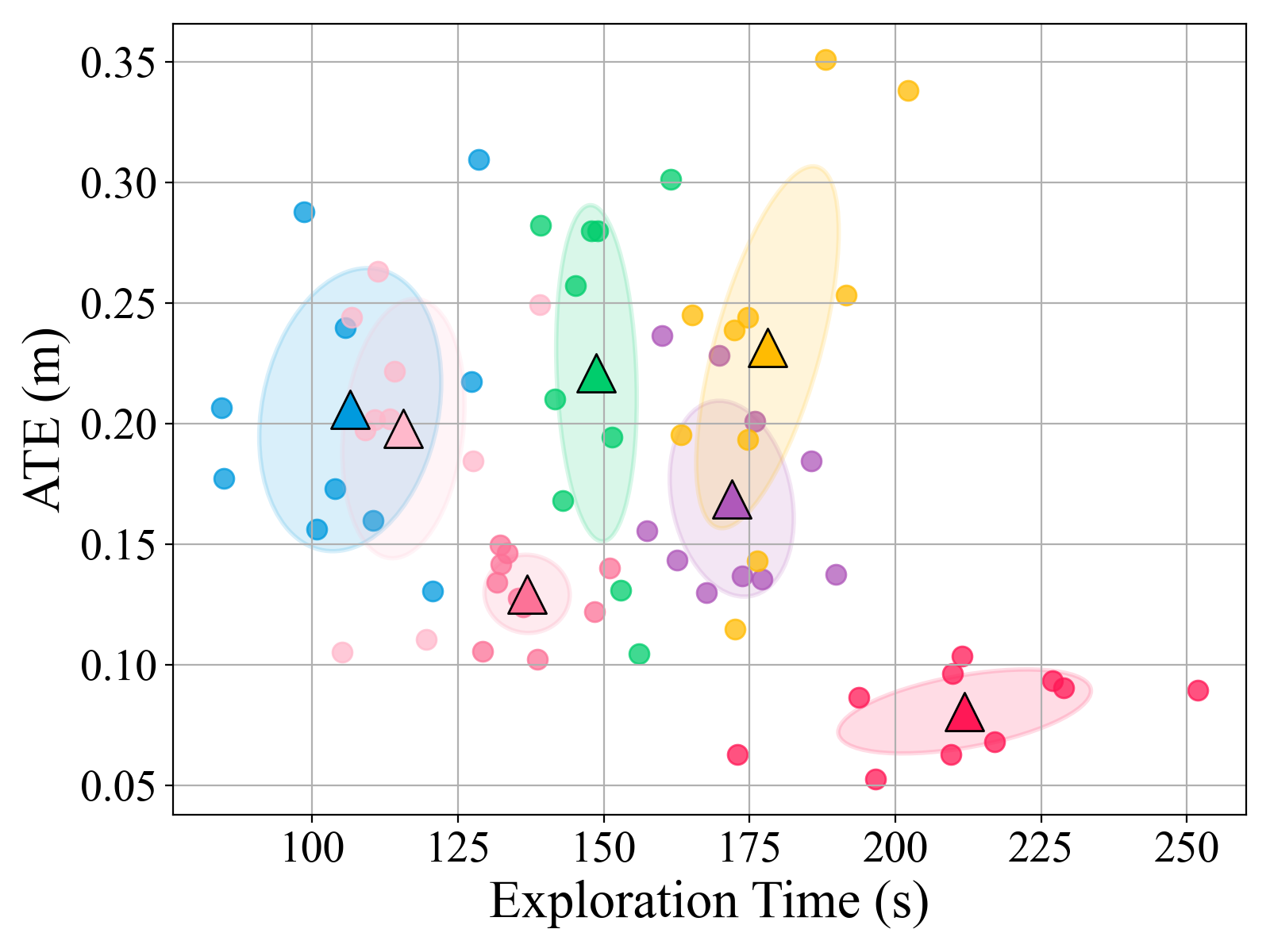}} \hfill
  \vspace{-0.1cm}
  \caption{\label{fig:exp_pareto_plot} Benchmark result plots of exploration time versus ATE, where average values are marked as triangles and $1$-$\sigma$ confidence ellipses are depicted.}
  \vspace{-0.6cm}
\end{figure*}

\subsubsection{Result Discussions}
\begin{figure*}[t]
  \captionsetup[subfloat]{farskip=-1pt, captionskip=-2pt}
	\centering
  \subfloat[Ground Truth]{\includegraphics[width=0.48\columnwidth]{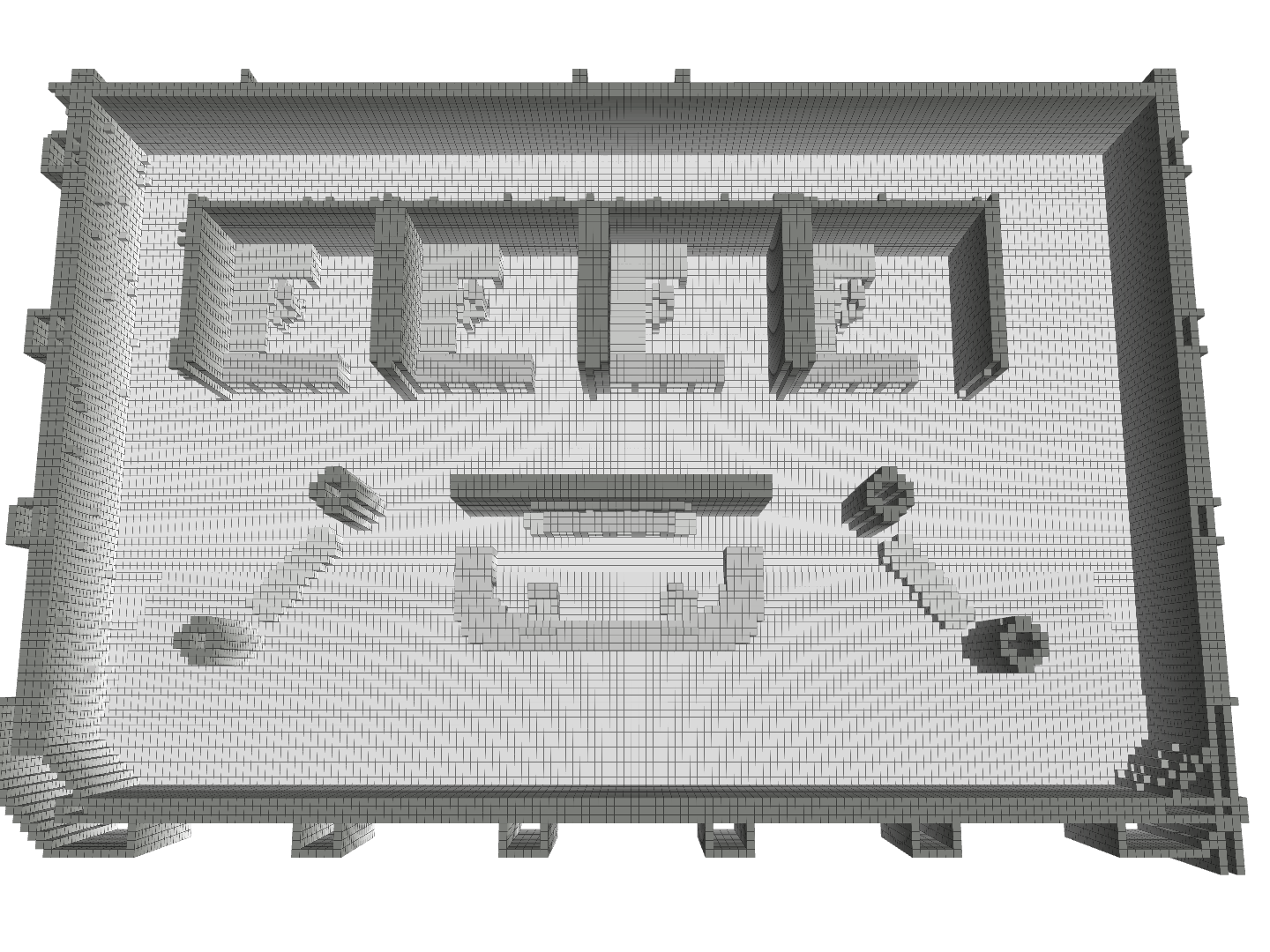}} \hfill
  \subfloat[SRE]{\includegraphics[width=0.48\columnwidth]{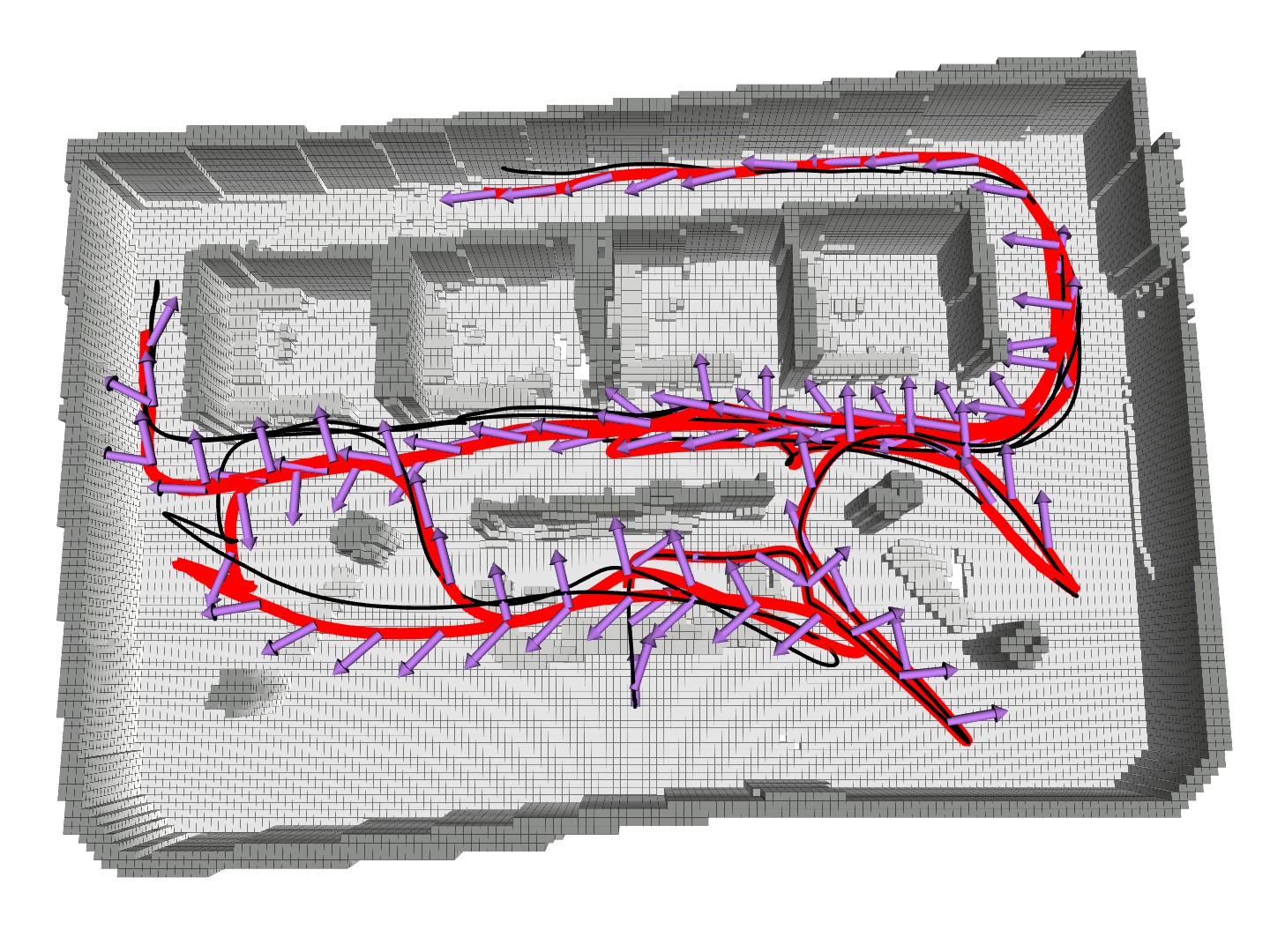}} \hfill
  \subfloat[REAL]{\includegraphics[width=0.48\columnwidth]{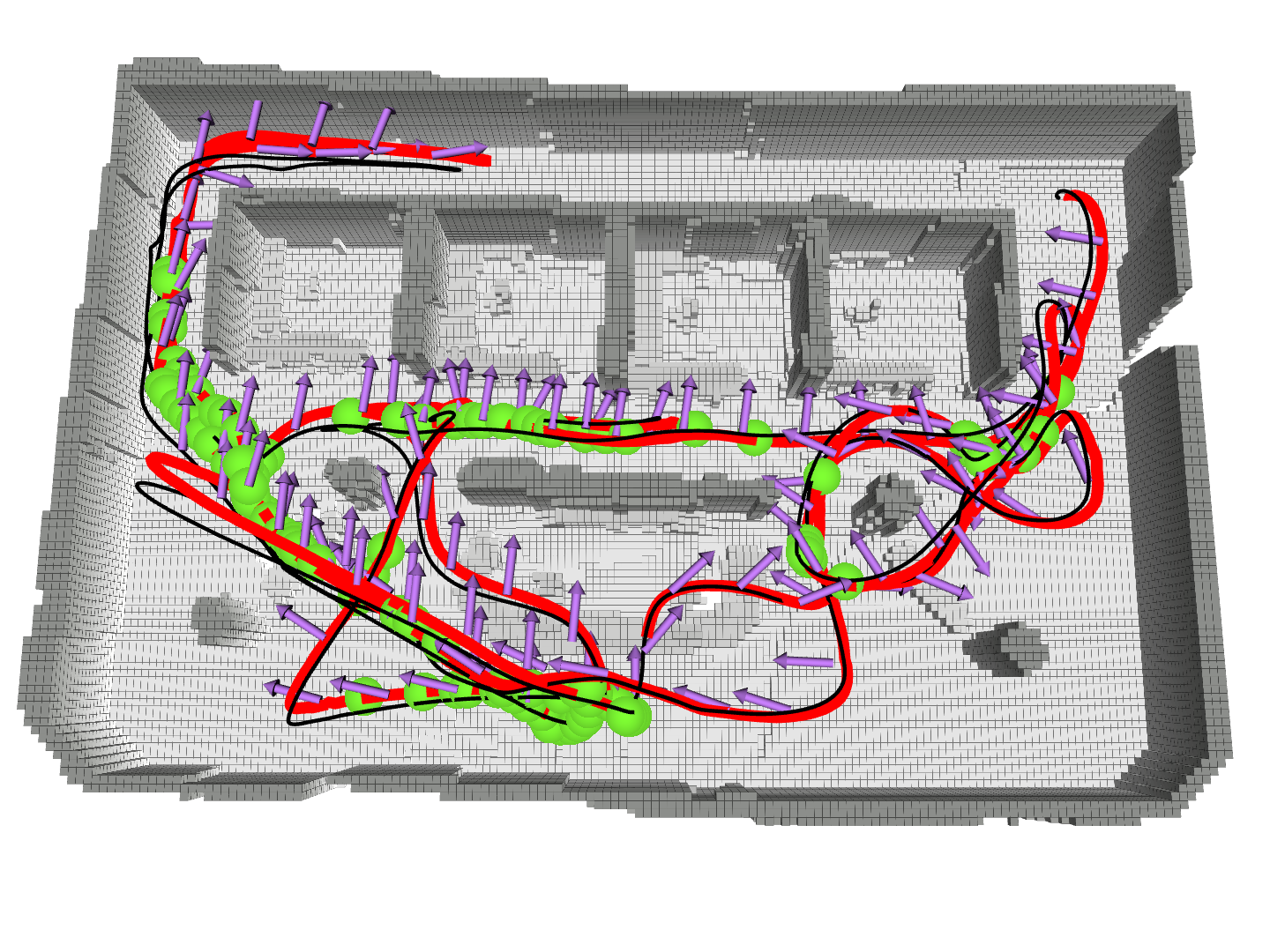}} \hfill
  \subfloat[ExplORB]{\includegraphics[width=0.48\columnwidth]{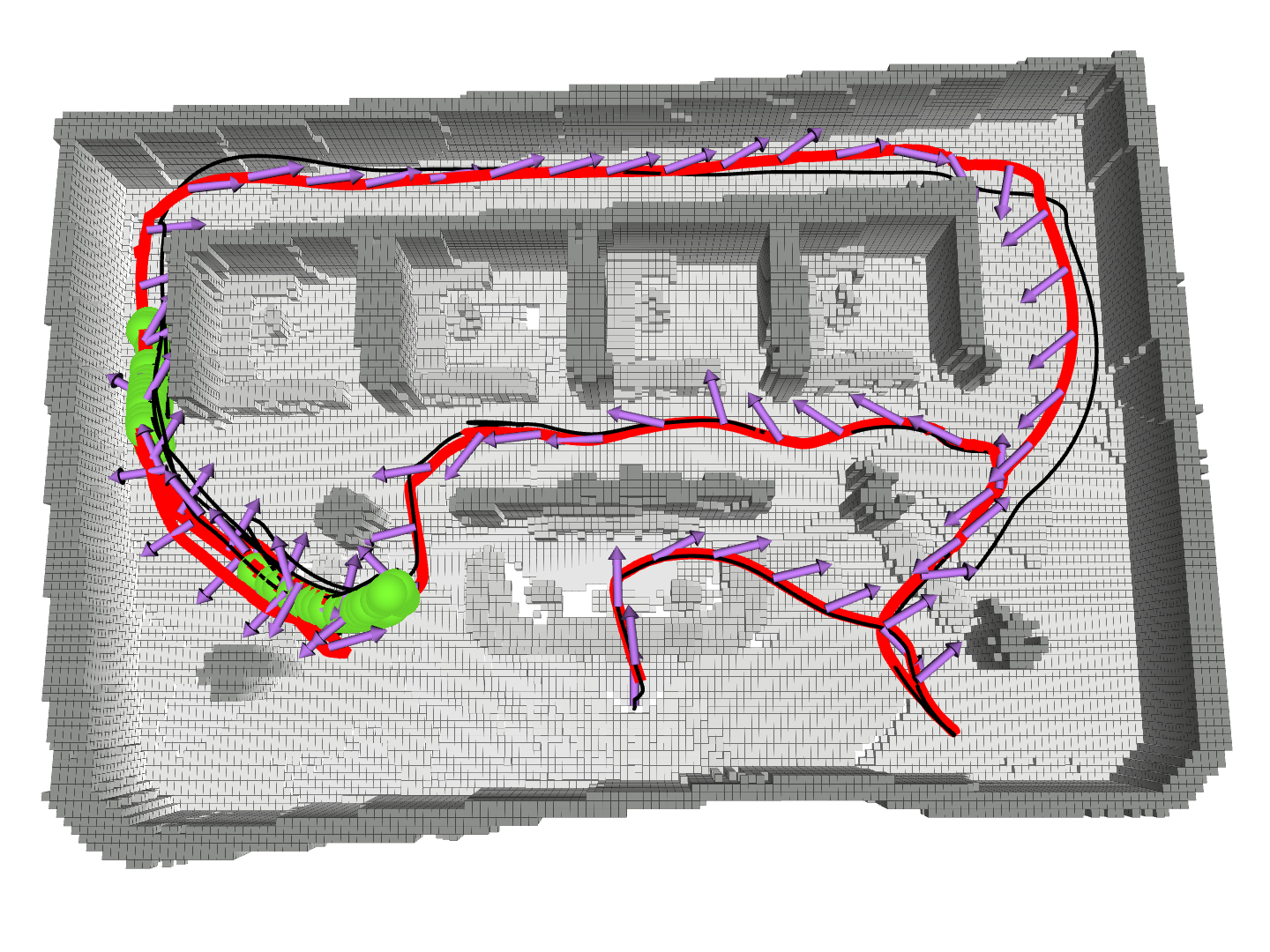}} \hfill \\
  \subfloat[FALCON]{\includegraphics[width=0.48\columnwidth]{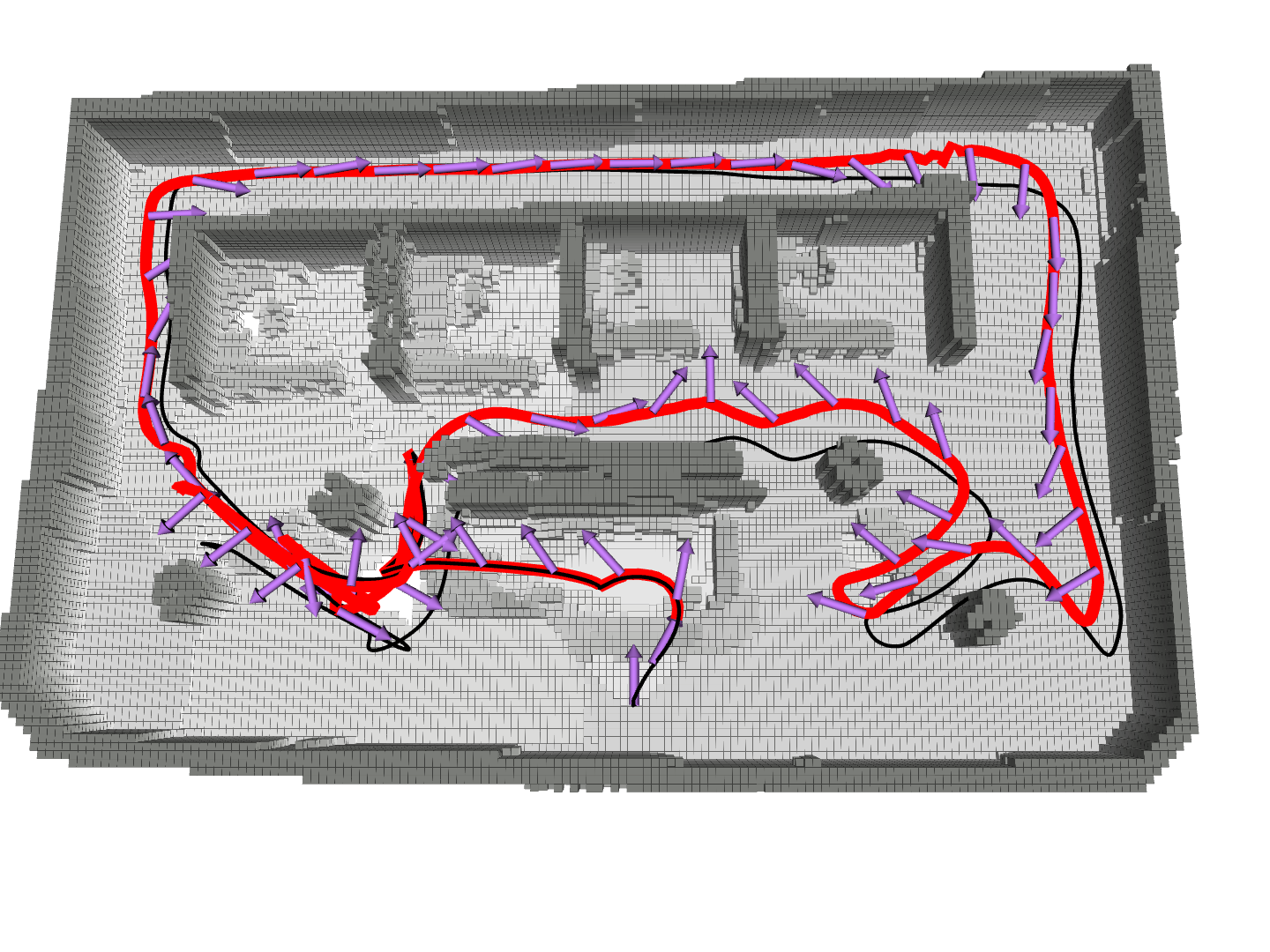}} \hfill
  \subfloat[RAVEN-0.0]{\includegraphics[width=0.48\columnwidth]{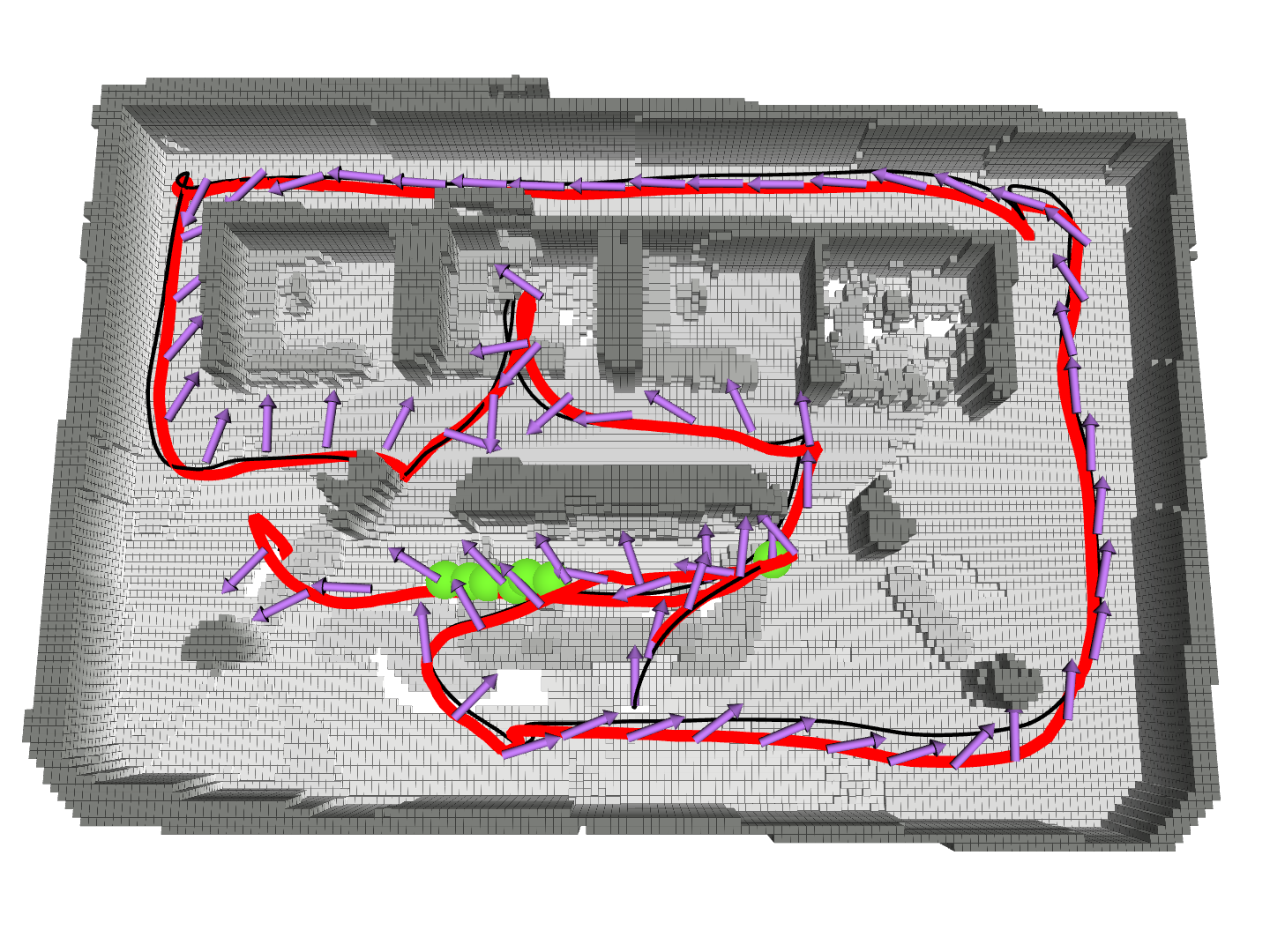}} \hfill
  \subfloat[RAVEN-0.5]{\includegraphics[width=0.48\columnwidth]{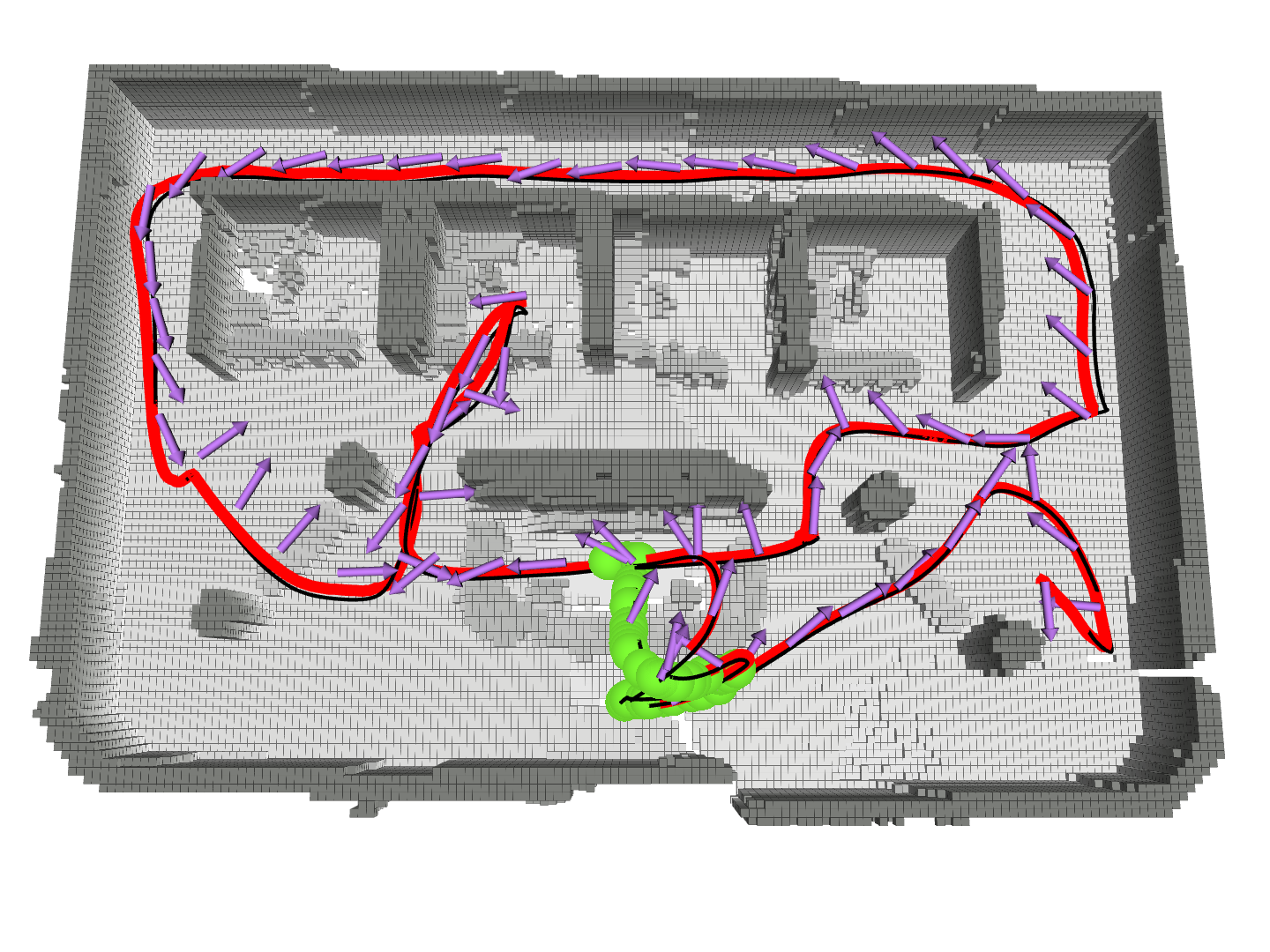}} \hfill
  \subfloat[RAVEN-1.0]{\includegraphics[width=0.48\columnwidth]{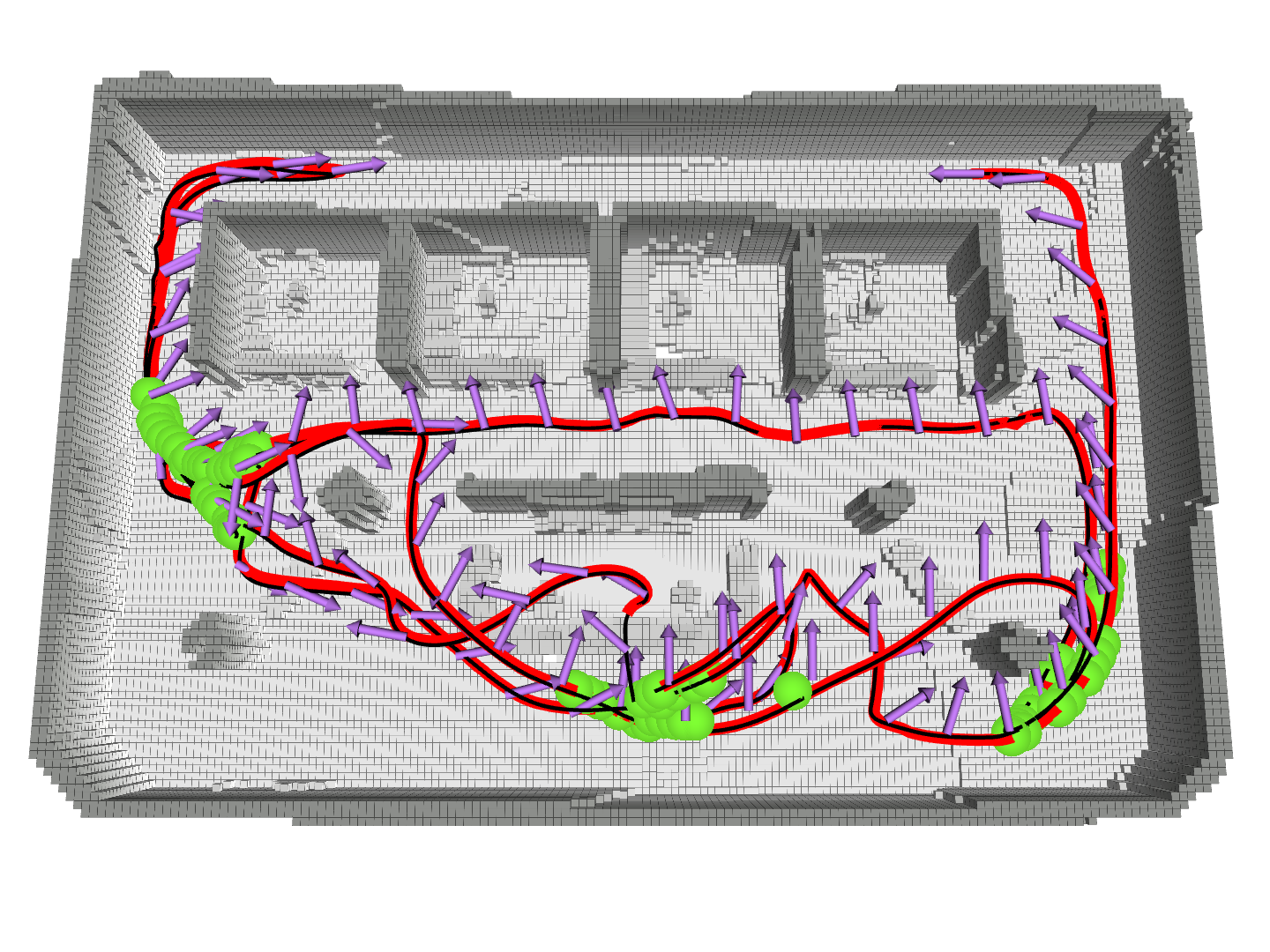}} \hfill
  \caption{\label{fig:sim_exe_traj} The final executed trajectories and reconstructed volumetric maps of all the benchmark candidates in the \textit{Office} scenario. The ground truth trajectories are shown in black curves, and detected loop closures are highlighted with green dots.}
  \vspace{-0.6cm}
\end{figure*}

\begin{figure}[t]
  \centering
  \includegraphics[width=0.9\columnwidth]{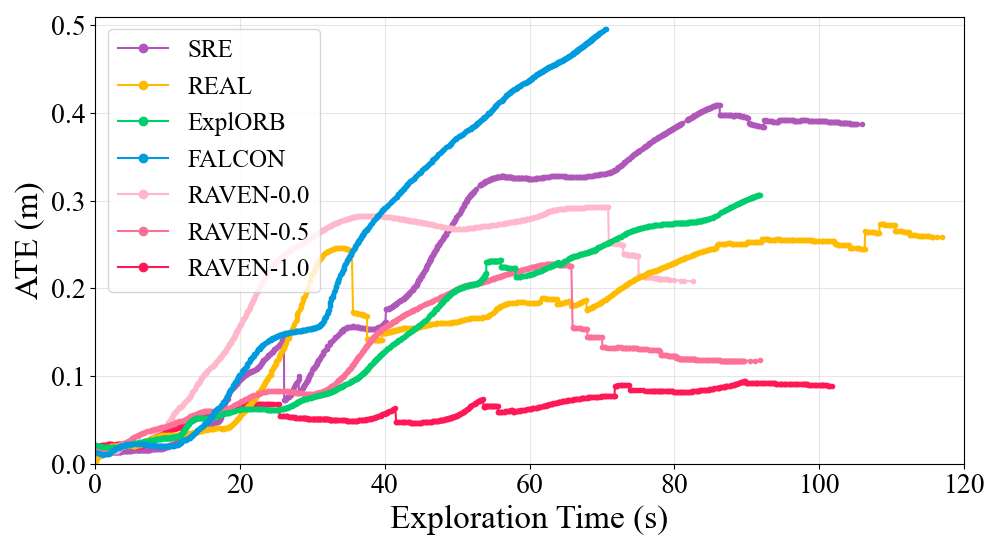}  
  \vspace{-0.4cm}
  \caption{\label{fig:exp_ate} The change in ATE throughout the exploration trails in Fig.~\ref{fig:sim_exe_traj}, where discontinuities indicate loop closures and pose graph rectification.}
  \vspace{-0.4cm}
\end{figure}

\begin{figure}[t]
  \centering
  \includegraphics[width=0.9\columnwidth]{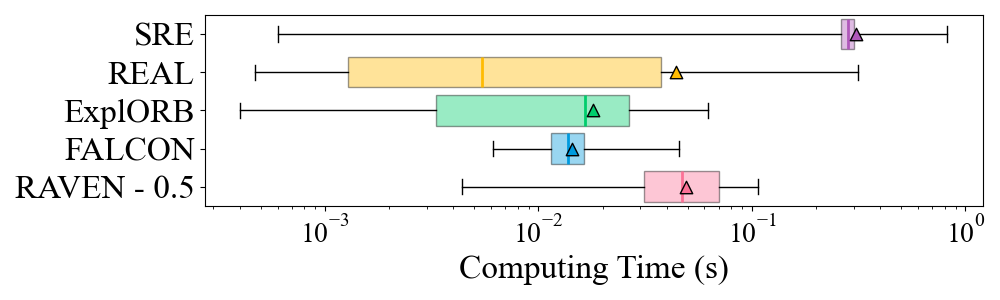}  
  \vspace{-0.4cm}
  \caption{\label{fig:exp_time} Box plot of planning time per iteration for all benchmarked planners in the \textit{Warehouse} scenario with medians/averages indicated by lines/triangles.}
  \vspace{-1.2cm}
\end{figure}

\begin{figure}[t]
  \centering
  \subfloat[RAVEN-0.0]{\includegraphics[width=0.5\columnwidth]{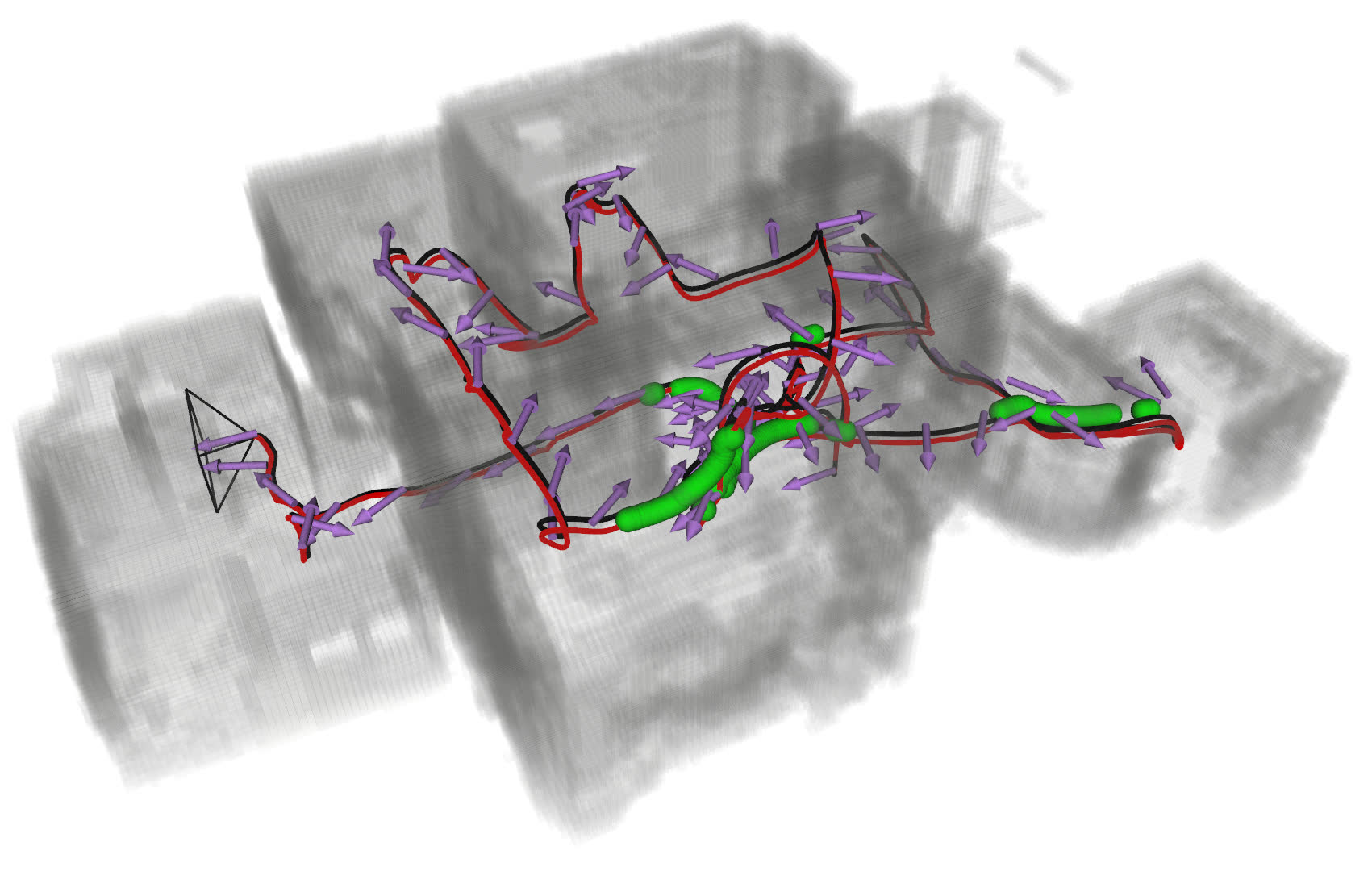}} \hfill
  \subfloat[RAVEN-1.0]{\includegraphics[width=0.5\columnwidth]{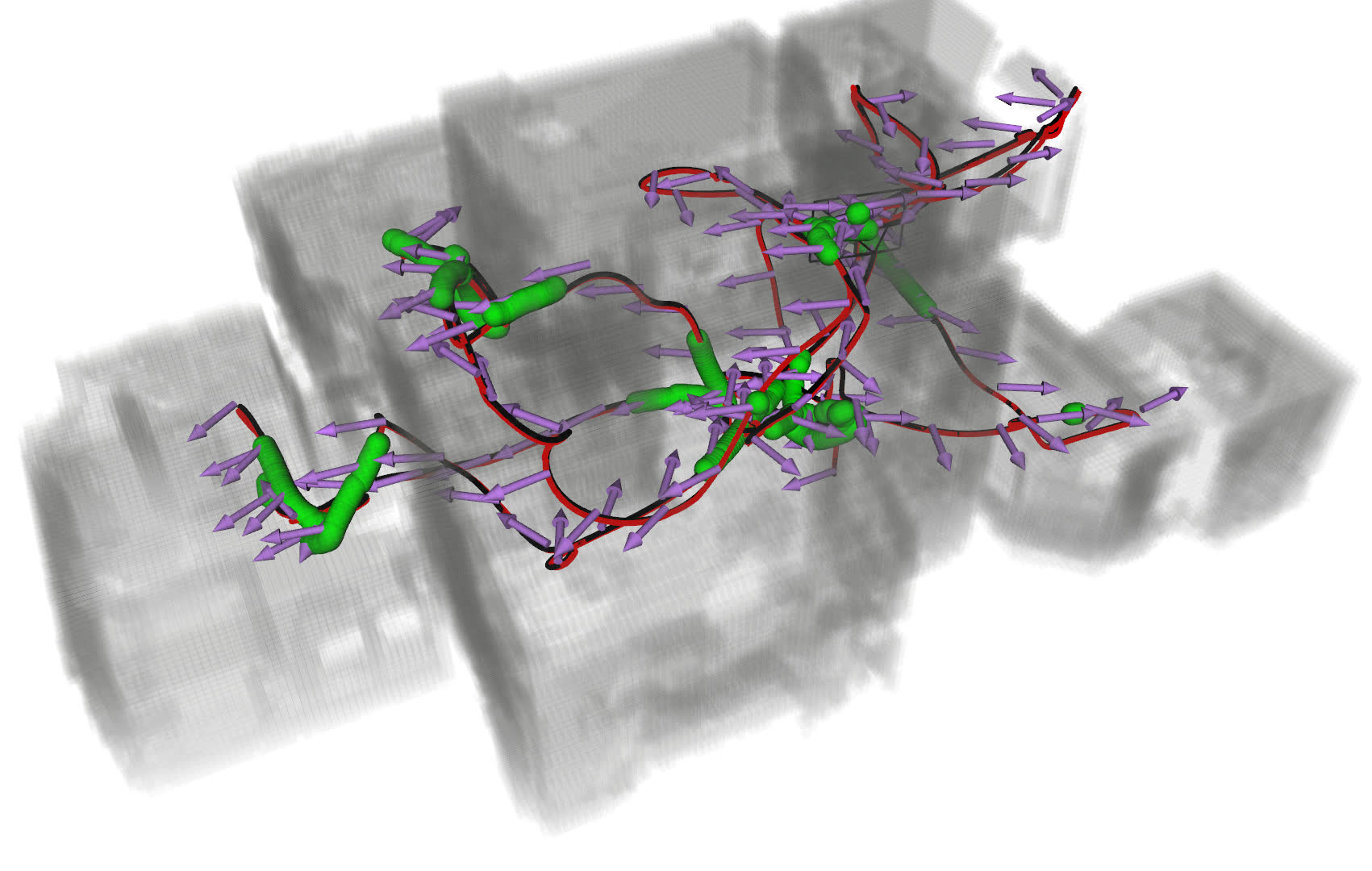}} \hfill
  \vspace{-0.1cm}
  \caption{\label{fig:sim_exe_traj2} The final executed trajectories and reconstructed maps of RAVEN-0.0 and RAVEN-1.0 in the \textit{Apartment} scenario.}
  \vspace{-0.4cm}
\end{figure}

The benchmark experiments are conducted in four scenarios, namely \textit{Corridor}, \textit{Warehouse}, \textit{Office}, and \textit{Apartment}\cite{ramakrishnan2021hm3d}, as shown in Fig.~\ref{fig:exp_scenarios}.
For each planner in each scenario, statistics from $10$ runs are summarized in Table~\ref{tab:result} and visualized in Fig.~\ref{fig:exp_pareto_plot}, where average values and $1$-$\sigma$ confidence ellipses are highlighted.
The final executed trajectories and the ATE changes of a single exploration trial in the \textit{Office} scenario are presented in Fig.~\ref{fig:sim_exe_traj} and Fig.~\ref{fig:exp_ate}, respectively. 
Due to space limitations, the complete exploration experiments are provided in the supplementary video.
As shown in Fig.~\ref{fig:exp_pareto_plot}, RAVEN and FALCON define the Pareto front in terms of exploration time and ATE in all four scenarios.
Since the classic exploration planner FALCON focuses solely on exploration efficiency, it achieves the shortest exploration duration but suffers from substantially higher localization drift compared to the other active SLAM planners.
The proposed planner RAVEN, when configured with $\lambda=0$ to prioritize exploration efficiency, achieves a comparable exploration duration to FALCON while significantly reducing localization drift in terms of average ATE, thanks to the perception-aware trajectory planning.
When set to $\lambda=1$ prioritizing localization accuracy, RAVEN performs frequent active loop closures, yielding consistently low localization error with minimal average and standard deviation in ATE.
With $\lambda=0.5$, RAVEN offers a balanced trade-off between exploration efficiency and localization accuracy, delivering reasonable exploration durations alongside acceptable localization drift.
In contrast, the other three active SLAM algorithms, SRE, REAL, and ExplORB, exhibit longer exploration times and higher localization errors compared to RAVEN, producing distorted maps as shown in Fig.~\ref{fig:sim_exe_traj}(b)-(d).
Their performance is hindered by oscillatory back-and-forth trajectories caused by less effective active loop closure strategies and the absence of perception-aware navigation planning.
Furthermore, Fig.~\ref{fig:exp_time} plots the one iteration planning time of each planner in the \textit{Warehouse} scenario, where RAVEN-0.5 serves as a representative of this unified framework.
Fig.~\ref{fig:sim_exe_traj2} shows the exploration results in the \textit{Apartment} scene, where RAVEN-1.0 yields an estimated trajectory (red) closer to the ground truth (black) with more detected loop closures (green spheres), at the cost of exploration efficiency.
In summary, the benchmark experiments demonstrate that RAVEN achieves a more effective balance between exploration efficiency and localization accuracy compared to the state-of-the-art methods across diverse scenarios.

\section{Ablation Studies}
\label{sec:ablation}

\subsection{Ablation Study on \texorpdfstring{$\lambda$}{lambda} and Perception-aware Navigation}

\begin{figure}[t]
	\centering
  \includegraphics[width=0.9\columnwidth]{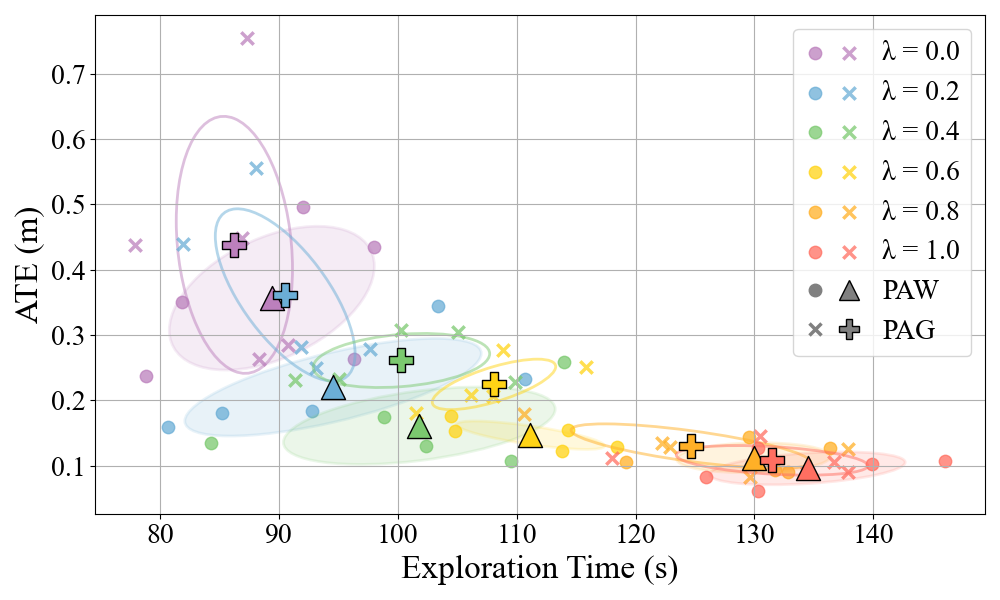}  
  \vspace{-0.4cm}
  \caption{\label{fig:abla_lambda} Ablation study results for a series of prioritization parameter $\lambda$ values in the \textit{Warehouse} scenario. 
  Triangle (Plus) markers and filled (hollow) ellipses indicate means and 1-$\sigma$ confidence intervals using a perception-aware (PAW) (perception-agnostic (PAG)) local planner respectively.}
  \vspace{-1.2cm}
\end{figure}

This ablation study investigates the effect of the prioritization parameter $\lambda$ on RAVEN's performance.
We also compare the proposed perception-aware local planner (PAW) with a perception-agnostic variant (PAG) that does not consider visual feature distribution in trajectory generation.
We evaluate six $\lambda$ values ranging from $0$ to $1$ in increments of $0.2$ in the \textit{Warehouse} scenario with $5$ runs per setting.
Fig.~\ref{fig:abla_lambda} reports the results in terms of exploration time and ATE, along with the mean and standard deviation for each $\lambda$ and local planner variant.
With a perception-agnostic local planner, the average ATE is $12.74-64.19\%$ higher across $\lambda$ values compared to the PAW planner, and the average exploration time is $1.48-4.26\%$ shorter as ignorance of perception constraints gives more freedom to the planner to select shorter paths.
In general, the Pareto front of PAG is dominated by that of PAW, validating the effectiveness of the perception-aware navigation strategy.
Moreover, as $\lambda$ increases, the exploration time gradually increases while the ATE decreases, demonstrating a consistent trade-off between exploration efficiency and localization accuracy.
These results validate $\lambda$ as an effective control variable, allowing RAVEN to flexibly adapt performance along a spectrum that aligns with varying task requirements for general-purpose autonomous exploration.

\subsection{Ablation Study on Loop Closure Probability Weighting}

\begin{figure}[t]
  \centering
  \captionsetup[subfloat]{captionskip=-0.5pt}
  \subfloat[No Weighting]{\includegraphics[width=0.3\columnwidth]{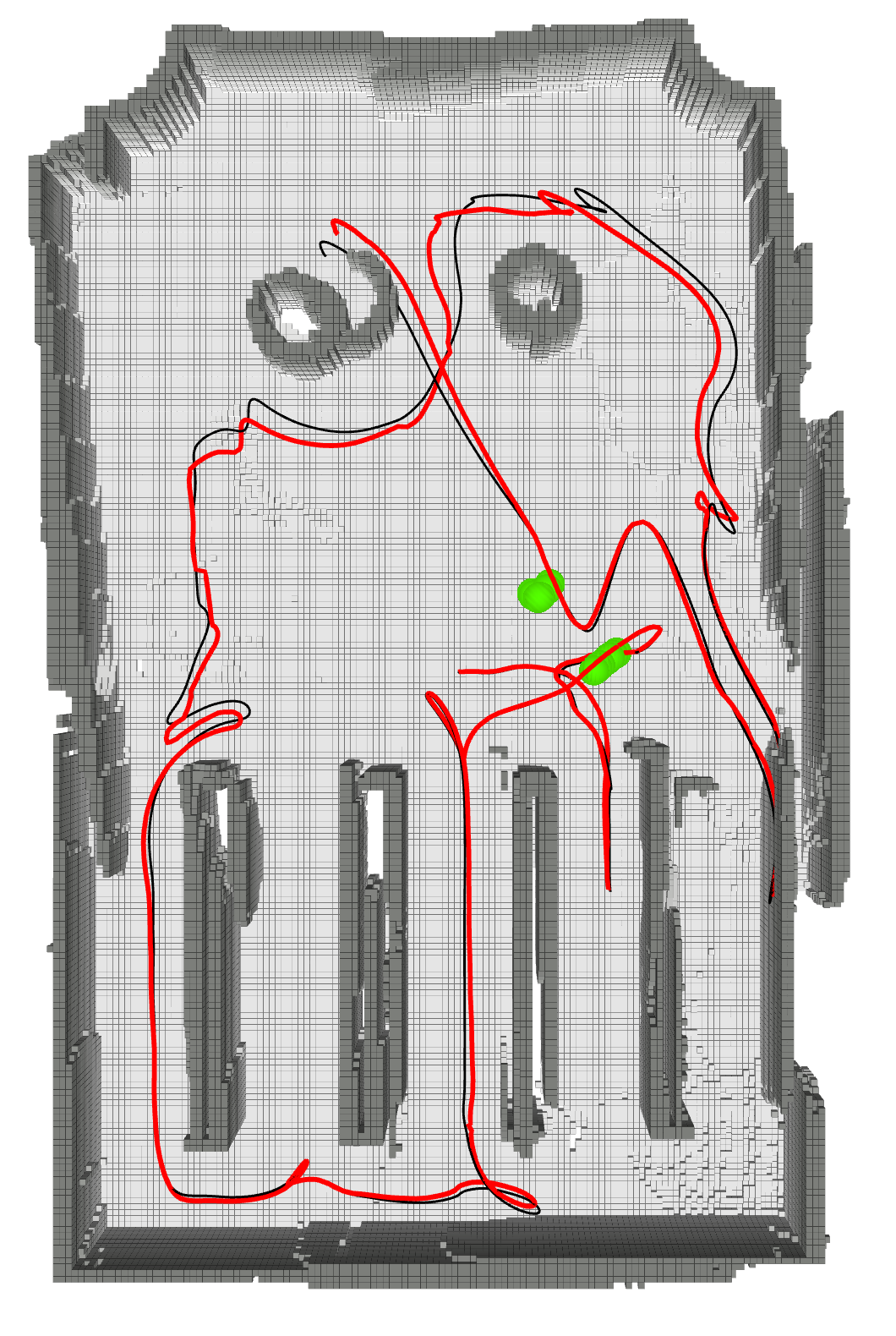}}
  \subfloat[Uniform Threshold]{\includegraphics[width=0.3\columnwidth]{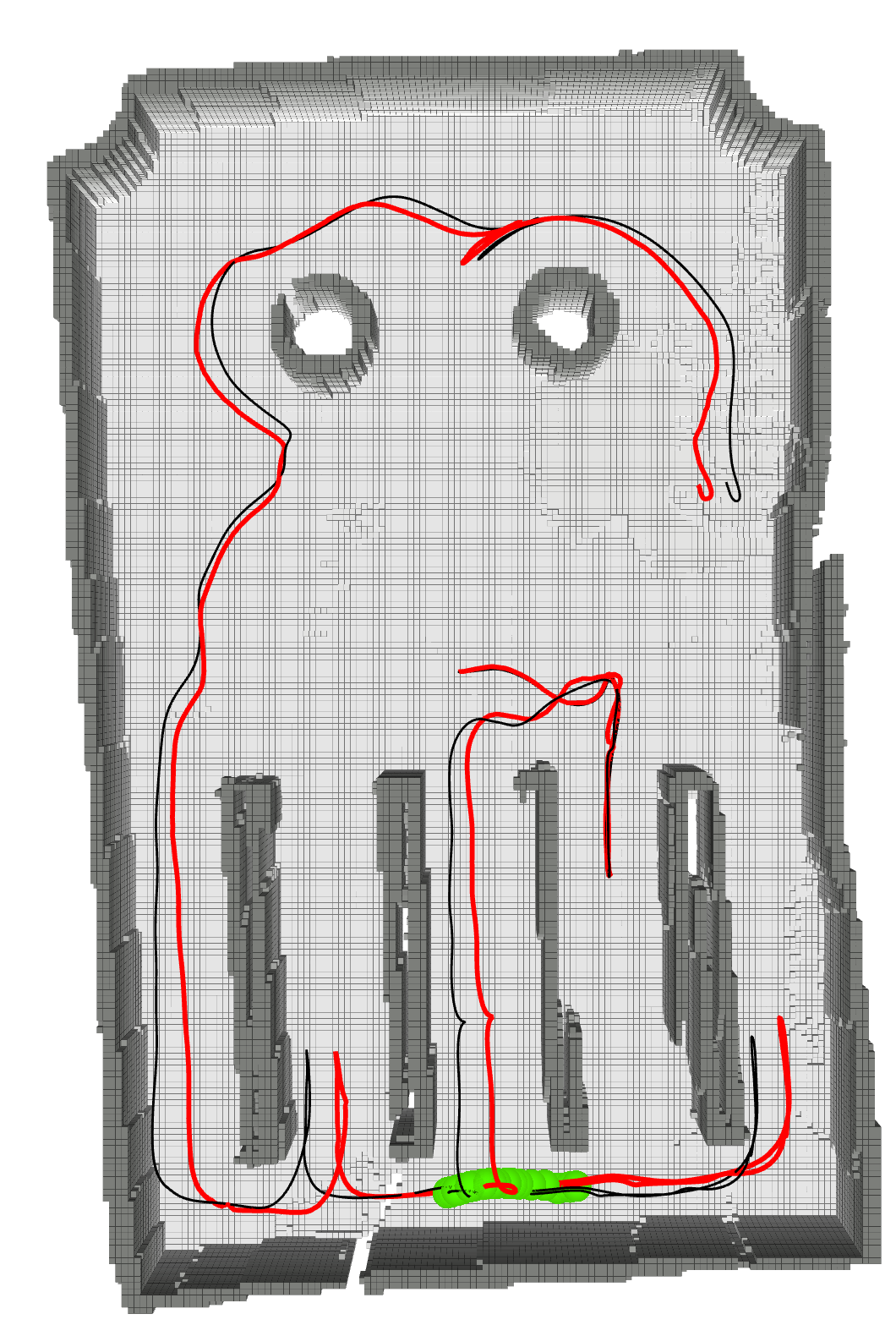}}
  \subfloat[Proposed]{\includegraphics[width=0.3\columnwidth]{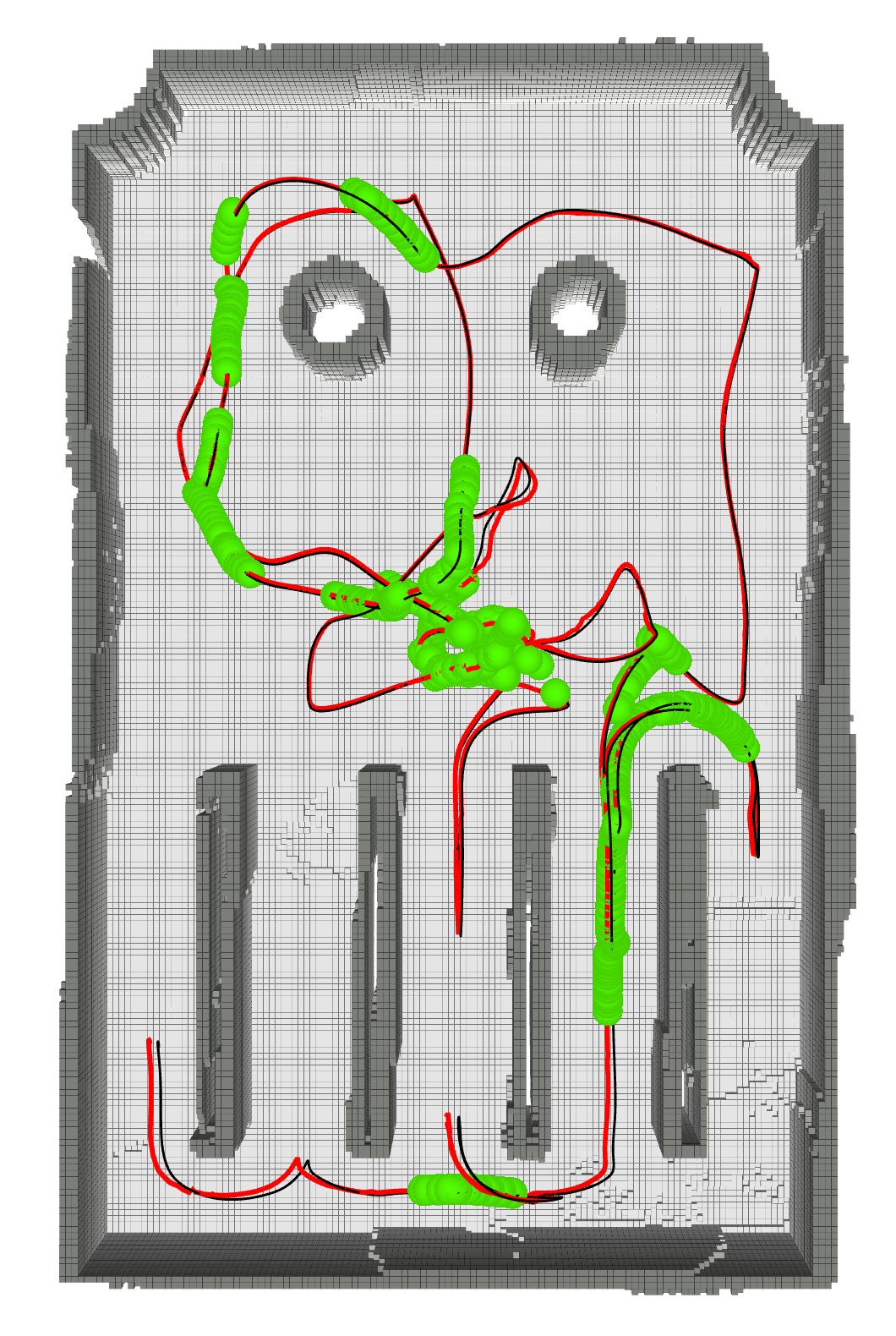}}
  \vspace{-0.1cm}
  \caption{\label{fig:abla_succ_prob} Ablation study results on loop closure probability weighting, showing the reconstructed maps, the estimated and ground truth trajectories in red and black curves, as well as the detected loop closures in green dots.}
  \vspace{-0.2cm}
\end{figure}

To validate the necessity of probability weighting in localization accuracy utility evaluation (Sec.~\ref{subsec:loc_util}), we compare three strategies in the \textit{Warehouse} scene: (1) no probability weighting, which assumes $p_\text{succ}=1$ for all ALC candidates; (2) uniform thresholding, which rejects ALC candidates with $p_\text{succ} < 0.7$; and (3) our proposed model (\ref{eqn:loc_util}). 
The first two baselines compute the localization accuracy utility from the success graph $\mathcal{G}^+(\pi^k_{i,j})$, assuming all active loop closures will succeed.
We conducted $5$ independent trials for each strategy with $\lambda=1.0$ and representative results for each baseline are illustrated in Fig.~\ref{fig:abla_succ_prob}.
The first strategy consistently defers loop closure decisions, yielding an average of only $10.0$ looped image pairs and a final ATE of $0.45$ m. 
The uniform thresholding approach won't take risk for large loop closures, resulting in a final ATE of $0.46$ m after completing an average of $71.8$ loop closures.
In contrast, our proposed method effectively produces the highest localization accuracy with an average final ATE of $0.09$ m with $150.6$ loop pairs that are both beneficial and feasible.
These results confirm that probability weighting is essential for balancing theoretical benefit against practical feasibility in active loop closure planning.

\begin{figure}[t]
  \centering
  \captionsetup[subfloat]{farskip=-1pt, captionskip=-1pt}
  \subfloat{\includegraphics[width=0.9\columnwidth]{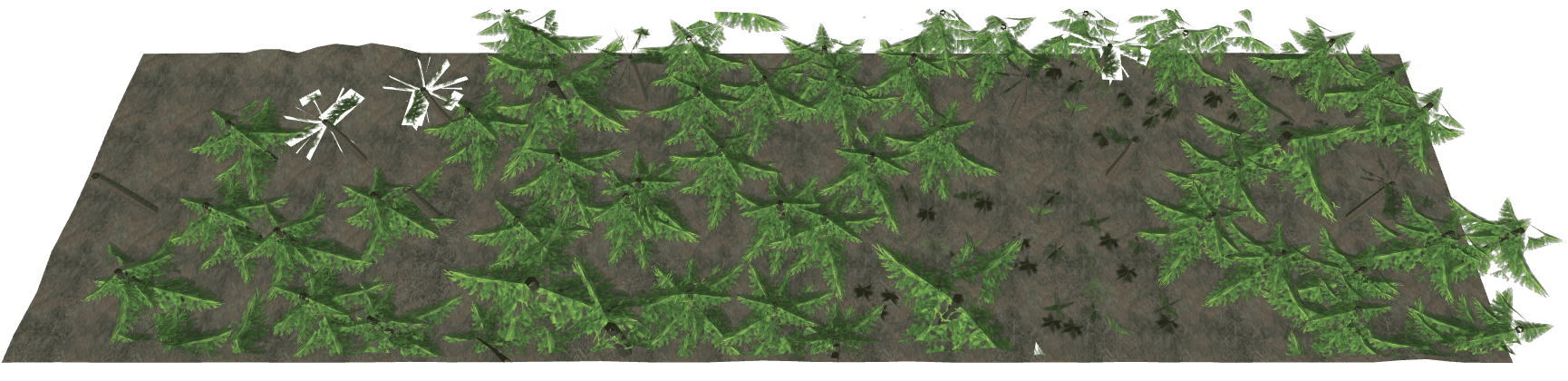}}\\
  \subfloat{\includegraphics[width=0.9\columnwidth]{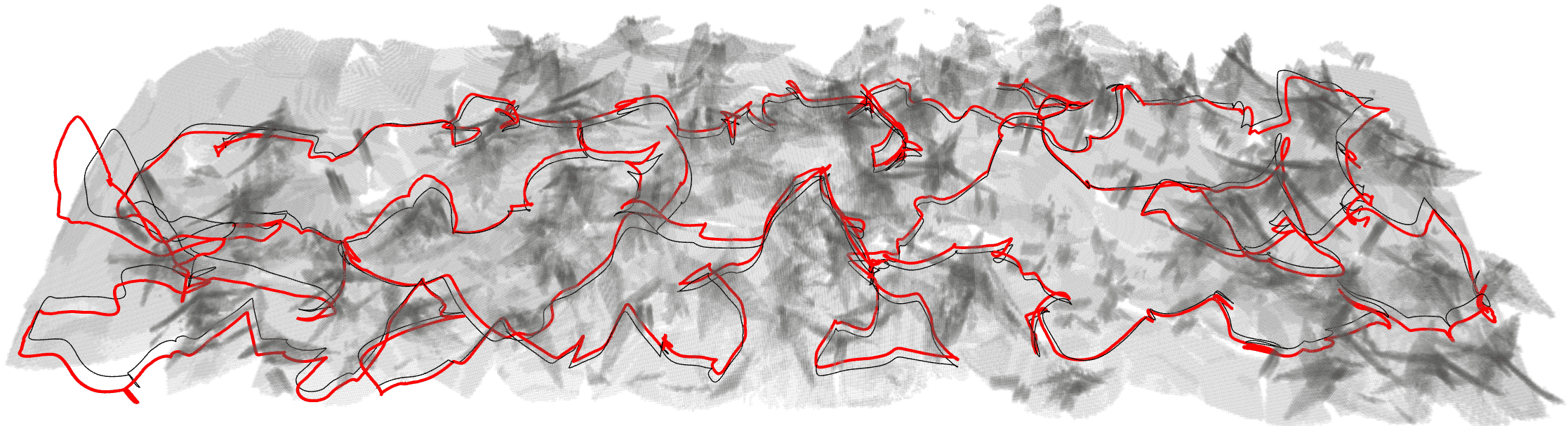}} \\
  \subfloat{\includegraphics[width=0.9\columnwidth]{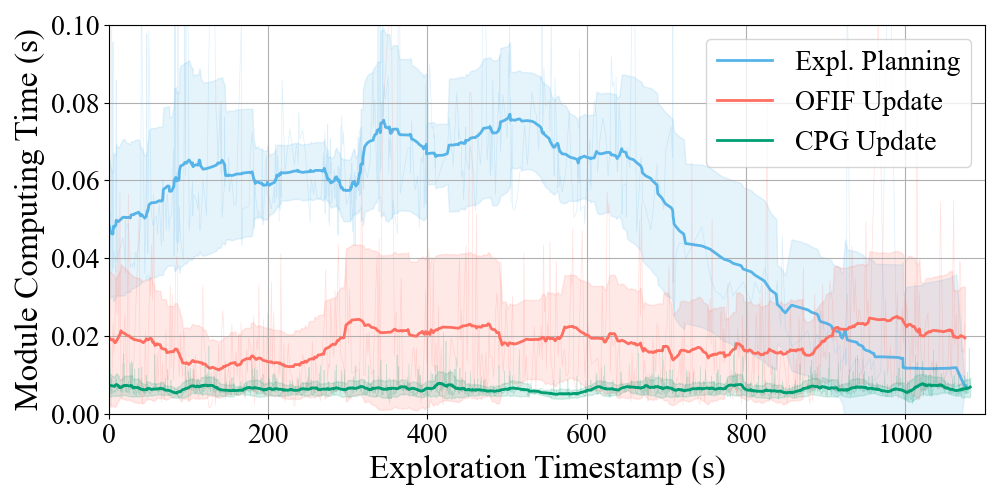}}
  \vspace{-0.4cm}
  \caption{\label{fig:abla_large} Top: Scene photo of the \textit{Forest} scenario. Middle: Exploration results from RAVEN - 0.5. Bottom: Computation breakdown of RAVEN modules throughout the exploration.}
  \vspace{-1.2cm}
\end{figure}

\subsection{Exploration in Large-Scale Environment}
To validate the scalability of the proposed framework RAVEN, we conduct exploration experiments in the complex large-scale \textit{Forest}\footnote{\url{https://skfb.ly/69RVV}} scene with dimensions $80 \times 20 \times 2\text{ m}^3$, which features dense vegetation and undulating terrain, as shown at the top of Fig.~\ref{fig:abla_large}.
Deploying RAVEN with $\lambda=0.5$, the system completed the exploration task in $1081.6$ seconds with a trajectory length of $759.6$~m and a final trajectory ATE of $0.664$~m.
The reconstructed volumetric map as well as the executed (red) and ground truth (black) trajectories are visualized in the middle of Fig.~\ref{fig:abla_large}.
The computation breakdown for each of the RAVEN core modules throughout the exploration process is shown at the bottom of Fig.~\ref{fig:abla_large}.
As the OFIF and CPG updates run in separate threads periodically, each replan iteration only performs exploration planning, which takes $0.057$~s on average and shows no significant increase during the exploration.
Memory usage remains capped at $2.8$~GB throughout the exploration, as RAVEN incurs minimal additional memory beyond the volumetric map storage.
These results validate the scalability and effectiveness of the RAVEN framework in large-scale environments and its feasibility for real-world deployments detailed in the next section.

\begin{table}[t]
  \caption{Real-World Experiments Statistics}
  \vspace{-0.2cm}
  \begin{tabular}{M{0.75cm}M{0.25cm}M{0.28cm}M{0.28cm}M{0.28cm}M{0.28cm}M{0.28cm}M{0.28cm}M{0.28cm}M{0.28cm}M{0.28cm}M{0.28cm}}
    \toprule\toprule
     & & \multicolumn{4}{c}{\textbf{Expl. Efficiency}} & \multicolumn{6}{c}{\textbf{Localization Accuracy}} \\
    \cmidrule(lr){3-6} \cmidrule(lr){7-12} 
    \multirow{2}{*}{\begin{tabular}[c]{@{}c@{}}\textbf{Scene}\end{tabular}} & \multirow{2}{*}{\begin{tabular}[c]{@{}c@{}}\boldsymbol{$\lambda$}\end{tabular}} 
          & \multicolumn{2}{c}{\begin{tabular}[c]{@{}c@{}}\textbf{Explore}\\\textbf{Time(s)}\end{tabular}}
          & \multicolumn{2}{c}{\begin{tabular}[c]{@{}c@{}}\textbf{Flight}\\\textbf{Dist.(m)}\end{tabular}}
          & \multicolumn{2}{c}{\begin{tabular}[c]{@{}c@{}}\textbf{$^\dagger$ATE/}\\\textbf{NND(m)}\end{tabular}}
          & \multicolumn{2}{c}{\begin{tabular}[c]{@{}c@{}}\textbf{Looped}\\\textbf{Images}\end{tabular}}
          & \multicolumn{2}{c}{\begin{tabular}[c]{@{}c@{}}\textbf{$^\ddagger$Graph}\\\textbf{Info.}\end{tabular}} \\
          & & \textbf{Avg} & \textbf{Std} & \textbf{Avg} & \textbf{Std} & \textbf{Avg} & \textbf{Std} & \textbf{Avg} & \textbf{Std} & \textbf{Avg} & \textbf{Std} \\
    \midrule
    \multirow{3}{*}{\begin{tabular}[c]{@{}c@{}}Flight\\Arena\end{tabular}} 
      & 0.0 & 44.8 & 5.19 & 33.6 & 3.84 & 0.18 & 0.06 & 4.40 & 6.47 & 16.1 & 5.93 \\
      & 0.5 & 48.7 & 6.09 & 38.2 & 5.17 & 0.14 & 0.05 & 17.8 & 17.3 & 20.1 & 7.69 \\
      & 1.0 & 51.5 & 4.97 & 39.3 & 6.11 & 0.10 & 0.02 & 36.6 & 27.3 & 25.2 & 8.97 \\
    \midrule
    \multirow{3}{*}{\begin{tabular}[c]{@{}c@{}}Campus\\Hallway\end{tabular}} 
      & 0.0 & 61.2 & 7.78 & 57.6 & 7.72 & 0.28 & 0.04 & 23.7 & 20.9 & 18.1 & 6.56 \\
      & 0.5 & 80.8 & 9.54 & 67.7 & 13.1 & 0.21 & 0.03 & 67.3 & 20.1 & 29.3 & 3.25 \\
      & 1.0 & 107 & 16.9 & 86.7 & 10.6 & 0.14 & 0.02 & 168 & 75.3 & 46.6 & 13.6 \\
    \midrule
    \multirow{3}{*}{\begin{tabular}[c]{@{}c@{}}Parking\\Lot\end{tabular}} 
      & 0.0 & 175 & 18.9 & 153 & 14.5 & 0.48 & 0.14 & 19.0 & 21.5 & 11.0 & 0.68 \\
      & 0.5 & 182 & 18.5 & 149 & 8.03 & 0.22 & 0.06 & 36.0 & 11.9 & 13.6 & 1.10 \\
      & 1.0 & 242 & 10.6 & 196 & 5.40 & 0.18 & 0.05 & 167 & 43.6 & 24.5 & 4.12 \\
    \toprule\toprule
  \end{tabular} \\
  \footnotesize{$^\dagger$ For the first scene, we report ATE against the ground truth trajectory obtained from the motion capture system. For the other two scenes without ground truth odometry, we instead evaluate the NND between the reconstructed map and a ground truth scene point cloud. \\
  $^\ddagger$ Graph information is the approximated Fisher information of the entire CPG graph at exploration completion.
  }\\
  \label{tab:realworld_data}
  \vspace{-0.8cm}
\end{table}

\begin{figure}[t]
  \centering
  \captionsetup[subfigure]{justification=centering}
  \subfloat[Flight Arena\\$8.5\times9\times2$ m$^3$]{\includegraphics[width=0.33\columnwidth]{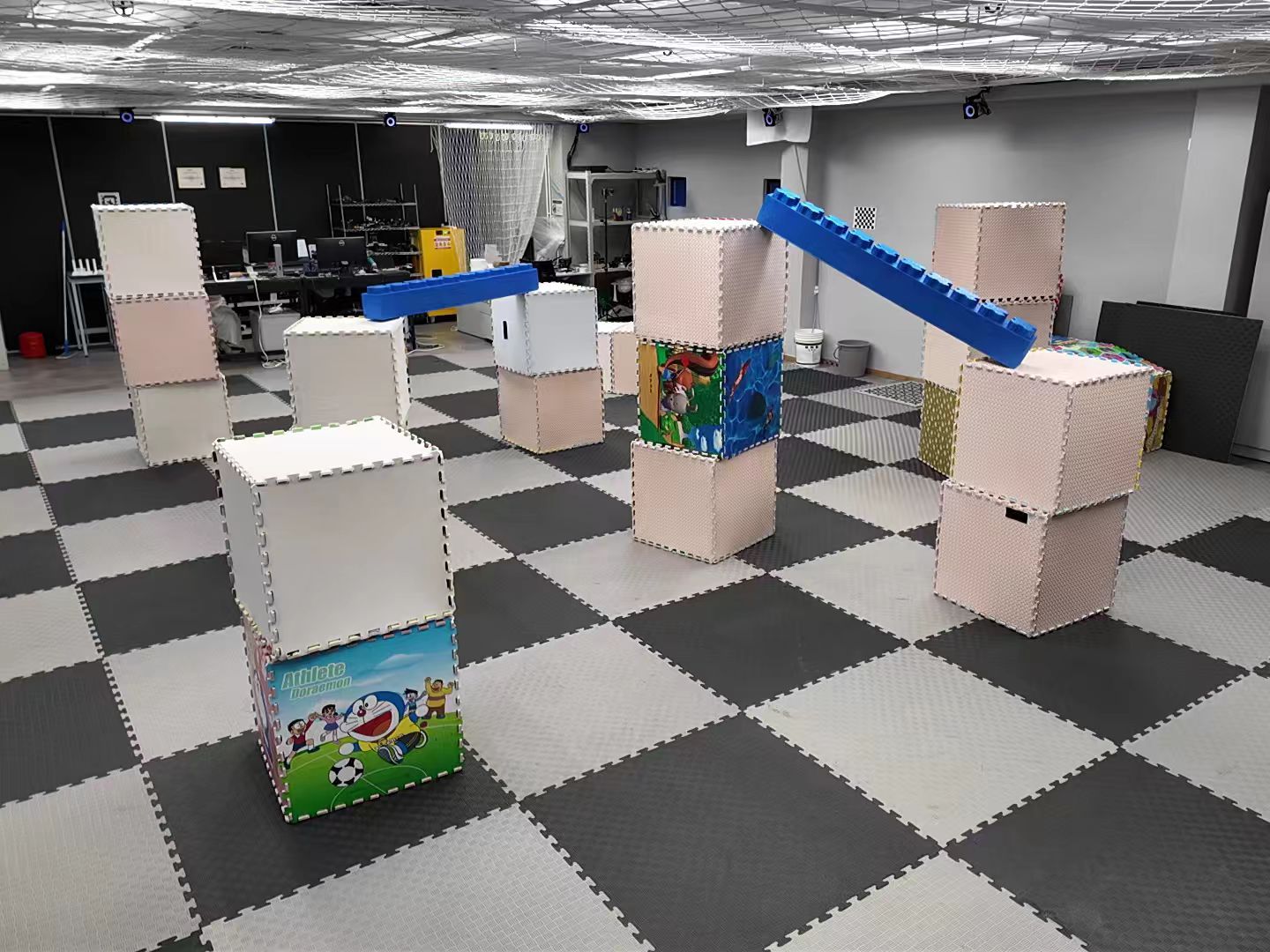}} \hfill
  \subfloat[Campus Hallway\\$17\times10\times2$ m$^3$]{\includegraphics[width=0.33\columnwidth]{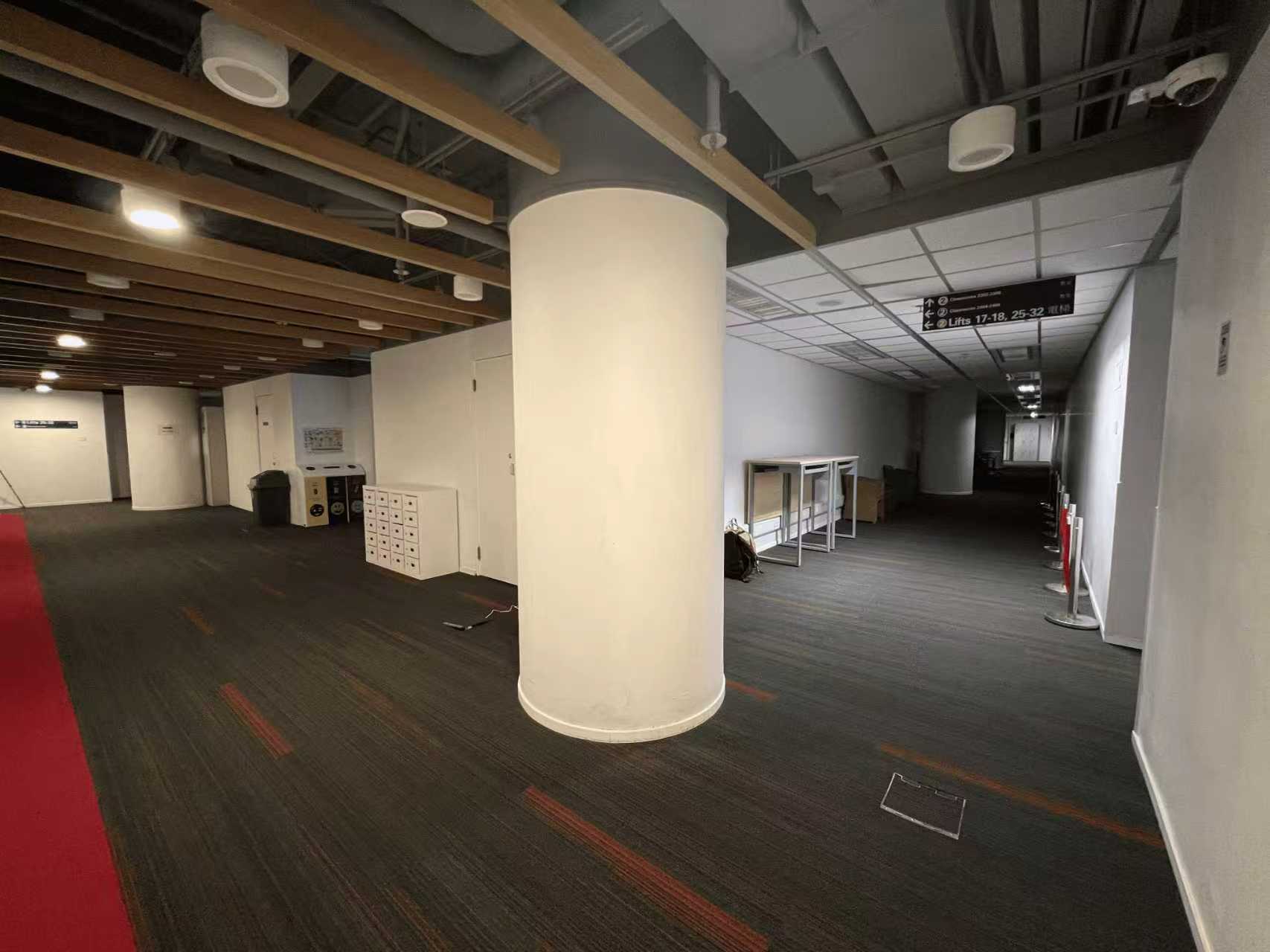}} \hfill
  \subfloat[Parking Lot\\$28.5\times16\times2$ m$^3$]{\includegraphics[width=0.33\columnwidth]{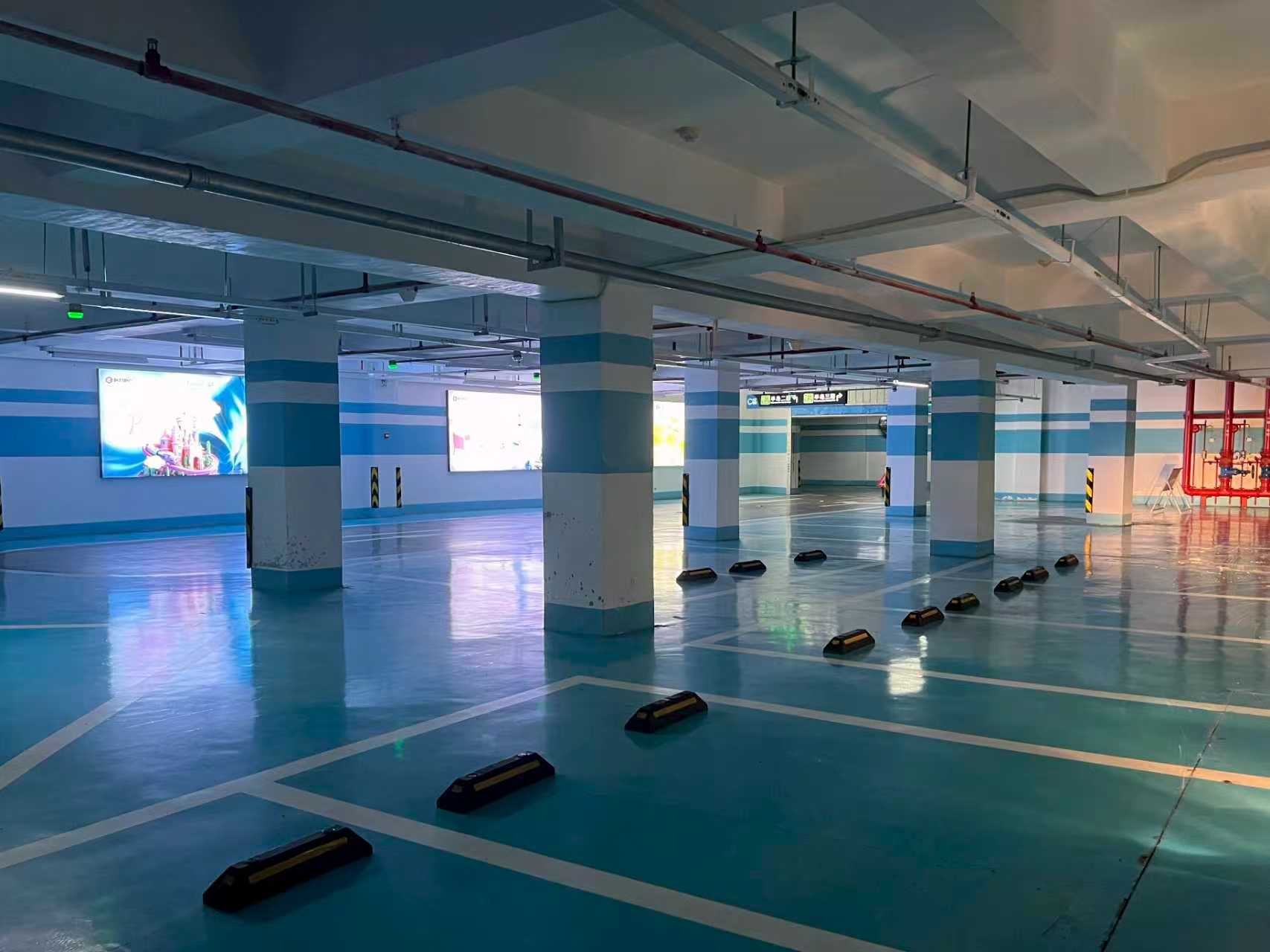}}
  \vspace{-0.1cm}
  \caption{\label{fig:realworld_arena} The real-world experiment scene photos.}
  \vspace{-1.4cm}
\end{figure}

\begin{figure*}[t]
  \centering
  \subfloat[Flight Arena, $\lambda=0.0$]{\includegraphics[width=0.66\columnwidth]{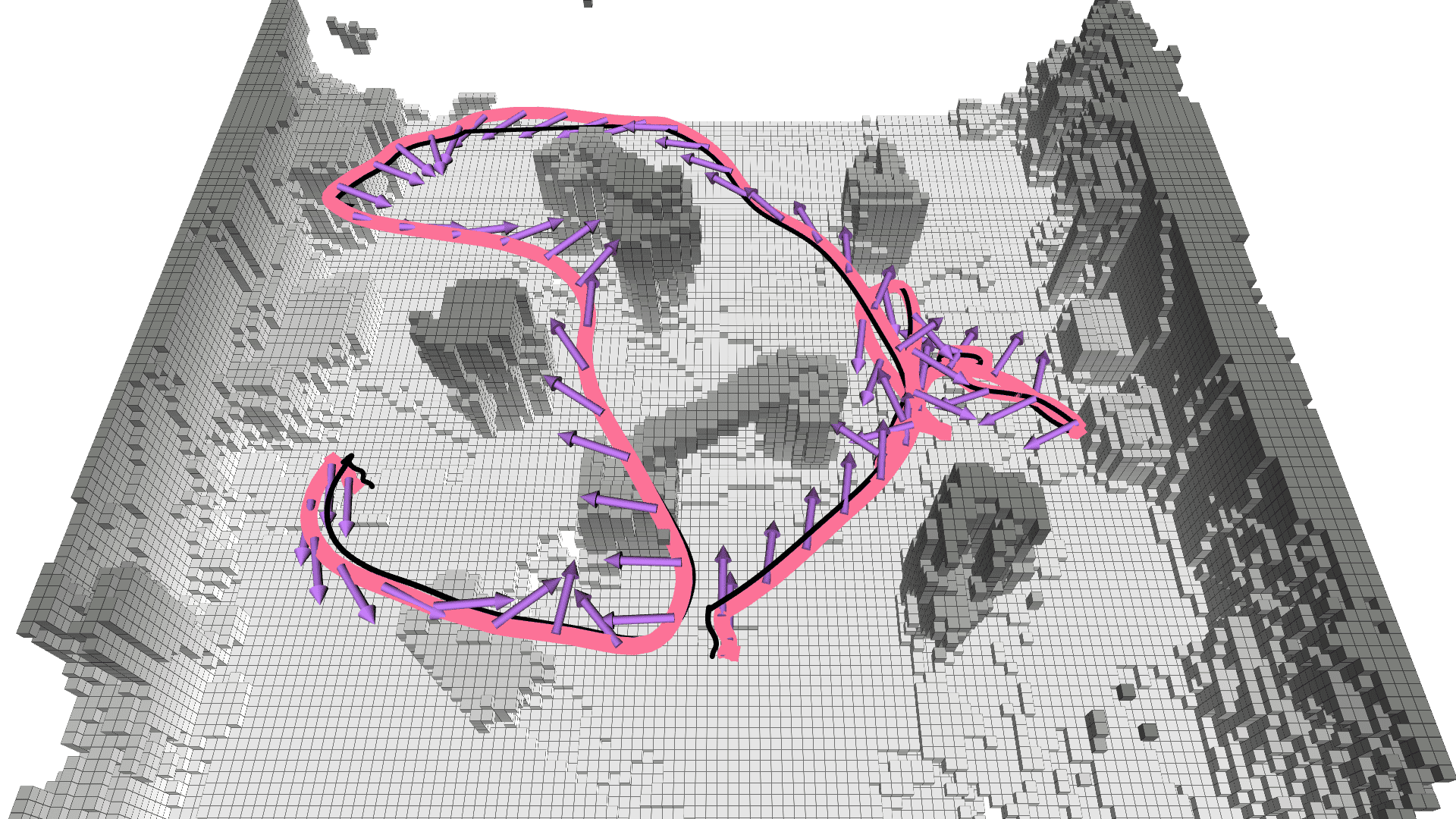}} \hfill
  \subfloat[Flight Arena, $\lambda=0.5$]{\includegraphics[width=0.66\columnwidth]{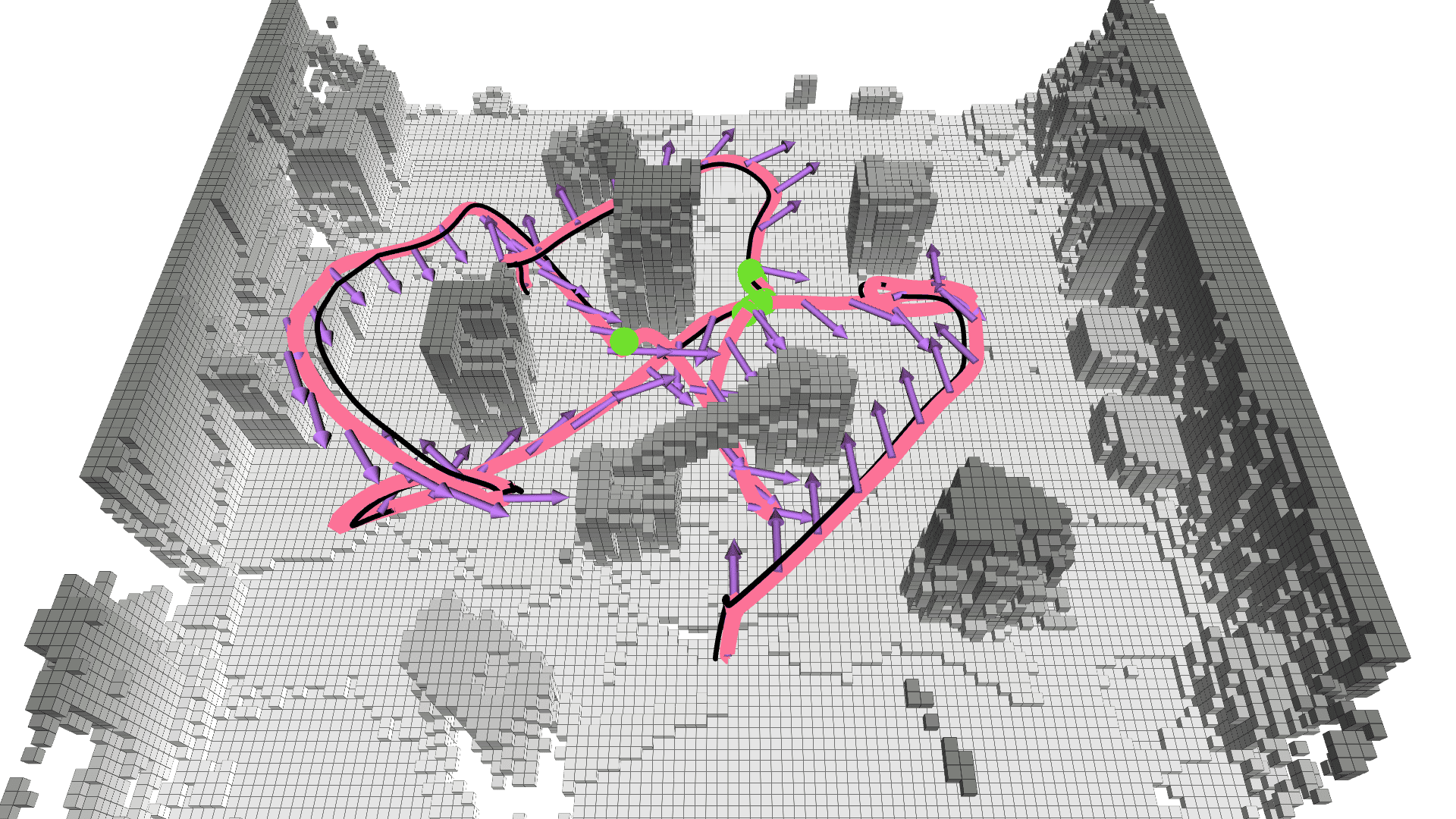}} \hfill 
  \subfloat[Flight Arena, $\lambda=1.0$]{\includegraphics[width=0.66\columnwidth]{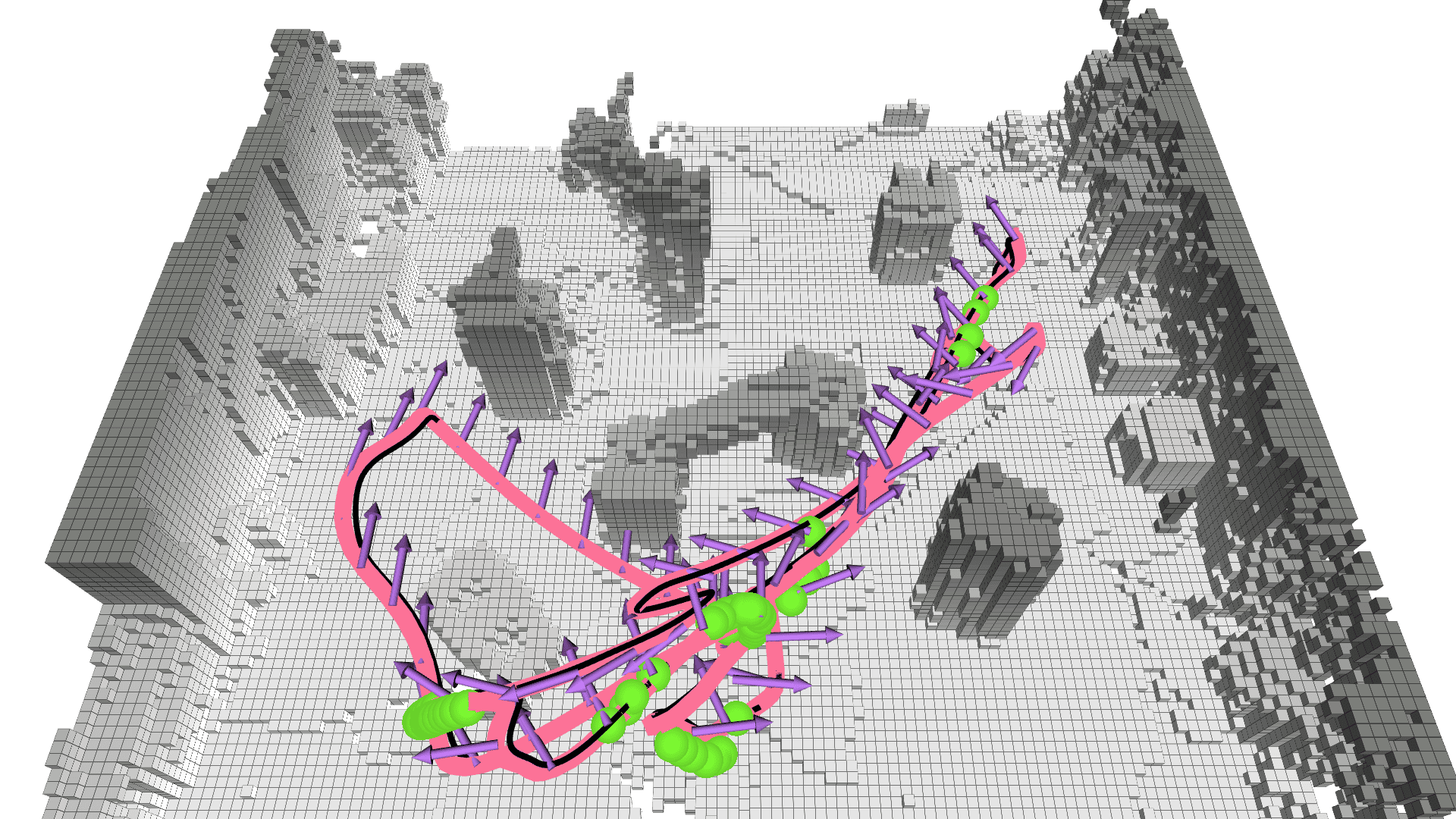}} 

  \captionsetup[subfloat]{farskip=1pt,captionskip=-2pt}
  \subfloat[Campus Hallway, $\lambda=0.0$]{\includegraphics[width=0.66\columnwidth]{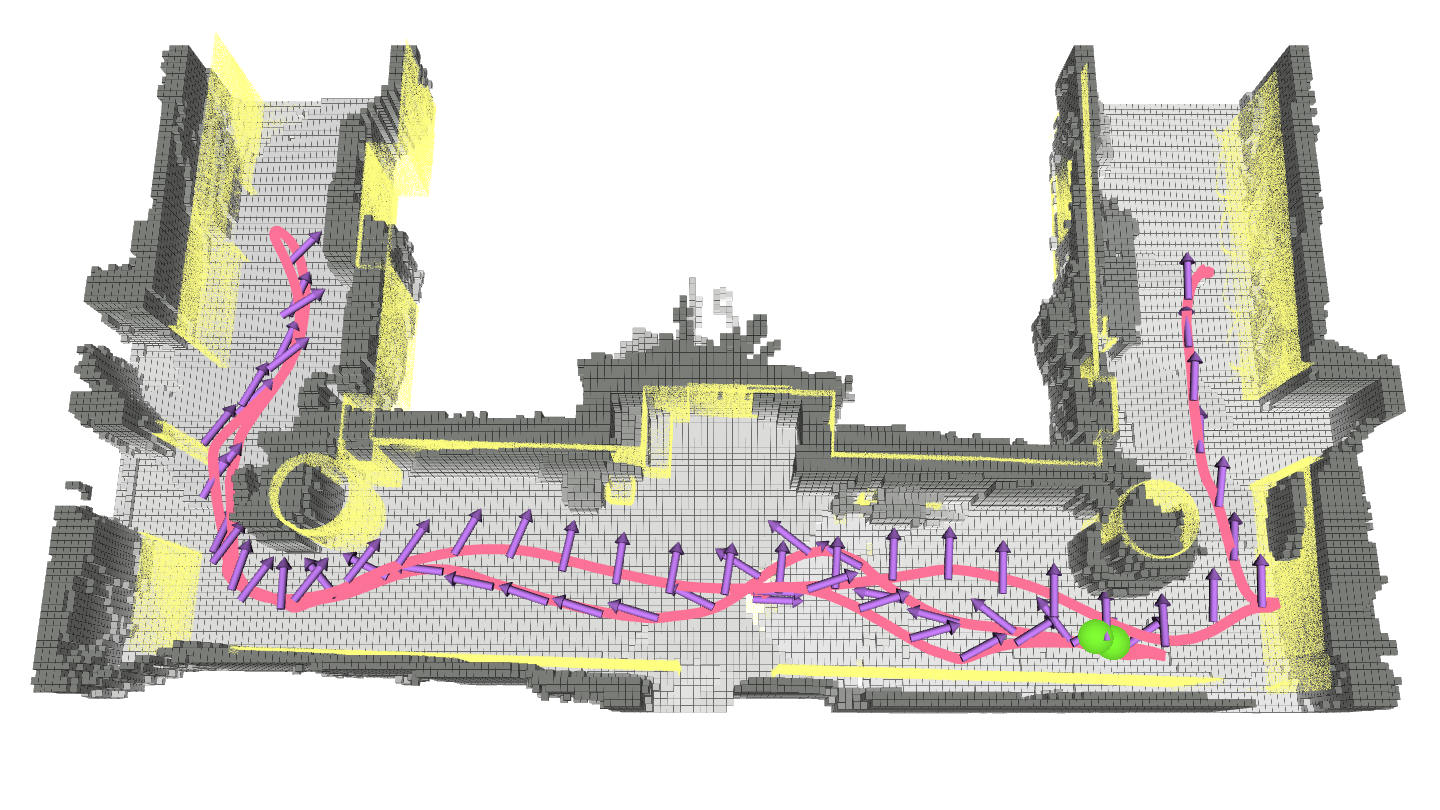}} \hfill
  \subfloat[Campus Hallway, $\lambda=0.5$]{\includegraphics[width=0.66\columnwidth]{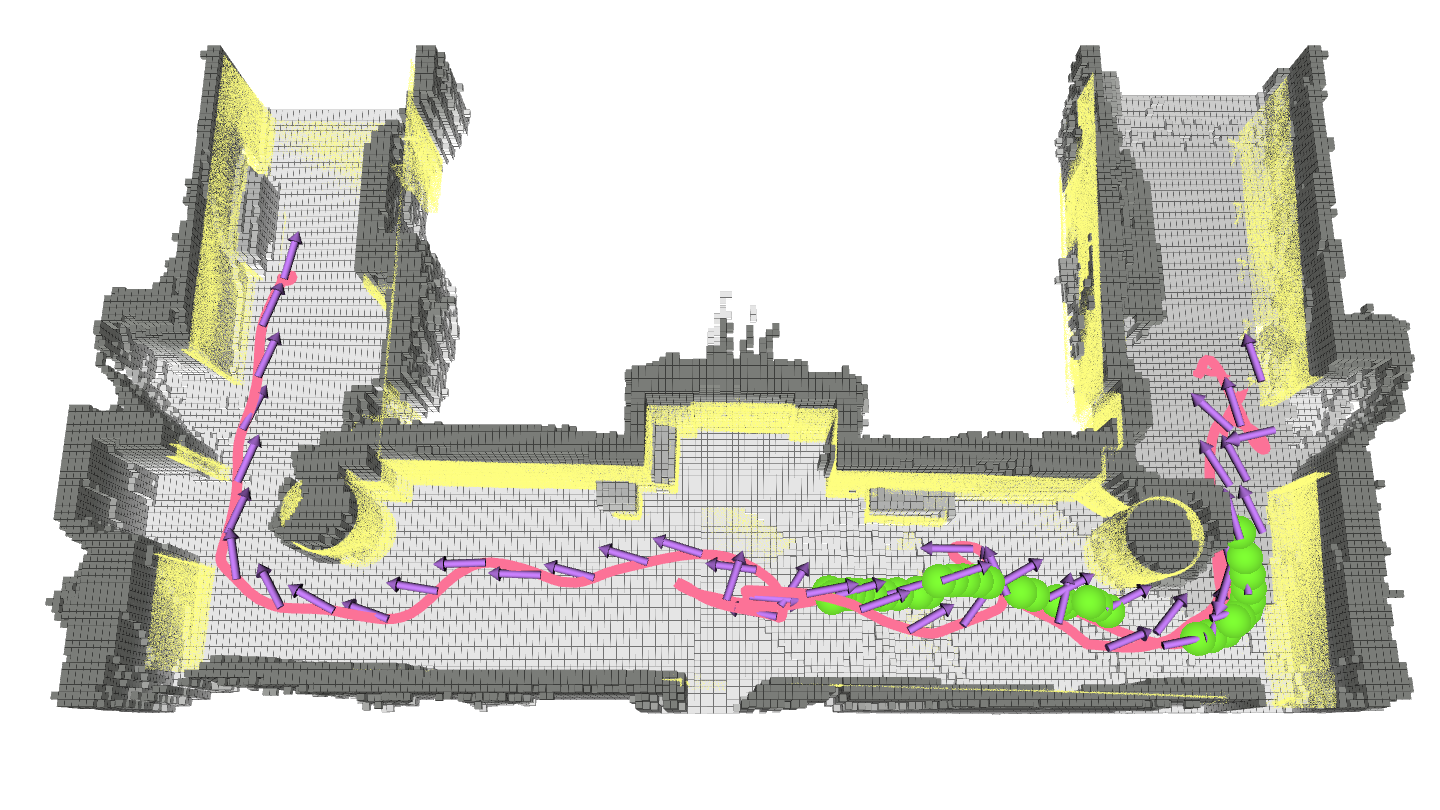}} \hfill 
  \subfloat[Campus Hallway, $\lambda=1.0$]{\includegraphics[width=0.66\columnwidth]{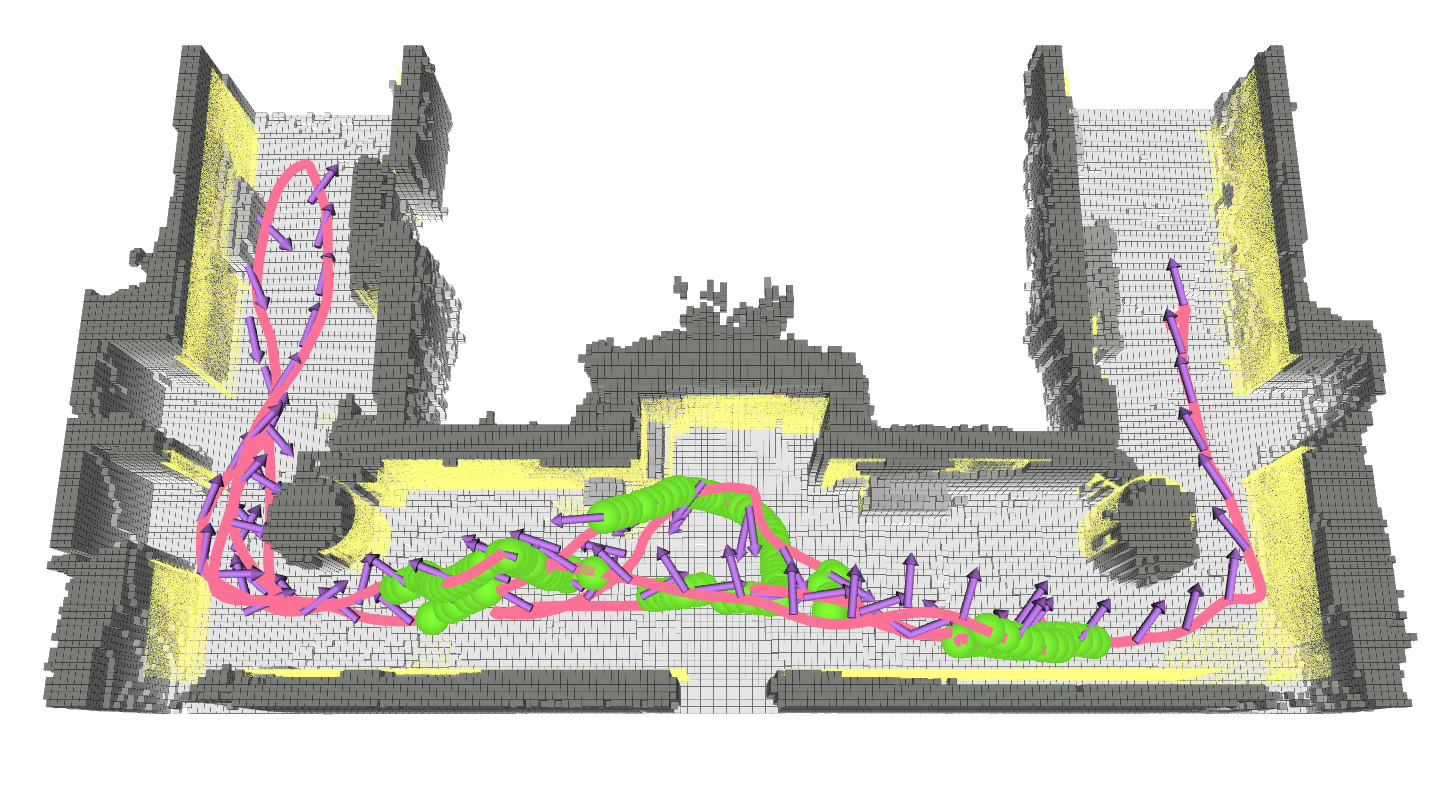}} 

  \subfloat[Parking Lot, $\lambda=0.0$]{\includegraphics[width=0.66\columnwidth]{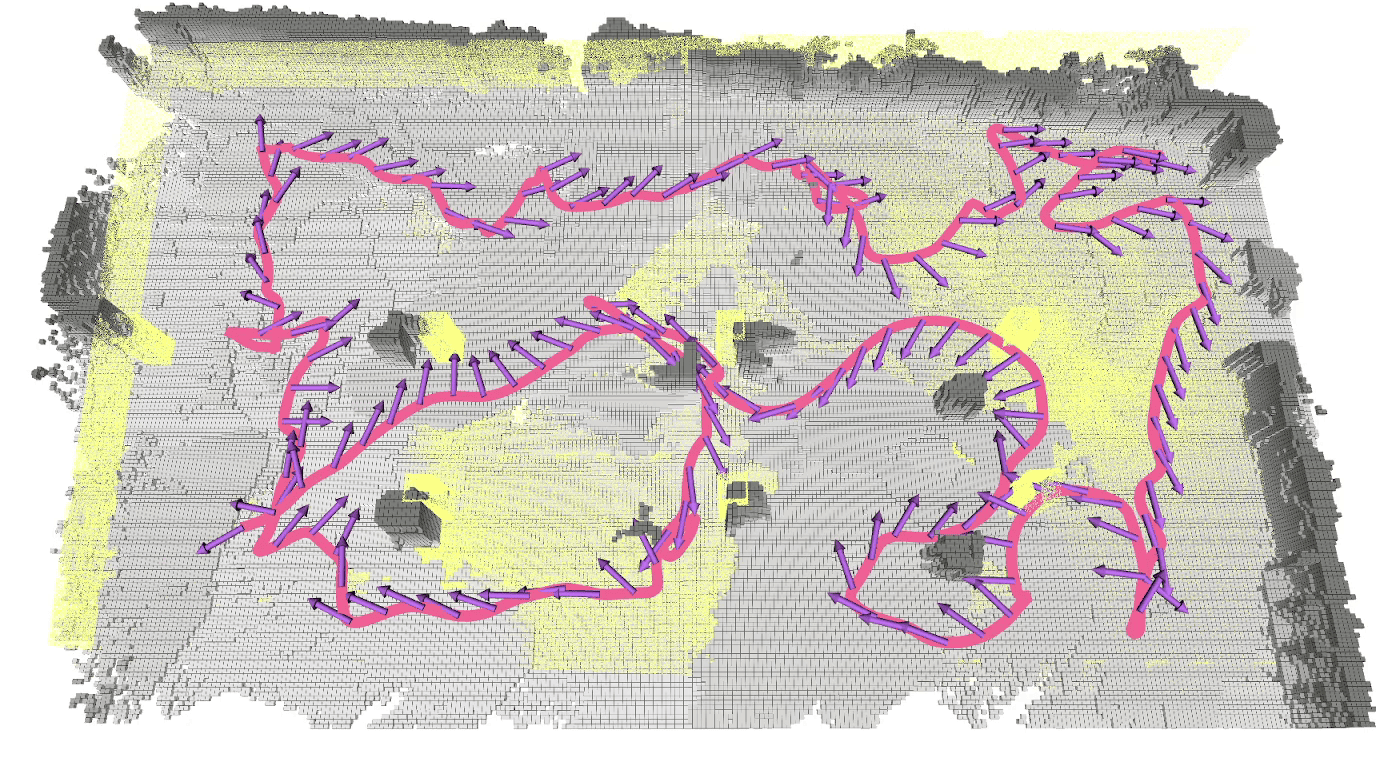}} \hfill
  \subfloat[Parking Lot, $\lambda=0.5$]{\includegraphics[width=0.66\columnwidth]{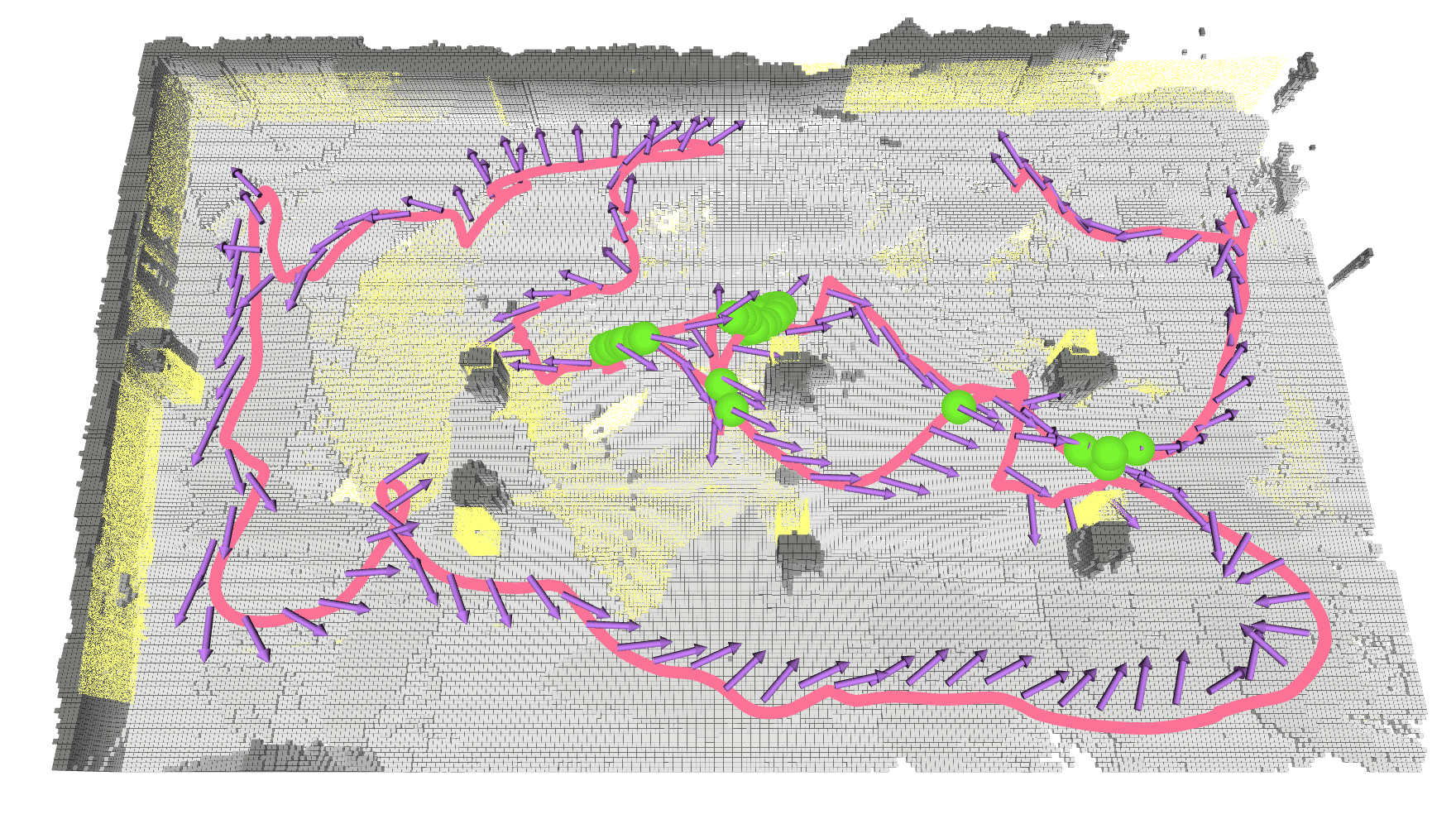}} \hfill 
  \subfloat[Parking Lot, $\lambda=1.0$]{\includegraphics[width=0.66\columnwidth]{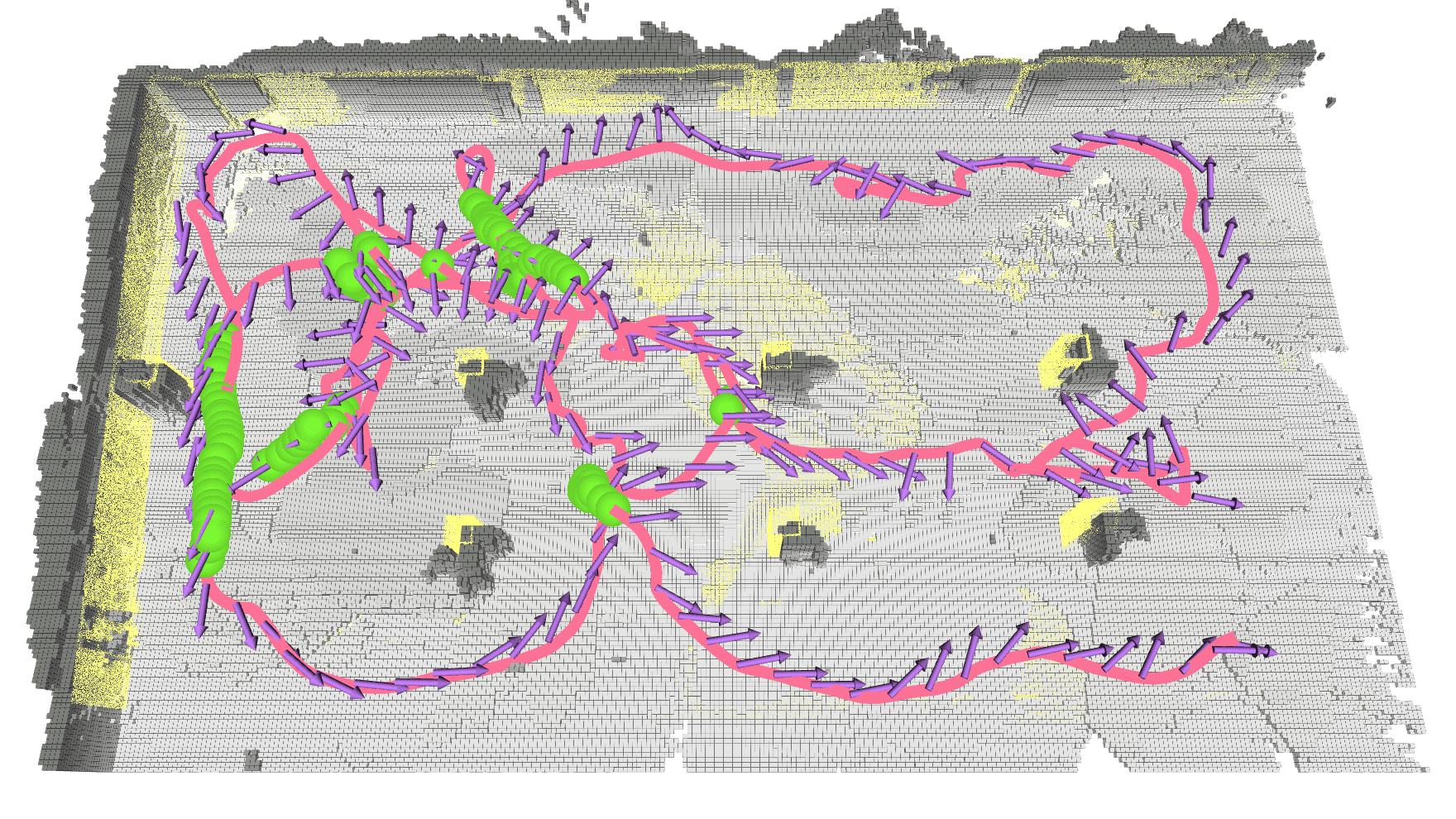}} 
  \caption{\label{fig:realworld_vis} Visualization of the real-world experiment results. The three rows show results in the Flight Arena, Campus Hallway, and Parking Lot from top to down. 
  The three columns correspond to prioritization parameter $\lambda$ of $0.0, 0.5, 1.0$ from left to right. 
  In the figures, the odometry trajectory is shown in pink, with purple arrows indicating the vehicle's yaw directions.
  The reconstructed occupancy grid is shown in grey, and detected loop closures are marked by green dots.
  The ground truth odometry trajectory is shown as a black curve in the first row, and the ground truth point cloud is represented in light yellow in the second and third rows. For more details on the complete exploration process, please refer to the supplementary video.}
  \vspace{-0.6cm}
\end{figure*}

\section{Real-World Experiments}
\label{sec:real_exp}

To validate the practical applicability of the proposed active SLAM framework RAVEN, we conduct field experiments in three challenging real-world environments for fully autonomous aerial exploration.
All experiments were performed using a customized quadrotor built on an open-source platform \cite{zhang2024uniquad}.
The quadrotor, with a take-off weight of $1.3$ kg, is equipped with a Morefine Mini PC\footnote{\url{https://morefine.com/product/morefine-ms-02/}} with an Intel i7-1260P CPU, an NxtPx4 autopilot\footnote{\url{https://github.com/HKUST-Aerial-Robotics/NxtPX4}}, a forward-looking RealSense D435i\footnote{\url{https://www.intelrealsense.com/depth-camera-d435}} camera, and a 6S LiPo battery. 
The exploration planning settings are kept consistent with those used in the simulation experiments, as detailed in Table~\ref{tab:params}. 
During the flights, the quadrotor operated fully autonomously with planning, mapping, and localization executed entirely onboard, without any human intervention or reliance on external infrastructure.

Field experiments were conducted in three scenes as illustrated in Fig.~\ref{fig:realworld_arena}.
The first scene is a cluttered \textit{Flight Arena} containing randomly placed boxes and bridges.
This arena is equipped with an OptiTrack\footnote{\url{https://www.optitrack.com/}} motion capture system, which provides ground truth vehicle pose used solely for quantitative evaluation of ATE and not used by the quadrotor during flight.
The second scene is a \textit{Campus Hallway} featuring two cylindrical pillars and visually challenging plain white walls, and the third scene is a large underground \textit{Parking Lot}.
Localization accuracy in the latter two scenes is indirectly evaluated by the RMSE of nearest neighbor distance (NND) between the reconstructed volumetric map and a ground truth point cloud captured with a handheld 3D laser scanner\footnote{\url{https://skylandx.com/metacam-air/}}.

\begin{figure*}[t]
  \centering
  \captionsetup[subfloat]{farskip=-2pt,captionskip=-8pt}
  \subfloat[]{\includegraphics[height=0.5\columnwidth]{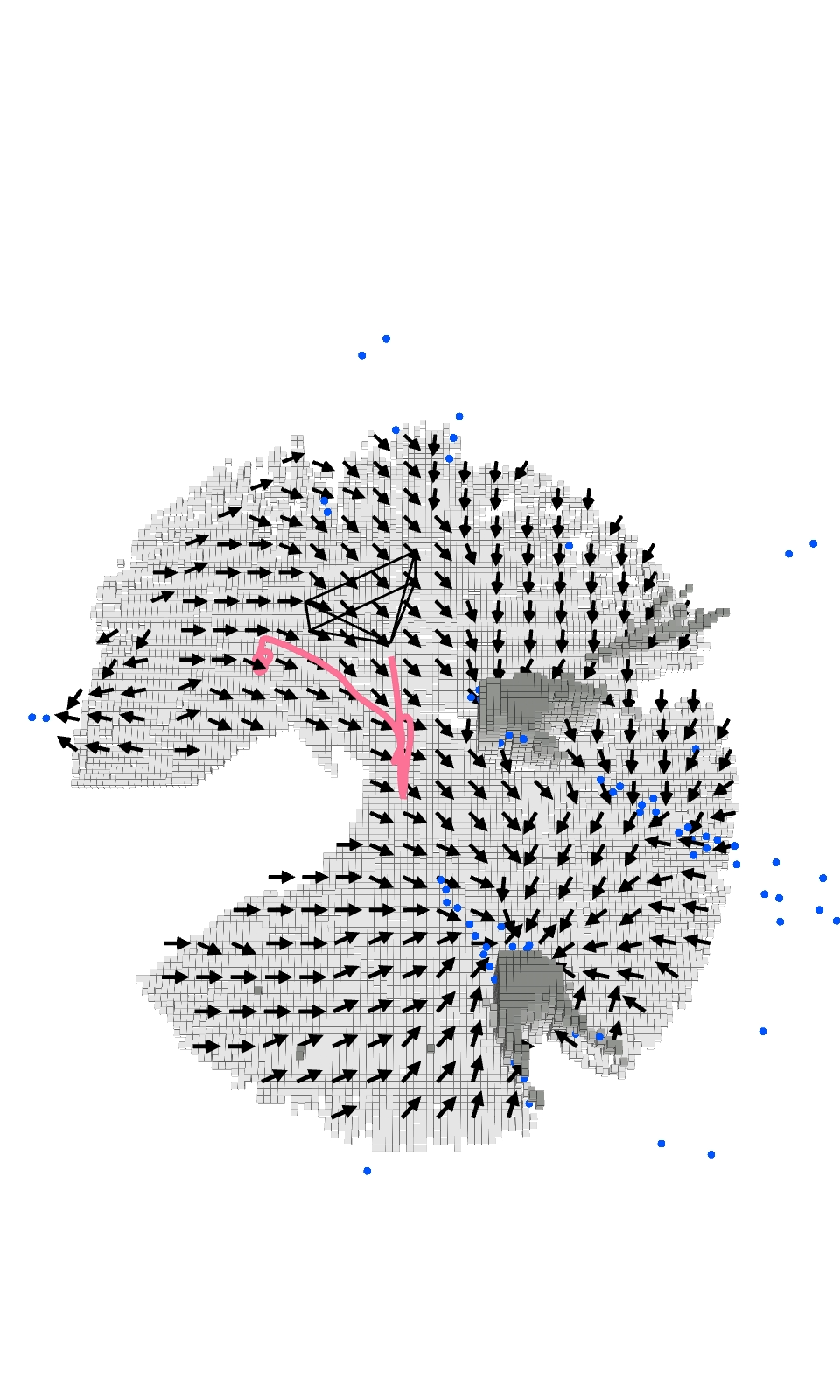}} \hspace{0.2cm}
  \subfloat[]{\includegraphics[height=0.5\columnwidth]{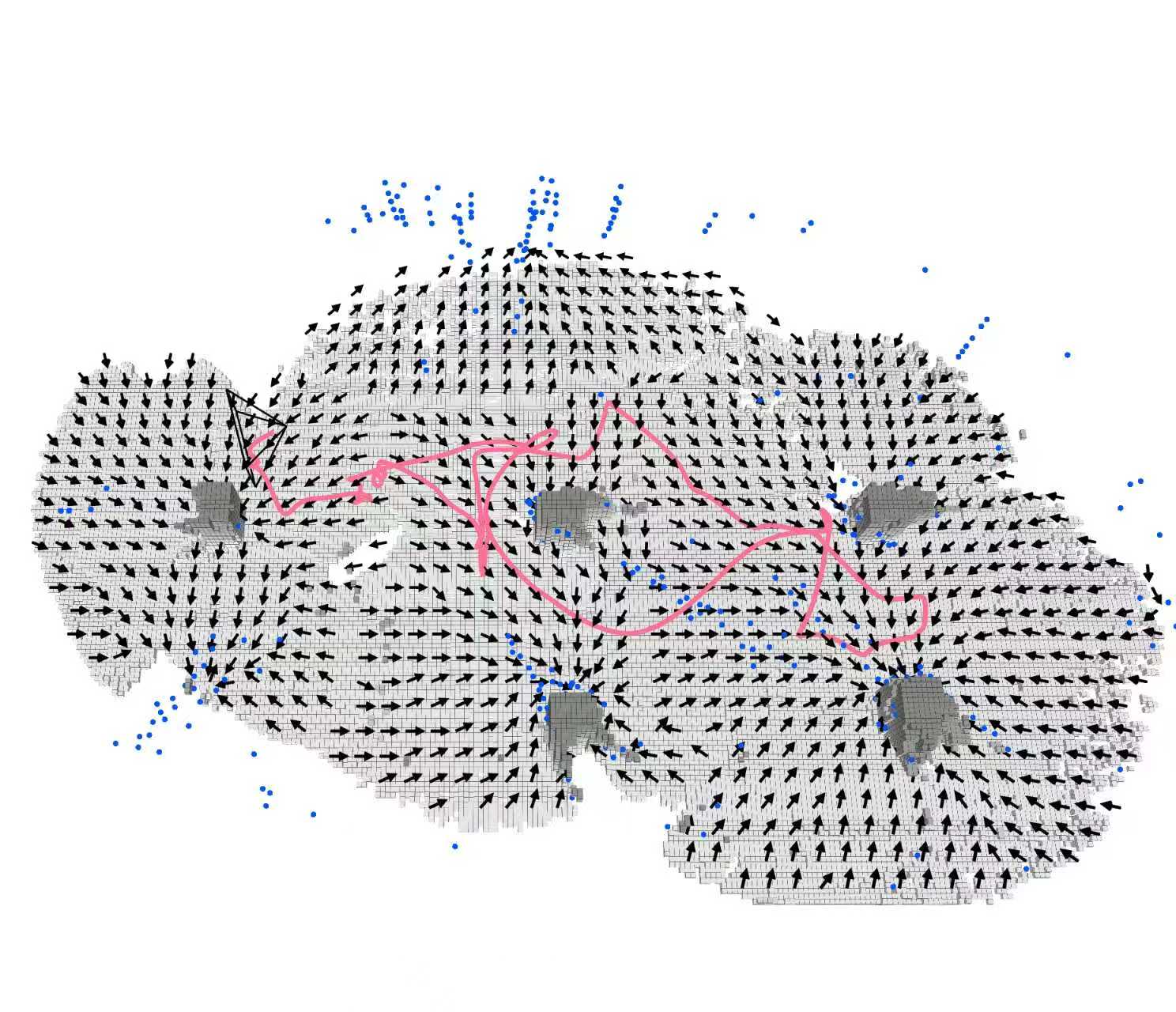}} \hspace{0.2cm}
  \subfloat[]{\includegraphics[height=0.5\columnwidth]{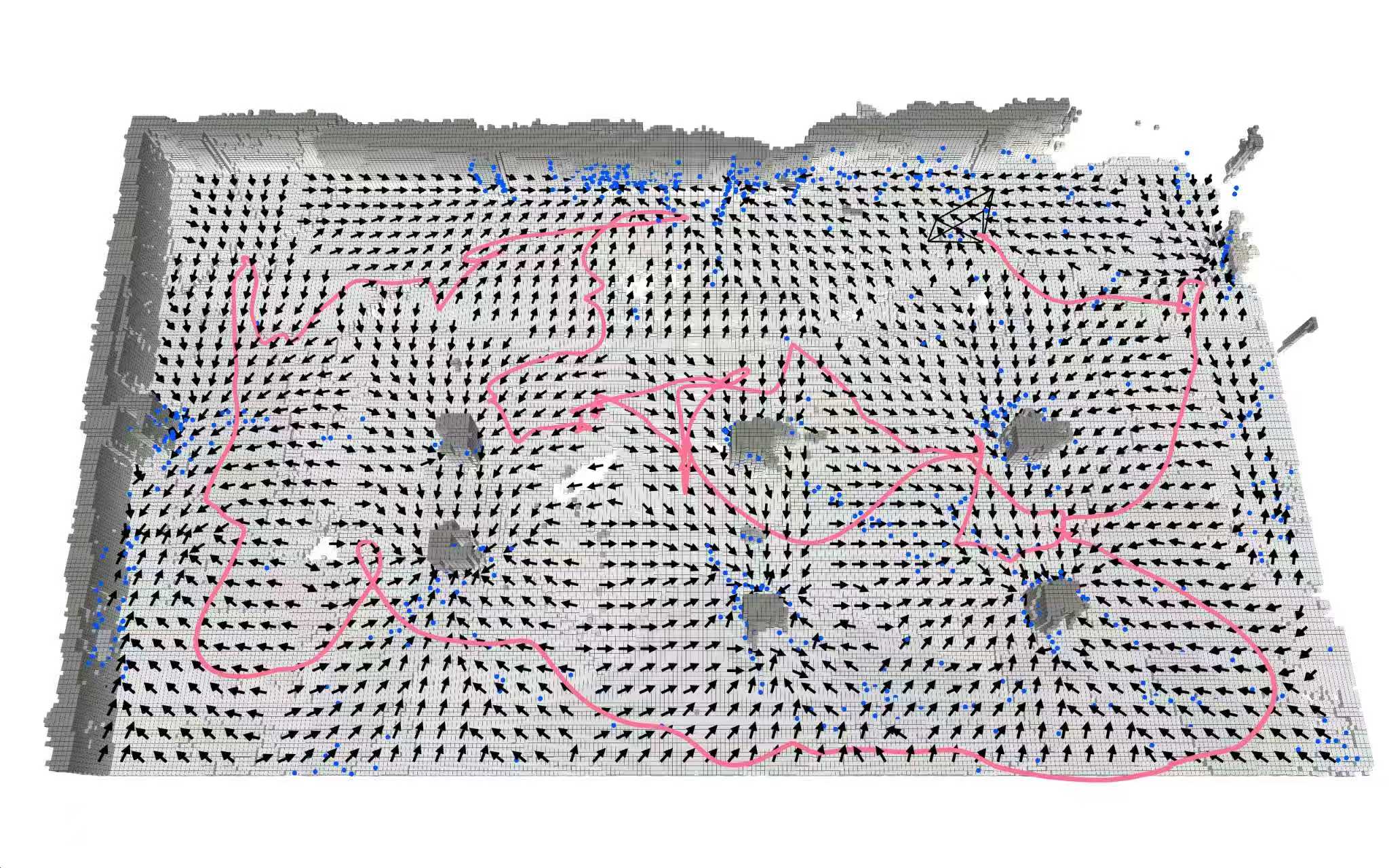}} 
  \vspace{-0.2cm}
  \caption{\label{fig:realworld_ofif} Visualization of the OFIF incremental construction process, illustrating the optimal observation directions in black arrows and localizable features in blue dots during exploration of the \textit{Parking Lot} scene with $\lambda=0.5$.}
  \vspace{-0.4cm}
\end{figure*}

We deploy the proposed active SLAM framework RAVEN for autonomous aerial exploration in the three real-world scenes under three prioritization parameters $\lambda = 0.0, 0.5, 1.0$.
In the \textit{Flight Arena} with ground truth pose, RAVEN was tested for five trials for each prioritization setting for a quantitative evaluation.
In the \textit{Campus Hallway} and \textit{Parking Lot}, three trials were performed per setting.
The result statistics in Table~\ref{tab:realworld_data} demonstrate that increasing $\lambda$ leads to reduced exploration efficiency, evidenced by longer exploration time and distance, and higher localization accuracy, reflected in lower ATE or NND values, more looped image pairs, and higher CPG graph information.
Fig.~\ref{fig:realworld_vis} displays the reconstructed maps and executed trajectories in all three scenes under different $\lambda$ settings.
In the \textit{Flight Arena} (Fig.~\ref{fig:realworld_vis}(a)-(c)), although more loop closures were detected with the increase of $\lambda$, as indicated by green dots, the reconstructed maps were of good quality across all settings due to the small scene scale and limited localization drift.
In more challenging scenes \textit{Campus Hallway} and \textit{Parking Lot}, the reconstructed maps exhibit stronger differences.
As shown in Fig.~\ref{fig:realworld_vis}(d)-(i), in comparison with the ground truth point clouds in light yellow, maps reconstructed with $\lambda=0.0$ suffered from noticeable distortions due to significant visual odometry drift, whereas $\lambda=1.0$ yielded high-fidelity reconstructed maps, reducing the average NND error by $50\%$ and $62.5\%$ respectively.
For instance, the two cylindrical pillars in \textit{Campus Hallway} were both deformed in Fig.~\ref{fig:realworld_vis}(d) under $\lambda=0.0$, but were accurately reconstructed in Fig.~\ref{fig:realworld_vis}(f) under $\lambda=1.0$.
In \textit{Parking Lot} scene with larger scale, the rectangular pillars and walls were severely distorted in Fig.~\ref{fig:realworld_vis}(g), yet were well preserved in Fig.~\ref{fig:realworld_vis}(i).
Moreover, the incremental construction of the OFIF during one exploration trial in the \textit{Parking Lot} with $\lambda=0.5$ is visualized in Fig.~\ref{fig:realworld_ofif}.
As the OFIF is designed for constant-time information queries, the visualization here illustrates the optimal observation directions obtained by uniform sampling.
Leveraging the OFIF, the perception-aware navigation planner effectively avoids directing the quadrotor toward textureless areas with poor visual features, such as the plain white wall at the bottom of Fig.~\ref{fig:realworld_vis}(d)-(f).
The field experiments validate RAVEN's capability for real-world autonomous aerial exploration, effectively balancing exploration efficiency and localization accuracy according to the prioritization parameter $\lambda$, while maintaining robust visual odometry during the flights.

\section{Conclusions}
\label{sec:conclude}
This article presents RAVEN, a unified active SLAM framework for autonomous aerial exploration that systematically accounts for localization uncertainty throughout the planning workflow.
The framework addresses the exploration-exploitation dilemma by selecting the global tour among Pareto-optimal solutions according to a user-defined prioritization parameter.
A perception-aware local planner generates execution trajectories that minimize visual state estimation drift while ensuring safety and feasibility.
An online Fisher information field is incrementally constructed to provide constant-time queries of localization information from an observation to support real-time planning.
Extensive benchmark experiments and ablation studies demonstrate that RAVEN effectively balances exploration efficiency and localization accuracy, offering a spectrum of performance across prioritizations for general-purpose autonomous exploration.
Fully onboard experiments with a customized quadrotor further validate RAVEN's practical robustness in challenging real-world scenarios including partially feature-poor regions.
Nevertheless, RAVEN is scoped to visually tractable environments under normal indoor lighting and is not intended for scenes with almost no visual features where visual SLAM becomes fundamentally infeasible.
Addressing active SLAM in such extreme conditions with additional sensing modalities, such as LiDAR or radar, represents a potential direction for future research.

\appendix
\renewcommand{\thesubsection}{\Alph{subsection}}

\subsection{Alternative Definitions of D-optimality}
\label{apx:a}

\begin{proof}[Proof of (\ref{eqn:dopt}) and (\ref{eqn:pgo_fim})] The following proves the eigenvalue expression of D-optimality. From the general definition of optimality criteria (\ref{eqn:p-opt}),
\begin{align*}
    & \Dopt(\matr{\Sigma}) \triangleq \lim_{p\to 0}\Bigl(\frac{1}{n}\tr(\matr{\Sigma}^p)\Bigr)^\frac{1}{p} 
    = \lim_{p\to 0}\Bigl(\frac{1}{n}\sum_{i=1}^{n}\lambda_i^p\Bigr)^\frac{1}{p} \\
    & = \lim_{p\to 0}\Bigl(\frac{1}{n}\sum_{i=1}^{n}\exp(p\log\lambda_i)\Bigr)^\frac{1}{p} 
    \approx \lim_{p\to 0}\Bigl(\frac{1}{n}\sum_{i=1}^{n}(1 + p\log\lambda_i)\Bigr)^\frac{1}{p} \\
    & = \lim_{p\to 0}\exp\Bigl(\frac{1}{p}\log(1 + \frac{p}{n}\sum_{i=1}^{n}\log\lambda_i)\Bigr) \\
    & \approx \lim_{p\to 0}\exp\Bigl(\frac{1}{p}\cdot\frac{p}{n}\sum_{i=1}^{n}\log\lambda_i\Bigr)
    = \exp\Bigl(\frac{1}{n}\sum_{i=1}^{n}\log\lambda_i\Bigr),
\end{align*}
where the two approximations are derived from the Taylor expansion. We further arrange this equation for special cases:

1) When $\matr{\Sigma}$ is non-singular,
\begin{align*}
    \Dopt(\matr{\Sigma}) &= \exp\Bigl(\frac{1}{n}\sum_{i=1}^{n}\log\lambda_i\Bigr) = \exp\Bigl(\frac{1}{n}\log\prod_{i=1}^{n}\lambda_i\Bigr) \\
    & = \exp\Bigl(\frac{1}{n}\log(\det(\matr{\Sigma}))\Bigr) = (\det(\matr{\Sigma}))^\frac{1}{n}.
\end{align*}

2) When $\matr{\Sigma} = \matr{L}$ is the Laplacian matrix of a connected graph, the eigenvalues of $\matr{L}$ are $(\lambda_1=0, \lambda_2,\dots, \lambda_n)$.
\begin{align*}
    \Dopt(\matr{L}) & = \lim_{p\to 0}\Bigl(\frac{1}{n}\sum_{i=2}^{n}\lambda_i^p\Bigr)^\frac{1}{p}
    = \exp\Bigl(\frac{1}{n}\sum_{i=2}^{n}\log\lambda_i\Bigr) \\
    & = \Bigl(\prod_{i=2}^{n}\lambda_i\Bigr)^{\frac{1}{n}}
    = (n t(\mathcal{G}))^{\frac{1}{n}} = (n \det(\matr{L}'))^{\frac{1}{n}},
\end{align*}
where the last two equations come from the Kirchhoff's Matrix Tree Theorem mentioned in Sec.~\ref{subsec:preli_graph_theory}.

From these two cases, it is important to note that while the term D-optimality originates from the concept of determinant, it is not necessarily true that $\Dopt(\matr{\Sigma})$ directly relates to $\det{\matr{\Sigma}}$.
Particularly, a singular Laplacian matrix $\matr{L}$ has a determinant of zero, but its D-optimality is still well-defined.
\end{proof}

\subsection{Closed-form Expression of FIM}
\label{apx:fim_closed_form}
\begin{proof}[Proof of (\ref{eqn:fim_closed_form})] The following proves the closed-form expression of T-optimality of the FIM obtained from the right perturbation Jacobian. For brevity, we omit the subscript $i$ and move the coordinate indicator from superscript to subscript in (\ref{eqn:right_perturb}). Let $\matr{M} \triangleq \frac{1}{n^2} \matr{I}_{3}-\frac{1}{n^4} \vect{p}_\text{c} \vect{p}_\text{c}^T$, we have
\begin{equation*}
    \matr{J}^T\matr{J} = \begin{bmatrix}-\matr{I}_{3} \\ (\vect{p}_\text{c}^{\wedge})^T \end{bmatrix} \matr{M} \begin{bmatrix} -\matr{I}_{3} & \vect{p}_\text{c}^{\wedge} \end{bmatrix}
    = \begin{bmatrix} \matr{M}  & \cdots \\ \cdots & -\vect{p}_\text{c}^{\wedge} \matr{M} \vect{p}_\text{c}^{\wedge} \end{bmatrix}
\end{equation*}
Then since the $\sigma$ in (\ref{eqn:jacobian_and_fim}) is a constant scalar, we have
\begin{align*}
    \tr(\mathcal{I}) &\approx \tr(\matr{J}^T\matr{J}) = \tr(\matr{M}) + \tr(-\vect{p}_\text{c}^{\wedge} \matr{M} \vect{p}_\text{c}^{\wedge}), \\
    \tr(\matr{M}) &= \tr(\frac{1}{n^2} \matr{I}_{3}-\frac{1}{n^4}\vect{p}_\text{c} \vect{p}_\text{c}^T) = \frac{3}{n^2}-\frac{n^2}{n^4} = \frac{2}{n^2}, \\
    \tr(-\vect{p}_\text{c}^{\wedge} \matr{M} \vect{p}_\text{c}^{\wedge}) &= \tr(-\frac{1}{n^2}\vect{p}_\text{c}^{\wedge}\vect{p}_\text{c}^{\wedge}) + \tr(\frac{1}{n^4}\vect{p}_\text{c}^{\wedge} \vect{p}_\text{c} \vect{p}_\text{c}^T \vect{p}_\text{c}^{\wedge}) \\
    &= -\frac{1}{n^2}(-2n^2) + 0 = 2
\end{align*}
where $\vect{p}_\text{c}^{\wedge} \vect{p}_\text{c} \vect{p}_\text{c}^T \vect{p}_\text{c}^{\wedge} = -\vect{p}_\text{c}\times \vect{p}_\text{c} \cdot (\vect{p}_\text{c}\times \vect{p}_\text{c})^T = \vect{0}$
\end{proof}

\subsection{Covariance Computation of CPG Edges}
\label{apx:cpg_cov}
For two consecutive SLAM frames, given their 3D feature correspondences $\{\vect{p}\}, \{\vect{q}\}$ and relative transformation $\matr{T} = (\matr{R}, \vect{t})$, the following propagates the reprojection residual covariance $\Sigma_\matr{Y}$ to relative transformation covariance $\Sigma_\matr{T}$.
Supposing $\boldsymbol{\xi} = (\boldsymbol{\theta}, \boldsymbol{\rho})^T \in \mathfrak{se}(3)$ is the linearized transformation error and $\boldsymbol{\epsilon}_i$ is the correspondence noise, the reprojection residual $\vect{y}_i$ can be written as
\begin{align*}
    \vect{q}_i &= (\matr{I}+\boldsymbol{\theta}^\wedge) \matr{R} \vect{p}_i + (\vect{t} + \boldsymbol{\rho}) + \boldsymbol{\epsilon}_i,\\
    \vect{y}_i &= \vect{q}_i - (\matr{R}\vect{p}_i + \vect{t}) = \boldsymbol{\theta}^\wedge \matr{R} \vect{p}_i + \boldsymbol{\rho} + \boldsymbol{\epsilon}_i \\
    &= -(\matr{R} \vect{p}_i)^\wedge \boldsymbol{\theta} + \boldsymbol{\rho} + \boldsymbol{\epsilon}_i = \begin{bmatrix} -(\matr{R} \vect{p}_i)^\wedge & \matr{I} \end{bmatrix} \boldsymbol{\xi} + \boldsymbol{\epsilon}_i \\
    &\triangleq \matr{J}_i \boldsymbol{\xi} + \boldsymbol{\epsilon}_i.
\end{align*}
Stacking all correspondences, we have the linear system and the desired covariance
\begin{equation*}
    \matr{Y} = \matr{J} \xi + \boldsymbol{\epsilon},\quad
    \Sigma_\matr{T} = (\matr{J}^T \Sigma_\matr{Y}^{-1} \matr{J})^{-1}.
\end{equation*}


\addtolength{\textheight}{0.cm}   

\newlength{\bibitemsep}\setlength{\bibitemsep}{0.0\baselineskip}
\newlength{\bibparskip}\setlength{\bibparskip}{0.1pt}
\let\oldthebibliography\thebibliography
\renewcommand\thebibliography[1]{%
\oldthebibliography{#1}%
\setlength{\parskip}{\bibitemsep}%
\setlength{\itemsep}{\bibparskip}%
}

\bibliography{ref}

\end{document}